\documentclass[10pt,journal,compsoc]{IEEEtran} 

\usepackage[nocompress]{cite}
\usepackage{multirow}
\usepackage{booktabs}
\usepackage{threeparttable}
\usepackage{array}
\newcolumntype{H}{>{\setbox0=\hbox\bgroup}c<{\egroup}@{}}
\newcommand{\PreserveBackslash}[1]{\let\temp=\\#1\let\\=\temp}
\newcolumntype{C}[1]{>{\PreserveBackslash\centering}p{#1}}
\newcolumntype{R}[1]{>{\PreserveBackslash\raggedleft}p{#1}}
\newcolumntype{L}[1]{>{\PreserveBackslash\raggedright}p{#1}}
\PassOptionsToPackage{table,xcdraw}{xcolor}
\usepackage{xcolor}
\usepackage{comment}
\usepackage{graphicx}
\usepackage{blindtext}
\usepackage{adjustbox}
\usepackage{wrapfig}
\usepackage{amsmath}
\usepackage{amssymb}
\usepackage{afterpage}
\usepackage{algorithm}
\usepackage{algorithmic}
\usepackage{amsfonts}
\usepackage{bm}
\usepackage{caption}
\usepackage{overpic}
\usepackage{stfloats}
\usepackage{subcaption}
\usepackage{textcomp}
\usepackage{url}
\usepackage{verbatim}
\usepackage{xspace}
\usepackage[colorlinks,linkcolor=black,anchorcolor=black,citecolor=black]{hyperref}

\makeatletter
\DeclareRobustCommand\onedot{\futurelet\@let@token\@onedot}
\def\@onedot{\ifx\@let@token.\else.\null\fi\xspace}
\def\eg{\emph{e.g}\onedot}

\def\etc{\emph{etc}\onedot}

\makeatother

\begin{document}

\title{UNICE: Training A Universal \\ Image Contrast Enhancer}

\author{Ruodai~Cui,
        and~Lei~Zhang,~\IEEEmembership{Fellow,~IEEE}%
\IEEEcompsocitemizethanks{\IEEEcompsocthanksitem R. Cui and L. Zhang are with the Department of Computing, The
Hong Kong Polytechnic University, Hong Kong. (email: ruodai.cui@connect.polyu.hk, cslzhang@comp.polyu.edu.hk).
}}

\IEEEtitleabstractindextext{%
\begin{abstract}
Existing image contrast enhancement methods are typically designed for specific tasks such as under-/over-exposure correction, low-light and backlit image enhancement, etc. The learned models, however, exhibit poor generalization performance across different tasks, even across different datasets of a specific task. It is important to explore whether we can learn a universal and generalized model for various contrast enhancement tasks. In this work, we observe that the common key factor of these tasks lies in the need of exposure and contrast adjustment, which can be well-addressed if high-dynamic range (HDR) inputs are available. We hence collect  46,928 HDR raw images from public sources, and render 328,496 sRGB images to build multi-exposure sequences (MES) and the corresponding pseudo sRGB ground-truths via multi-exposure fusion. Consequently, we train a network to generate an MES from a single sRGB image, followed by training another network to fuse the generated MES into an enhanced image. Our proposed method, namely \textbf{UN}iversal \textbf{I}mage \textbf{C}ontrast \textbf{E}nhancer (\textbf{UNICE}), is free of  costly human labeling. However, it demonstrates significantly stronger generalization performance than existing image contrast enhancement methods across and within different tasks, even outperforming manually created ground-truths in multiple no-reference image quality metrics. The dataset, code and model are available at \href{https://github.com/BeyondHeaven/UNICE}{https://github.com/BeyondHeaven/UNICE}.
\end{abstract}
    
}

\maketitle
\IEEEdisplaynontitleabstractindextext
\IEEEpeerreviewmaketitle

\begin{figure}[!t]
    \centering
    \begin{subfigure}[t]{0.22\textwidth}
        \centering
        \includegraphics[width=\textwidth]{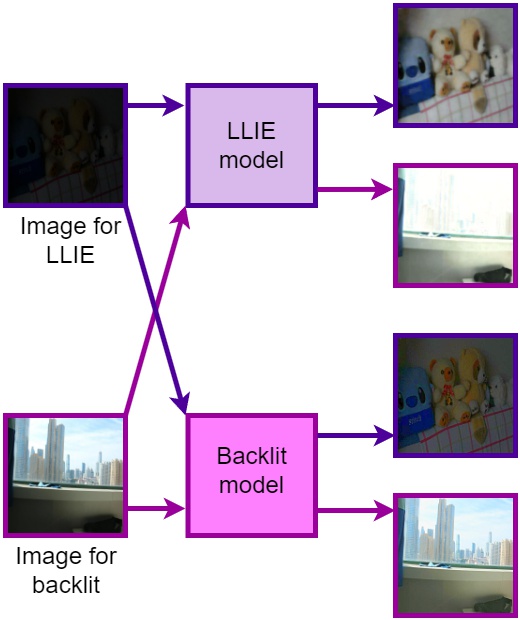}
        \caption{Cross-task test}
        \label{fig:cross_task}
    \end{subfigure}
    \vrule width 0.5pt 
    \begin{subfigure}[t]{0.22\textwidth}
        \centering
        \includegraphics[width=\textwidth]{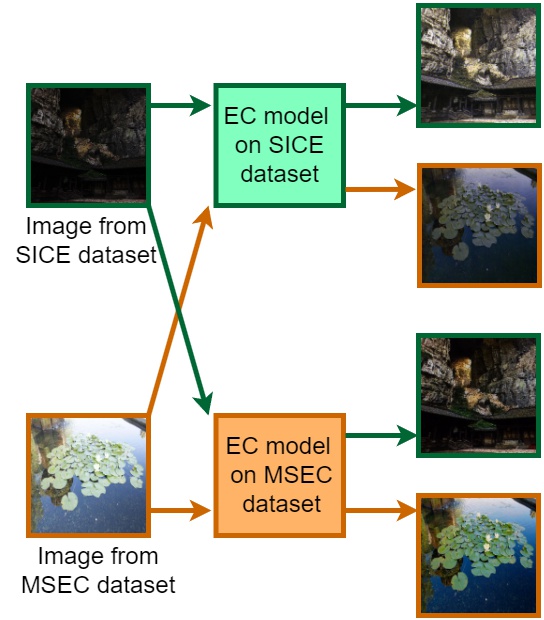}
        \caption{Cross-dataset test}
        \label{fig:cross_dataset}
    \end{subfigure}
    \begin{subfigure}[t]{0.45\textwidth}
        \centering
        \includegraphics[width=\textwidth]{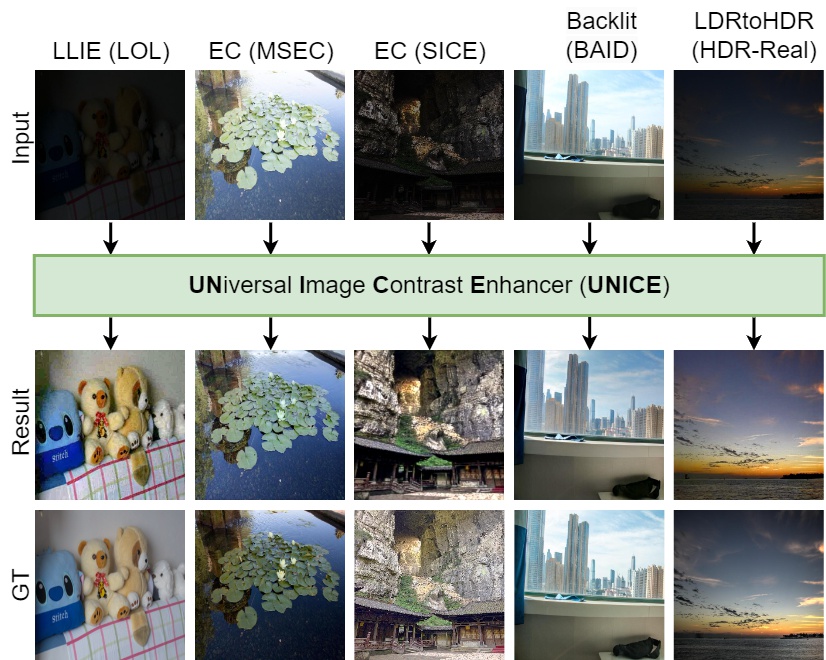}
        \caption{Comparison between our results and manually labeled ground-truths (GTs)}
        \label{fig:our_results}
    \end{subfigure}
    \caption{(a) Existing methods typically focus on specific tasks, such as low-light image enhancement (LLIE) and exposure correction (EC), but often perform poorly across different tasks. (b) Performance may also degrade across datasets of the same task, such as MSEC~\cite{afifi2021msec} and SICE~\cite{cai2018sice} for EC. (c) Our model performs well across tasks and datasets, such as LOL~\cite{wei2018lol}, MSEC~\cite{afifi2021msec}, SICE~\cite{cai2018sice}, BAID~\cite{lv2022backlitnet} and HDR-Real~\cite{liu2020reverse}, even exceeding manually labeled ground-truth in NR-IQA metrics. FLLIE~\cite{fu2023llie} and CSEC~\cite{li_2024_cvpr_csec} are used as baselines for LLIE and EC, respectively.}
\end{figure}

\begin{figure}[ht]
    \centering
    \includegraphics[width=1\linewidth]{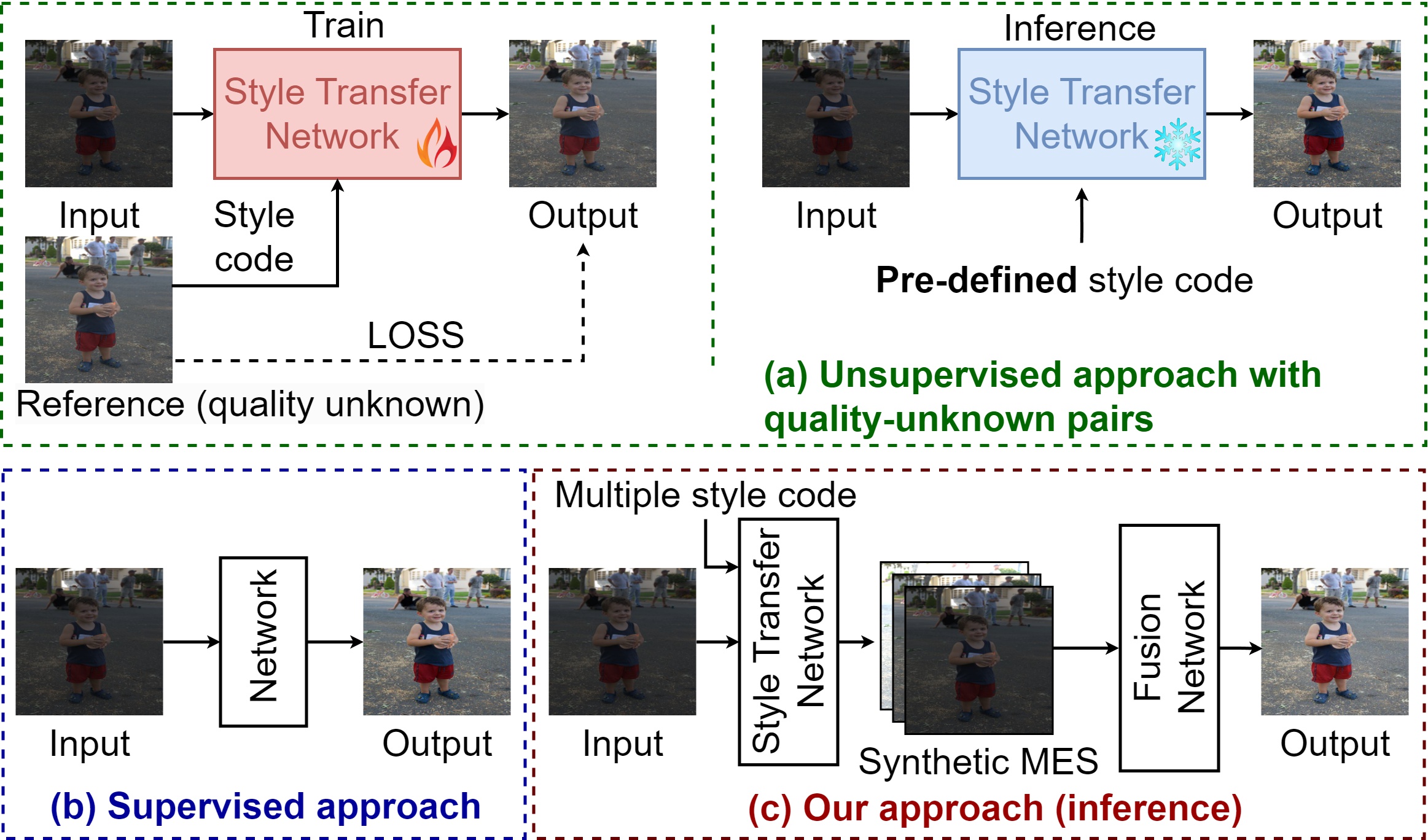}
    \caption{Comparison between previous methods and our proposed universal image contrast enhancement pipeline. \textcolor{red}{Flames} and \textcolor{blue}{snowflakes} refer to learnable and frozen parameters, respectively.}
    \label{fig:pipeline_cmp}
\end{figure}
\section{Introduction}
\label{sec:intro}

\IEEEPARstart{E}xtreme lighting conditions or improper Image Signal Processor (ISP) configurations can significantly degrade image quality in various ways, such as under-/over-exposure or backlit scenes, where uneven illuminance often leads to poor contrast of image intensities. These factors result in a low dynamic range image, either globally (under-/over-exposure) or locally (backlit), damaging the image visual quality and local details. While effective solutions have been developed for each of these challenges, such as low-light image enhancement (LLIE)~\cite{wang2019underexposed, xu2023llie, wu2023llie, cai2023retinexformer, yang2023implicit, fu2023llie, li2023fourier}, exposure correction (EC)~\cite{eyiokur2022exposure, afifi2021msec, huang2023sample_relationship, wang2022lcdp, huang2022exposure_normalization, wang2023fourllie}, backlit image enhancement (BIE)~\cite{liang2023iterative, ueda2020backlit, akai2023backlit, zhao2021backlit} and L2HT transformation (L2HT)~\cite{eilertsen2017hdr1, liu2020reverse, chen2021Hdrunet, wang2022hdr4, chen2023CEVR, li2024hdr6}, \etc, they often lack robustness and generalization performance across different tasks and datasets, as illustrated in Figs.~\ref{fig:cross_task} and~\ref{fig:cross_dataset}.

These issues of existing image contrast enhancement methods primarily stem from limited training data, typically constructed by collecting input and ground-truth (GT) pairs under manually controlled adverse and ideal conditions. Table~\ref{tab:dataset} summarizes the main existing datasets, highlighting their limited number of images and scenes. Although manual intervention is a straightforward way to create images with high human-perceptual quality, it is not only expensive but also leads to inconsistent annotation styles~\cite{bychkovsky2011fivek, jie2021PPR10K}. As shown in Figs.~\ref{fig:cross_task} and~\ref{fig:cross_dataset}, models trained on these datasets fail to generalize across different enhancement tasks and even different datasets of the same task.

Some unsupervised image contrast enhancement methods~\cite{fu2023PairedLIE, jiang2024lightendiffusion, ruodai2024uec} have been proposed to learn from image pairs of unknown quality. Unlike conventional supervised approaches (Fig.~\ref{fig:pipeline_cmp}(a)), which rely on input-GT pairs with manual annotated high-quality reference images, these unsupervised methods (Fig.~\ref{fig:pipeline_cmp}(b)) operate on image pairs of unknown quality. Typically, they adopt a two-stage framework: in the first stage, a style transfer network is trained on quality-unknown image pairs to simulate illumination variations under diverse lighting conditions without requiring quality labels. In the second stage, the pre-trained network is frozen, and a small set of manually designed priors (\eg, brightness constraints~\cite{fu2023PairedLIE} or a few high-quality reference images~\cite{ruodai2024uec, jiang2024lightendiffusion}) is used to identify optimal style codes. 
Modern cameras typically capture high-bit-depth images (\eg, 12 or 16-bit), which are then processed by the ISP to produce 8-bit images suitable for display. During this quantization and truncation process, much of the HDR information is lost, complicating subsequent enhancement tasks. In these unsupervised methods, the first stage preserves HDR signal fidelity, thereby facilitating the optimization of perceptual quality in the second stage. Inspired by these works, we extend the idea of using quality-unknown pairs to using image sequences that span a wide range of illumination levels, akin to a Multi-Exposure Sequence (MES). This enables us to employ Multi-Exposure Fusion (MEF) to fuse the sequence for improved image quality. Furthermore, we can construct a large-scale dataset using an emulated ISP under diverse and adverse illumination conditions, enabling a unified model to address multiple enhancement challenges.

\begin{table}[t]
\small
\caption{Summary of existing datasets for image enhancement. 
``\#images'' indicates the number of training images, while 
``\#scenes'' represents the number of unique scenes. For datasets LOLv2Real~\cite{wei2018lol}, UHD-LL~\cite{Li2023uhdll}, LSRW~\cite{hai2023r2rnet} and BAID~\cite{lv2022backlitnet}, exact number of scenes is not provided. We treat images with a CLIP similarity score greater than 0.9 as duplicated images in the same scene.}
\label{tab:dataset}
\centering
\begin{tabular}{c l r r}
\hline
\multicolumn{1}{c}{Task} & Dataset & \multicolumn{1}{c}{\#images} & {\#scenes} \\
\hline
\multirow{3}{*}{LLIE} & LOLv2Real \cite{wei2018lol} & \multicolumn{1}{r}{789} & 263 \\
 & LSRW \cite{hai2023r2rnet} & \multicolumn{1}{r}{5,650} & 4,157 \\
 & UHD-LL \cite{Li2023uhdll} & \multicolumn{1}{r}{2,150} & 724 \\
\hline
\multirow{3}{*}{EC} & MSEC \cite{afifi2021msec} & \multicolumn{1}{r}{24,330} & 5,000 \\
 & LCDP \cite{wang2022lcdp} & \multicolumn{1}{r}{1,700} & 1,700 \\
 & SICE \cite{cai2018sice} & \multicolumn{1}{r}{4,800} & 589 \\
\hline
\multirow{2}{*}{BIE} & BAID \cite{lv2022backlitnet} & \multicolumn{1}{r}{3,000} & 1,782 \\
 & Backlit300 \cite{liang2023CLIP-LIT} & \multicolumn{1}{r}{300} & 300 \\
\hline
\multirow{2}{*}{L2HT} & HDR-EYE \cite{liu2020reverse} & \multicolumn{1}{r}{46} & 46 \\
 & HDR-REAL \cite{liu2020reverse} & \multicolumn{1}{r}{1,838} & 70 \\
\hline
\end{tabular}
\end{table}

Specifically, we propose to address the challenges of existing image contrast and dynamic range enhancement methods from three aspects: \textbf{task}, \textbf{data}, and \textbf{model}. A common and critical operation in tasks such as LLIE, EC, BIE, L2HT, \etc, is to adjust image exposure and contrast. Therefore, we propose the task of universal contrast enhancement for a generalized solution. To support this task, a large-scale dataset is necessary. Specifically, we need a fully automated pipeline to generate pairs of images before and after enhancement. For contrast enhancement tasks, the input is usually an 8-bit sRGB image for wide applicability, while modern cameras can produce high-bit high dynamic range (HDR) raw images, which can be converted to 8-bit sRGB images. Specifically, we employ an emulated ISP system \cite{afifi2021msec} to adjust the exposure values (EVs) for an HDR raw image, resulting in an MES of 8-bit sRGB images. Then, by taking the MES as input, we render a high-quality contrast enhanced 8-bit sRGB image by MEF~\cite{mertens2009pixelmef, bruce2014pixelmef, ulucan2021pixelmef, kinoshita2019pixelmef, ma2019mefnet, xu2020mefgan, xu2020u2fusion, ram2017deepfuse, jiang2023meflut}, and take it as the pseudo-GT of the sRGB image before contrast enhancement. Finally, a training pair can be built by taking one sRGB image in the MES as input, and the pseudo-GT as output. This fully automated pipeline is highly scalable, enables us to collect 48,361 HDR raw images from publicly available sources~\cite{li2024AODRaw, bychkovsky2011fivek, hasinoff2016HDRP, jie2021PPR10K, omid2014pascalraw, dang2015raise, morawski2022RAWNOD}, and render 338,527 sRGB images to construct the MES and MEF data.

As for the model, an intuitive idea is to train a mapping network between sRGB images before and after enhancement. However, this approach is ineffective due to the inherent task gaps in LLIE, EC, BIE, and L2HT, and it neglects signal fidelity. 
To recover the full dynamic range, several L2HT methods~\cite{eilertsen2017hdr1, liu2020reverse, wang2022hdr4, chen2023CEVR} have been proposed, which first predict a high-bit image from an 8-bit image and then apply tone mapping. However, synthesizing a high-bit image from an 8-bit input is highly challenging as it requires predicting ISP-related properties like the camera response function or an image's specific EV. Since different ISPs have varying properties L2HT methods are mostly limited to certain ISPs. Our objective is to develop a generalized method applicable to various contrast enhancement tasks and images generated by different ISPs. Therefore, the scope of signal fidelity should exclude ISP-related properties. Specifically, we define signal fidelity as the MES data, which do not require accurate ISP-related EV labels. Using our generated MES data, we first train a style-transfer-like network, termed MES-Net, to generate an under-/normal-/over-exposed image triplet from a single sRGB image, thereby synthesizing a realistic MES. Subsequently, using the pseudo GTs generated by MEF methods, we train another network, called MEF-Net, to fuse the MES images synthesized by MES-Net into a high-quality 8-bit image. In our implementation, both MES-Net and MEF-Net are fine-tuned from the pre-trained SD-Turbo~\cite{sauer2023SDturbo} model with LoRA~\cite{hu2022lora}. Notably, they are trained to perform inference in just one diffusion step, demonstrating strong generalization performance with good efficiency. 


In summary, our proposed method, namely \textbf{UN}iversal \textbf{I}mage \textbf{C}ontrast \textbf{E}nhancer (\textbf{UNICE}), is free of human-labeled training pairs. UNICE demonstrates significantly stronger generalization performance across and within different contrast enhancement tasks, including LLIE, EC, BIE and L2HT, as shown in Fig.~\ref{fig:our_results}. The enhanced images can even exceed manually labeled GTs in multiple no-reference image quality assessment (NR-IQA) metrics such as NIQE~\cite{NIQE}, PI~\cite{PI} and ARNIQA~\cite{agnolucci2024arniqa}.

\begin{figure*}[ht]
    \centering
    \setlength{\abovecaptionskip}{0.2mm} 
    \includegraphics[width=1\linewidth]{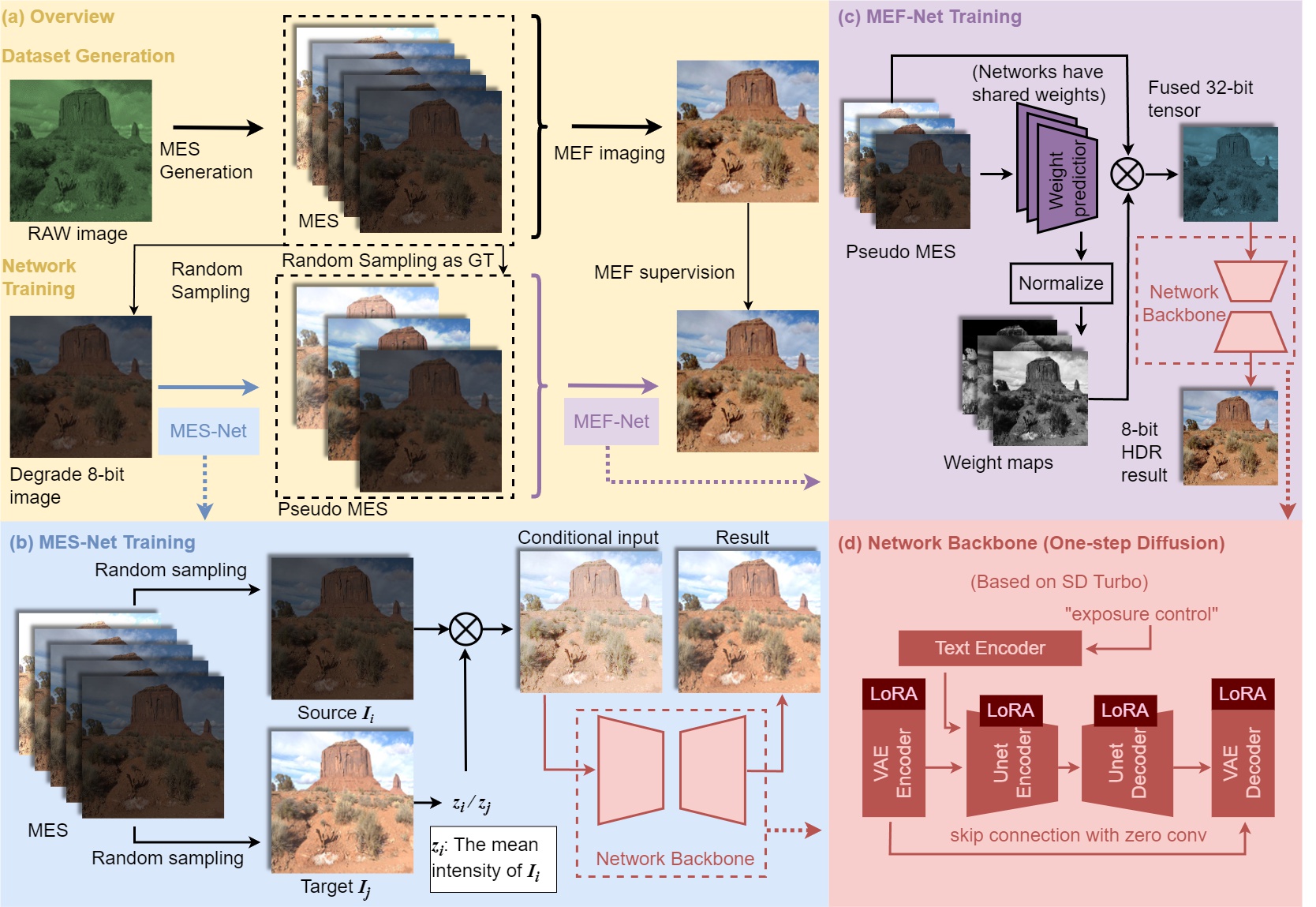}
    \vspace{+1mm}
    \caption{(a) The learning framework of UNICE. (b) Illustration of the training process of MES-Net. (c) Illustration of the training process of MEF-Net. (d) The architecture of the one-step diffusion network.}
    \label{fig:methods}
\end{figure*}

\section{Related Work}
\textbf{Contrast Enhancement.}
Contrast enhancement is a fundamental problem in image processing. Early approaches mostly rely on histogram equalization to adjust image contrast. In recent years, numerous learning-based methods have been proposed to address different contrast enhancement problems, including low-light conditions~\cite{chen2018sid,bychkovsky2011fivek,wei2018lol,yang2020lolv2,Li2023uhdll}, under-/over-exposure~\cite{wang2022lcdp,afifi2021msec,ruodai2024uec}, backlit scenarios\cite{lv2022backlitnet,liang2023CLIP-LIT}, and low dynamic range (LDR)~\cite{ignatov2017dslr, liu2020reverse}, \etc. However, these methods are data-hungry, turning the collection of low-quality input images and the corresponding high-quality GTs into a bottleneck. Low-quality images are often captured under poor lighting conditions or with deliberately mis-configured ISP parameters, while high-quality GTs are obtained under ideal lighting conditions or with proper camera ISP settings. In some cases, professional photographers are involved to manually retouch images to create the GT~\cite{bychkovsky2011fivek, jie2021PPR10K}, which is highly expensive.

\textbf{Multi-Exposure Sequence and Fusion.}
Due to the limited dynamic range of sRGB images, a lot of scene details can be lost, making contrast enhancement difficult. To address this challenge, many approaches ~\cite{mertens2009pixelmef,bruce2014pixelmef,ulucan2021pixelmef,kinoshita2019pixelmef, ma2017spdmef,li2020fmmef,huang2018patchmef} have been proposed to capture a multi-exposure sequence (MES) of the same scene using different exposure settings, and then apply multi-exposure fusion (MEF) to merge the MES into a single sRGB image. In static scenes where the MES images can be perfectly aligned, fusion can be achieved through simple weighted average. Some methods compute pixel-wise weights~\cite{mertens2009pixelmef,bruce2014pixelmef,ulucan2021pixelmef,kinoshita2019pixelmef} by using contrast, color saturation, and brightness as criteria. Some methods ~\cite{ma2017spdmef,li2020fmmef,huang2018patchmef} divide the image into patches and calculate weights for each patch. More recent methods leverage deep learning~\cite{ma2019mefnet,xu2020mefgan,xu2020u2fusion,ram2017deepfuse,jiang2023meflut} to optimize the weight through loss functions. While having shown promising results, it is difficult to capture a static MES in real-world scenarios due to object motion and camera shake. 

\textbf{Diffusion Prior based Image Enhancement.}
Recently, diffusion models ~\cite{yi2023Diff-retinex, fei2023gdp, jiang2023wavelet-diff, wang2023exposurediffusion, feng2024difflight} have achieved exceptional performance in image enhancement, largely due to the pre-training on vast datasets. However, these methods face computational challenges due to the multiple diffusion steps. To address this issue, methods like ExposureDiffusion~\cite{wang2023exposurediffusion} start denoising directly from noisy low-light images instead of random noise, improving convergence and restoration quality. Wavelet-based diffusion models~\cite{jiang2023wavelet-diff} reduce parameters for faster processing. DiffLight~\cite{feng2024difflight} uses a dual-branch structure to minimize inference time, especially for ultra-high definition images. Nonetheless, these models still require multiple steps to finish the enhancement process. In this work, we employ the pre-trained SD-Turbo~\cite{sauer2023SDturbo} as the backbone, and fine-tune it to build the MES-Net and MEF-Net in our UNICE model. Note that our model performs diffusion in just one single step, greatly improving the efficiency during inference.

\section{A Universal Image Contrast Enhancer}

We present in detail our proposed \textbf{UN}iversal \textbf{I}mage \textbf{C}ontrast \textbf{E}nhancer (\textbf{UNICE}). We first overview the learning framework in Sec.~\ref{sec:overall}, which includes a data generation stage and a network training stage. The data generation method is detailed in Sec.~\ref{sec:dataset}, and the network training details are discussed in Sec.~\ref{sec:stage1} and Sec.~\ref{sec:stage2}, respectively.

\subsection{Overview of UNICE}
\label{sec:overall}

The learning framework of UNICE is illustrated in Fig.~\ref{fig:methods}(a), which consists of two stages: a data generation stage (top part) and a network training stage (bottom part). In the data generation stage, we first collect a set of HDR raw images (see Sec. \ref{sec:dataset} for details). For a raw image, denoted by $\bm{I}^{raw}_i$, we employ an emulated ISP system~\cite{afifi2021msec, ruodai2024uec} to render it into 8-bit sRGB images with different exposure values (EVs), forming an MES \(\mathcal{I}_i = \{\bm{I}^i_1, \bm{I}^i_2, \bm{I}^i_3, \ldots\}\), where \(\bm{I}^i_j\) denotes an sRGB image with a specific EV $j$. Then we apply MEF to $\mathcal{I}_i$ to generate a high-quality sRGB pseudo-GT, denoted by \(\bm{I}_i^{GT}\). This fully automated process is highly scalable, enabling us to create a rich number of MES \(\{\mathcal{I}_1, \mathcal{I}_2, \mathcal{I}_3, \ldots\}\) and their corresponding pseudo-GTs. 

The data generated in the first stage are used to train two networks in the second stage. By randomly sampling sRGB images from $\mathcal{I}_i$ as the input and output, we can train a network, called MES-Net, which transfers an sRGB image with EV $j$ to another sRGB image with EV $k$. Then for any given sRGB image \(\bm{I}^i_j\), we can use MES-Net to generate multiple sRGB images with different EVs, synthesizing an MES $\mathcal{I}^i_j$ associated with \(\bm{I}^i_j\). Note that we have already generated a pseudo-GT \(\bm{I}_i^{GT}\) for the raw image $\bm{I}^{raw}_i$. Therefore, we can easily train another network, called MEF-Net, which maps $\mathcal{I}^i_j$ to \(\bm{I}_i^{GT}\). 

Finally, the trained MES-Net and MEF-Net build our UNICE model. During inference, for any given sRGB image to be processed, we pass it through MES-Net and MEF-Net, outputting an enhanced image.

\subsection{Dataset Generation}
\label{sec:dataset}

\begin{table}[t]
\small
\centering
\caption{RAW image datasets used to train UNICE.}
\label{tab:RAW image datasets}
\begin{tabular}{l r l}
\hline
Dataset & \# RAW & Purpose \\
\hline
AODRaw~\cite{li2024AODRaw} & 7,785 & Night Object Detection \\
Fivek~\cite{bychkovsky2011fivek} & 5,000 & Image Retouching \\
HDRP~\cite{hasinoff2016HDRP} & 3,640 & HDR Burst Photography \\
PASCAL-RAW~\cite{omid2014pascalraw} & 4,259 & Object Detection \\
PPR10K~\cite{jie2021PPR10K} & 11,161 & Portrait Photo Retouching  \\
RAISE~\cite{dang2015raise} & 8,155 & Digital Image Forensics \\
RAW-NOD~\cite{morawski2022RAWNOD} & 8,361 & Night Object Detection \\
\hline
TOTAL & 48,361 & \\
\hline
\end{tabular}
\end{table}

Our method does not require any manually labeled GT,  only needing HDR raw images to automatically generate the training data. We collect raw images from the publicly available datasets, including AODRaw~\cite{li2024AODRaw}, FiveK~\cite{bychkovsky2011fivek}, HDRP~\cite{hasinoff2016HDRP}, PPR10K~\cite{jie2021PPR10K}, PASCAL-RAW~\cite{omid2014pascalraw}, RAISE~\cite{dang2015raise}, and RAW-NOD~\cite{morawski2022RAWNOD}, whose  statistics are  summarized in Tab.~\ref{tab:RAW image datasets}.  Among them, FiveK, HDRP, RAISE and PPR10K are image enhancement datasets, while AODRaw, PASCAL-RAW and RAW-NOD are built for object detection from raw images. \textit{It should be noted that although some of these datasets provide manually enhanced images as GT, \textbf{we only use their raw data in the development of UNICE}}.

\subsubsection{Exposure Manipulation} 
\begin{figure*}[ht]
    \centering
    \makebox[0.135\linewidth]{-3EV}
    \makebox[0.135\linewidth]{-2EV} 
    \makebox[0.135\linewidth]{-1EV}
    \makebox[0.135\linewidth]{0EV}
    \makebox[0.135\linewidth]{+1EV} 
    \makebox[0.135\linewidth]{+2EV}
    \makebox[0.135\linewidth]{+3EV} \\
    \vspace{2mm}
    \includegraphics[width=1\linewidth]{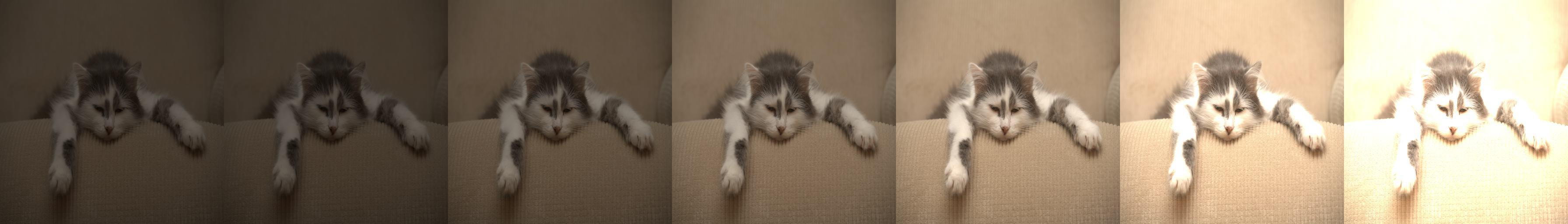}
    \includegraphics[width=1\linewidth]{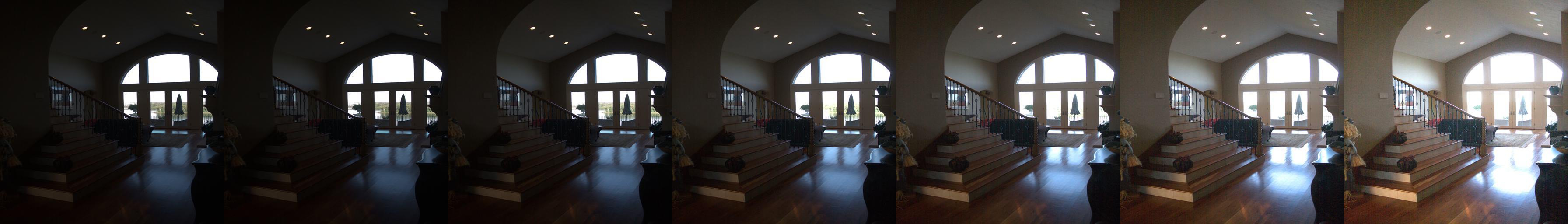}
    \caption{Examples of synthesized MES in our dataset.}
    \label{fig:Exposure Manipulation}
\end{figure*}

\begin{figure}
    \centering
    \includegraphics[width=1\linewidth, height=0.75\linewidth]{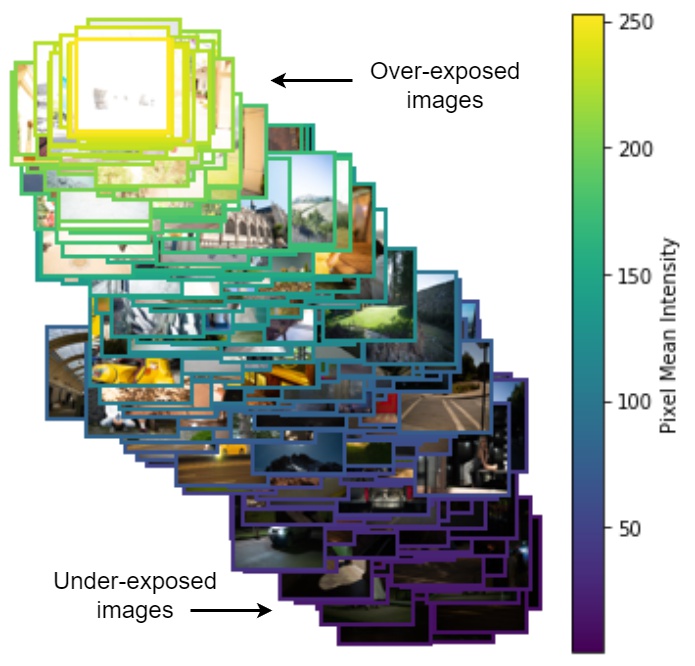}
    \caption{We randomly select 400 images from our dataset and visualize them using t-SNE \cite{van2008t-SNE}. Box colors indicate mean intensity. We see that our dataset covers a wide range of exposure levels. }
    \label{fig:tsne}
\end{figure}

We render sRGB images from HDR raw data. Previous works~\cite{afifi2021msec, ruodai2024uec} have demonstrated that camera ISP rendering can be emulated using metadata from DNG raw files. Following~\cite{afifi2021msec}, we use the Adobe Camera Raw SDK as the emulated ISP, which simulates images with various EVs while keeping other parameters consistent with the original metadata. Setting EVs to $\{+3, +2, +1, 0, -1, -2, -3\}$ proves effective for most raw images, as sensors typically cannot capture valid data beyond this range. From 48,361 raw images, we render 338,527 sRGB images and synthesize 48,361 MES.  Fig.~\ref{fig:Exposure Manipulation} visualizes two MES examples. The first row illustrates images ranging from low-light to over-exposure as EV increases. The second row presents an MES with imbalanced illumination, representing backlit scenarios. We randomly sample 400 rendered images for visualization in Fig.~\ref{fig:tsne}, showing that these images span the full 0-255 intensity spectrum and cover diverse scenes, lighting conditions, and camera sensors.

\begin{figure*}
    \centering
    \includegraphics[width=1\linewidth]{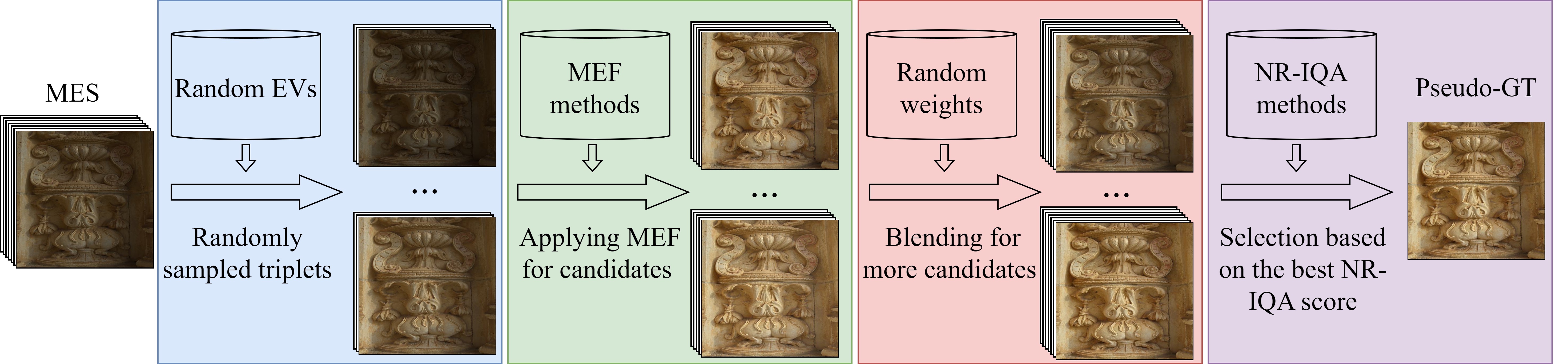}
    \caption{Our pseudo-GT generation and selection pipeline.}
    \label{fig:pseudo-GT generation}
\end{figure*}

\begin{figure*}
    \centering
    \includegraphics[width=1\linewidth]{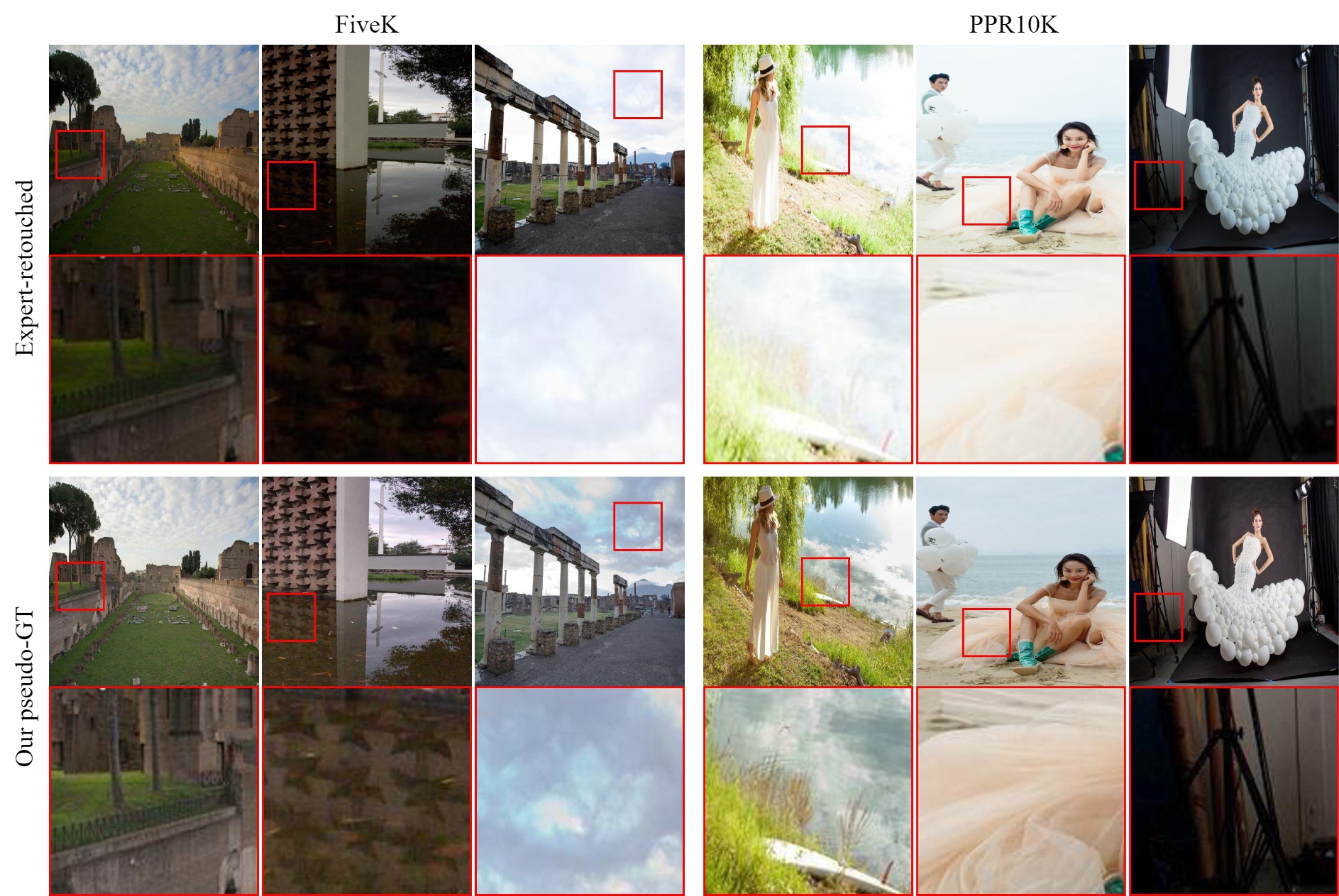}
    \caption{Comparison between the manually annotated GTs (top row) and our synthesized pseudo-GTs (bottom row). The images are from the FiveK (left) and PPR10K (right) datasets. Our synthesized pseudo GTs show more visual details by reducing under-/over-exposed areas, which are marked by the red boxes.}
    \label{fig:GTcmp}
\end{figure*}

\subsubsection{Pseudo-GT Generation} We explore various MEF techniques to generate high-quality pseudo-GTs. While methods such as FMMEF~\cite{li2020fmmef}, GradientMEF~\cite{Li2018GradientMEF}, MDO~\cite{ulucan178MDO}, Mertens~\cite{mertens2009pixelmef}, and PerceptualMEF~\cite{liu2022perceptual} perform well under specific conditions, they lack consistent performance across diverse scenarios. To leverage their respective strengths, we adopt an ensemble approach. Specifically, we select various fusion methods and apply them to randomly selected image triplets with different EVs to generate a large pool of candidate images. These candidates are then randomly blended to increase diversity. We evaluate the results using no-reference image quality assessment (NR-IQA) metrics, as illustrated in Fig.~\ref{fig:pseudo-GT generation}. To manage the large number of generated images, we apply filtering rules to ensure that each triplet includes images with \(EV > 0\), \(EV = 0\), and \(EV < 0\). The blending process is weighted based on NR-IQA scores. For instance, we select a fixed number of multi-exposure sets (\eg, 1000 groups) and apply five fusion methods to each. The results are ranked using various NR-IQA metrics, and we count how often each method ranks first. For example, if FMMEF ranks first in 600 out of 1000 cases, it receives a score of 600. For a given scene and EV combination, we randomly select three out of the five fusion results and combine them using weights proportional to their scores, normalized to sum to 1. We then use NR-IQA methods—including NIQE~\cite{NIQE}, BRISQUE~\cite{mittal2012brisque}, PI~\cite{PI}, and ARNIQA~\cite{agnolucci2024arniqa}—to rank both the original five fused images and the 10 weighted combinations, selecting the highest-quality result based on average scores. 

\subsubsection{Quality Control} Despite using an ensemble MEF approach to generate pseudo-GT, extreme adverse illumination or non-illumination factors (\eg, noise and motion blur) can occasionally compromise quality. To address this, we implement a quality threshold using ARNIQA, the state-of-the-art NR-IQA method. We exclude images with ARNIQA scores below 0.5, as they indicate poor quality. This results in the exclusion of 1,433 images, leaving 46,928 RAW images. Ultimately, we obtain 328,496 input-GT pairs. 

\subsection{MES-Net Training}
\label{sec:stage1}

According to the exposure physics in imaging process~\cite{liu2020reverse}, an LDR image $\bm{I}^i$ can be rendered from the HDR scene irradiance $\bm{H}^i$ with a simplified ISP system as follows:
\begin{equation}
    \bm{I}^i = \text{ISP}(\bm{H}^i), \quad \text{where} \quad \bm{H}^i = \bm{S}^i \Delta t,
\end{equation}
where $\bm{S}^i$ is the sensor response and $\Delta t$ is the exposure time. L2HT methods~\cite{liu2020reverse, chen2021Hdrunet, wang2022hdr4, chen2023CEVR, li2024hdr6} often try to reverse the ISP to estimate $\bm{H}^i$ from $\bm{I}^i$, which is very challenging. 
In this work, we synthesize an MES $\mathcal{I}^i$, which can acquire the HDR information of the scene, from a single sRGB image $\bm{I}^i_j$. Different from those L2HT methods, we formulate this process as a style transfer problem, and train an MES-Net to transform an sRGB image $\bm{I}^i_{j} = \text{ISP}(\bm{S}^i \Delta t_{j})$ to another sRGB image $\bm{I}^i_{k} = \text{ISP}(\bm{S}^i \Delta t_{k})$. The style code $z$ is extracted from $\bm{I}^i_{j}$ and $\bm{I}^i_{k}$ in training for exposure adjustment, which can be manipulated during testing. 
Therefore, by applying different style codes, we can generate multiple $\bm{I}^i_{k}$ for $\bm{I}^i$, forming a pseudo MES $\mathcal{I}^i$ of it. 

The training of MES-Net is illustrated in Fig.~\ref{fig:methods}(b). 
We leverage diffusion priors to enhance the network generalization performance by taking the pretrained SD-Turbo~\cite{parmar2024pix2pix_turbo} as backbone. 
During training, each time we randomly sample two images $\bm{I}^i_{j}$ and $\bm{I}^i_{k}$ from the MES synthesized in our data generation stage, taking $\bm{I}^i_{j}$ as the network input and $\bm{I}^i_{k}$ as the desired output. We define the style code $z$ as the mean intensity of the sRGB image, which corresponds well to the image exposure level. The MES-Net is trained to establish the following one-to-one mapping:
\begin{equation}
\label{eq:exposure control}
    \textbf{MES-Net}\left( (z_k / z_j) \cdot \bm{I}^i_{j} \right) \rightarrow \bm{I}^i_{k}.
\end{equation}
The simple $L_2$ loss is used to train our MES-Net.
Once trained, during testing, for any input $\bm{I}^i_{j}$ with style code $z_j$, the code $z_k$ is used to adjust the exposure to generate images with varying brightness levels. With both visual quality and runtime efficiency in consideration, we uniformly sample three values from $z_k$, specifically $z_k = 0.25$, $0.5$, and $0.75$, to generate a sequence of images.

As illustrated in Fig.~\ref{fig:methods}(d), we employ SD-Turbo to enable MES-Net to perform one-step image synthesis. To enhance flexibility, we incorporate LoRA adapters~\cite{hu2022lora} into the zero-convolution modules. Specifically, selected intermediate outputs from SD-Turbo are processed through zero-convolution layers and added as residuals to subsequent layers. This design introduces only a small number of trainable, task-specific parameters. During training, the parameters of SD-Turbo remain frozen, while only the zero-convolution layers are updated.


\subsection{MEF-Net Training}
\label{sec:stage2}

With the pseudo MES $\mathcal{I}^i$ generated by MES-Net as described in Sec.~\ref{sec:stage1}, we train an MEF-Net to convert $\mathcal{I}^i$ into a high-quality sRGB image \(\hat{\bm{I}}_i\), using the synthesized pseudo-GT \(\bm{I}^{GT}_i\) from the data generation stage as the supervision target. The training process of MEF-Net is illustrated in Fig.~\ref{fig:methods}(c), which can be formulated as:
\begin{equation}
\textbf{MEF-Net}\left(\sum_i W(\mathcal{I}^i) \circ \mathcal{I}^i\right) \rightarrow {\bm{I}}^{GT},
\end{equation}
where \(\circ\) denotes pixel-wise multiplication, and \(W(\cdot)\) is a network that combines the input sequence $\mathcal{I}^i$ into an implicit 32-bit HDR representation.

We implement \(W(\cdot)\) as a fully convolutional network \(g(\cdot)\), using a VGG encoder as the backbone. The final fully connected layers are removed, and the output channel is modified to 1. Each image \(\bm{I}^i_j\) in $\mathcal{I}^i$ is processed by \(W(\cdot)\) to generate a corresponding weight map \(\bm{w}_j\). These weight maps are normalized as follows:
\begin{equation}
\bm{w}'_j(x, y) = \frac{\bm{w}_j(x, y)}{\sum_{k=1}^{N} \bm{w}_k(x, y) + \epsilon},
\end{equation}
where \(\epsilon\) is a small constant to avoid division by zero, and \(N\) is the number of images in $\mathcal{I}^i$ (set to 3 in our experiments). The pixel-wise multiplication of \(W(\cdot)\) and $\mathcal{I}^i$ yields a 32-bit implicit HDR image. MEF-Net, which adopts the same one-step diffusion architecture as MES-Net, is then trained to produce \(\hat{\bm{I}}_i\) that closely matches \(\bm{I}^{GT}_i\). We jointly optimize \(g(W(\cdot))\) and MEF-Net using an $L_2$ loss.

\begin{table*}[]
\small
\caption{Quantitative comparison between manually adjusted GTs and our UNICE model outputs using NR-IQA metrics. ``↓(↑)'' indicates lower(higher) score is better. The best results are in \textbf{bold}.}
\label{tab:cmp_gt}
\centering
\begin{tabular}{l|cc|cc|cc|cc}
\hline
Dataset & \multicolumn{2}{c|}{UHD~\cite{Li2023uhdll} (LLIE)} & \multicolumn{2}{c|}{LSRW~\cite{hai2023r2rnet} (LLIE)} & \multicolumn{2}{c|}{LOLv2-Real~\cite{yang2020lolv2} (LLIE)} & \multicolumn{2}{c}{SICE~\cite{cai2018sice} (EC)} \\
\hline
Method & GT & Ours & GT & Ours & GT & Ours & GT & Ours \\
\hline
NIQE~\cite{NIQE}↓ & 6.738 & \textbf{5.684} & 6.165 & \textbf{5.813} & 5.559 & \textbf{4.990} & 8.050 & \textbf{5.327} \\
PI~\cite{PI}↓ & 5.013 & \textbf{4.100} & 4.473 & \textbf{4.137} & 4.171 & \textbf{3.285} & 5.110 & \textbf{3.429} \\
BRISQUE~\cite{mittal2012brisque}↓ & 26.62 & \textbf{21.058} & 21.951 & \textbf{16.909} & \textbf{18.492} & 21.22 & 32.065 & \textbf{11.914} \\
ARNIQA~\cite{agnolucci2024arniqa}↑ & 0.714 & \textbf{0.773} & 0.709 & \textbf{0.741} & 0.648 & \textbf{0.731} & 0.666 & \textbf{0.695} \\
\hline
Dataset & \multicolumn{2}{c|}{BAID~\cite{lv2022backlitnet} (BIE)} & \multicolumn{2}{c|}{HDREYE~\cite{liu2020reverse} (L2HT)} & \multicolumn{2}{c|}{HDR-REAL~\cite{liu2020reverse} (L2HT)} & \multicolumn{2}{c}{MSEC~\cite{afifi2021msec} (EC)} \\
\hline
Method & GT & Ours & GT & Ours & GT & Ours & GT & Ours \\
\hline
NIQE~\cite{NIQE}↓ & \textbf{5.309} & 5.314 & \textbf{5.506} & 6.012 & \textbf{5.117} & 5.604 & \textbf{5.444} & 5.457 \\
PI~\cite{PI}↓ & 3.538 & \textbf{3.523} & \textbf{3.498} & 3.826 & \textbf{3.296} & 3.762 & 3.697 & \textbf{3.612} \\
BRISQUE~\cite{mittal2012brisque}↓ & 22.069 & \textbf{20.664} & 18.299 & \textbf{14.845} & \textbf{21.022} & 22.808 & 19.856 & \textbf{19.122} \\
ARNIQA~\cite{agnolucci2024arniqa}↑ & 0.761 & \textbf{0.817} & 0.625 & \textbf{0.794} & \textbf{0.784} & 0.754 & 0.749 & \textbf{0.807} \\
\hline
\end{tabular}
\end{table*}

\begin{figure*}[t]
    \centering
    \includegraphics[width=1\linewidth]{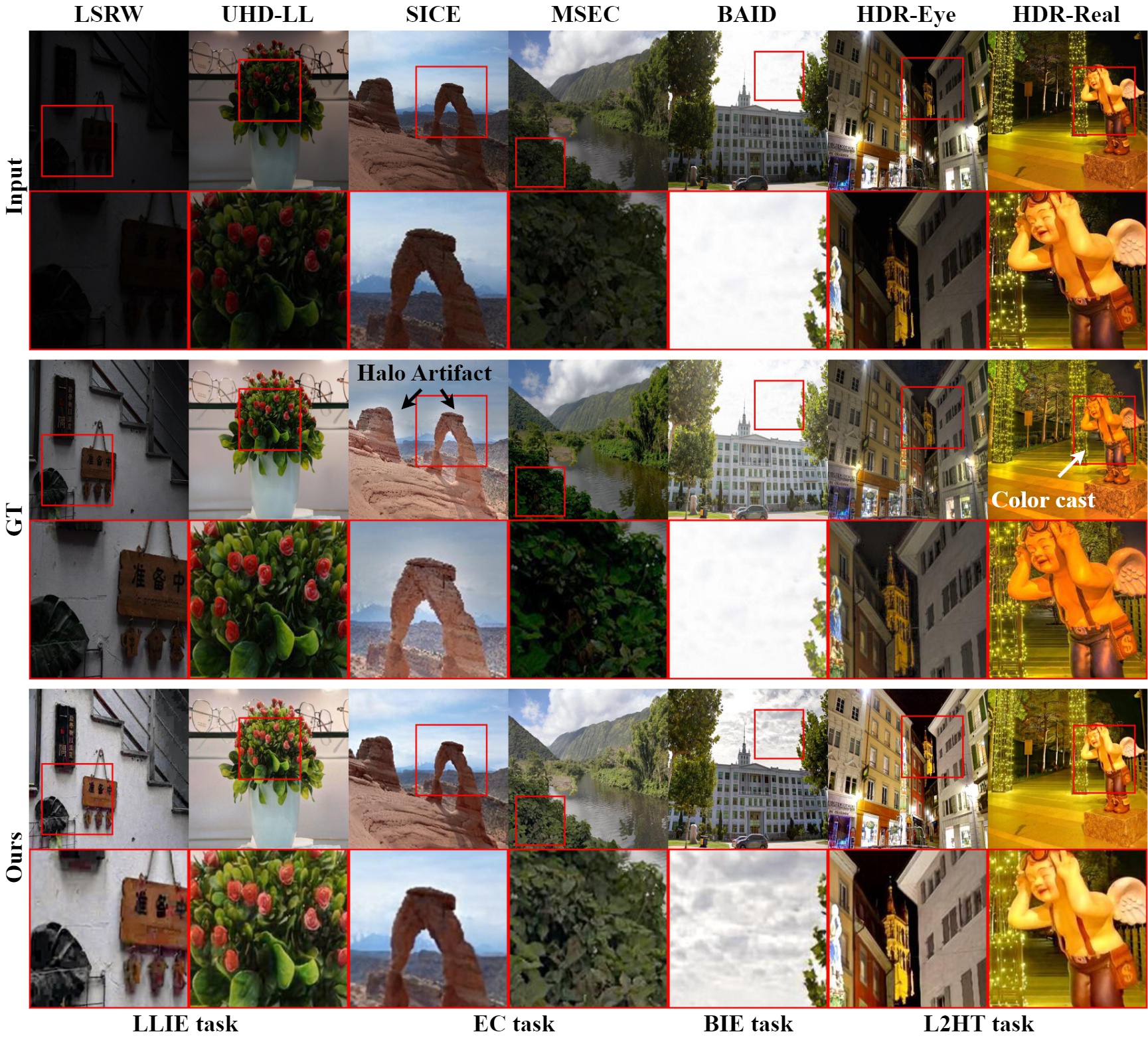}
    \caption{Visual comparison with GT on several datasets. The region marked with red box indicates locally under-/over-exposed with detail loss in the GT, which  may not rival the enhanced images by our UNICE model.}
    \label{fig:GT_unice_cmp}
\end{figure*}

\begin{table*}[h]
\centering
\caption{Task-specific datasets and methods used in our experiments. The training-free or unsupervised methods are marked with notion ``$^{*}$''. }
\begin{tabular}{ccc}
\toprule
\textbf{Task} & \textbf{Dataset} & \textbf{Method} \\
\midrule
LLIE & UHD-LL~\cite{Li2023uhdll}, LSRW~\cite{hai2023r2rnet} & CoLIE$^{*}$~\cite{chobola2024colie}, SCI$^{*}$~\cite{ma2022sci}, RF~\cite{cai2023retinexformer}, FLLIE~\cite{wang2023fourllie}, SD-T~\cite{parmar2024pix2pix_turbo} \\
EC & SICE~\cite{cai2018sice}, MSEC~\cite{afifi2021msec} & CoLIE$^{*}$~\cite{chobola2024colie}, LCDP~\cite{wang2022lcdp}, RF~\cite{cai2023retinexformer}, CSEC~\cite{li_2024_cvpr_csec}, SD-T~\cite{parmar2024pix2pix_turbo} \\
BIE & BAID~\cite{akai2023backlit}, Backlit300~\cite{liang2023CLIP-LIT} & CoLIE$^{*}$~\cite{chobola2024colie}, RF~\cite{cai2023retinexformer}, CSEC~\cite{li_2024_cvpr_csec}, CLIP-LIT$^{*}$~\cite{liang2023CLIP-LIT}, ZDCE$^{*}$~\cite{Zero-DCE++}, SCI$^{*}$~\cite{ma2022sci} \\
L2HT & HDREye~\cite{liu2020reverse}, HDRReal~\cite{liu2020reverse} & SHDR~\cite{liu2020reverse}, CEVR~\cite{chen2023CEVR}, CoLIE$^{*}$~\cite{chobola2024colie}, LCDP~\cite{wang2022lcdp}, RF~\cite{cai2023retinexformer}, CSEC~\cite{li_2024_cvpr_csec} \\
\bottomrule
\end{tabular}
\label{tab:datasets_methods}
\end{table*}

\section{Experiments}

We train our UNICE model using a single NVIDIA TESLA A100-PCIE-40GB GPU. The training configurations of MSE-Net and MFE-Net closely follow img2img-turbo~\cite{parmar2024pix2pix_turbo}. The batch size is set to 2. The Adam optimizer is employed with an initial learning rate of $5\times10^{-6}$, hyperparameters $\beta_1 = 0.9$, $\beta_2 = 0.999$, and a weight decay of 0.01. For the LoRA module, the rank is set to 8 and 4 for the U-Net and VAE parts of the SD-Turbo model, respectively. Considering the high memory consumption of diffusion models, we train MSE-Net and MFE-Net separately.

\subsection{Experiment Setting}

\textbf{Test Datasets}. For each task, we employ the widely-used and relatively large datasets for test. Specifically, for LLIE, we utilize LSRW~\cite{hai2023r2rnet} and UHD-LL~\cite{Li2023uhdll}. For EC, we employ  MSEC~\cite{afifi2021msec} and SICE~\cite{cai2018sice}. The LCDP~\cite{wang2022lcdp} dataset is not used because of its substantial data overlap with MSEC~\cite{afifi2021msec}. For BIE, we use BAID~\cite{lv2022backlitnet} as it is the only publicly available paired dataset. For L2HT, we employ HDR-EYE~\cite{liu2020reverse} and HDR-Real~\cite{liu2020reverse}. 

\textbf{Compared Methods}.
As we are the first to propose a universal contrast enhancer, we can only compare our UNICE method with representative methods that solve specific tasks, such as LLIE, EC, BIE, L2HT, including CoLIE~\cite{chobola2024colie}, SCI~\cite{ma2022sci}, FLLIE~\cite{fu2023llie}, RF~\cite{cai2023retinexformer}, LCDPNet~\cite{wang2022lcdp}, CSEC~\cite{li_2024_cvpr_csec}, CLIP-LIT~\cite{liang2023CLIP-LIT}, \etc. Furthermore, since we use SD-Turbo \cite{parmar2024pix2pix_turbo} as the backbone in UNICE, to verify that the effectiveness of UNICE does not come from SD-Turbo alone, we further implement a compared method using our SD-Turbo-based MES-Net to learn the mapping between input-GT image pairs directly without the MES generation and fusion process. We abbreviate this method as SD-T in the experiments.

\textbf{Evaluation Metrics}. We evaluate the competing methods using both full-reference image quality assessment (FR-IQA) and no-reference image quality assessment (NR-IQA) metrics. The FR-IQA metrics include PSNR, SSIM~\cite{wang2004ssim}, LPIPS~\cite{zhang2018lpips} and DISTS~\cite{ding2020dists}. For NR-IQA metrics, we employ NIQE~\cite{NIQE}, PI~\cite{PI}, BRISQUE~\cite{mittal2012brisque} and ARNIQA~\cite{agnolucci2024arniqa}.

\textbf{Experiments}. To comprehensively evaluate the proposed UNICE model and demonstrate its advantages over existing task-specific contrast enhancement methods, we perform a series of experiments.
\textbf{(1) Comparison with GT}. In this experiment, we compare the outputs of UNICE with the GT images across all four tasks using NR-IQA metrics.
\textbf{(2) Generalization Performance}. UNICE is a generalized image contrast enhancer. In this experiment, we select several SOTA methods for each task and compare UNICE with them on cross-task and cross-dataset generalization performance.
\textbf{(3) User Study}. In addition to objective metrics, we conduct a double-blinded user study to evaluate the subjective quality of the enhanced images. 
\textbf{(4) Complexity Analysis.} We analyze the computational efficiency of our model by evaluating its computational cost, including metrics such as FPS, FLOPS, \etc.
\textbf{(5) Ablation Study.} Finally, a series of ablation studies are conducted to validate the impact of different components of UNICE.


\subsection{Comparison with GT}
\label{sec:Comparison with GT}
This experiment aims to demonstrate that UNICE, despite being trained without human-labeled GT, can deliver high-quality contrast enhancement across various tasks. The quantitative results are presented in Tab.~\ref{tab:cmp_gt} and some visual examples are presented in Fig. \ref{fig:GT_unice_cmp}. (More visual results are provided in the \textbf{supplementary file}.) We can see that in most cases, the NR-IQA scores of UNICE can match or even surpass that of the GTs on these datasets. It is worth mentioning that most of the GT data are captured under optimal lighting conditions, while some GT data are retouched by experienced photographers. Nonetheless, some GTs may not rival the enhanced images by our UNICE model, which is trained on large-scale data. 
As can be seen in Fig. \ref{fig:GT_unice_cmp}, the GTs of LSRW~\cite{hai2023r2rnet}, UHD-LL~\cite{Li2023uhdll}, MSEC~\cite{afifi2021msec}, and HDR-Eye~\cite{liu2020reverse} tend to be dark and lack some details. In contrast, the GTs in BAID~\cite{lv2022backlitnet} tend to be locally over-exposed and also lack some details. Additionally, some of the GTs exhibit artifacts. For example, the GTs of SICE~\cite{cai2018sice} may contain halo artifacts, while the GTs of HDR-Real~\cite{liu2020reverse} may contain color cast artifacts.

\begin{table*}[ht]
    \small
    \caption{Cross-dataset evaluation on LLIE and EC tasks. ``↓(↑)'' indicates that a lower (higher) value is better. The best results are in \textbf{bold}. Note that the Backlit dataset does not have GT, therefore FR-IQA metrics cannot be calculated, denoted by symbol `-'. }
    \label{tab:cross_task}
    \centering
    \begin{tabular}{clccccccccc}
        \hline
        Task & Dataset & Method & PSNR↑ & SSIM↑ & LPIPS↓ & DISTS↓ & NIQE↓ & PI↓ & BRISQUE↓ & ARN.↑ \\
        \hline
        \multirow{8}{*}{\rotatebox{90}{LLIE}} 
            & \multirow{4}{*}{LSRW} 
            & LCDPNet (ECCV'22) & 15.348 & 0.552 & 0.213 & 0.170 & 6.734 & 4.981 & 31.593 & 0.662 \\
            &  & RF (ICCV'23) & 11.816 & 0.416 & 0.301 & 0.214 & 6.693 & 5.350 & 27.289 & 0.716 \\
            &  & CSEC (CVPR'24) & 15.899 & 0.659 & 0.252 & 0.202 & 6.040 & 5.177 & 29.663 & 0.534 \\
            &  & Ours & \textbf{19.399} & \textbf{0.718} & \textbf{0.153} & \textbf{0.129} & \textbf{5.813} & \textbf{4.137} & \textbf{16.909} & \textbf{0.741} \\
            \cline{2-11}
            & \multirow{4}{*}{UHDLL} 
            & LCDPNet (ECCV'22) & 19.148 & 0.835 & 0.114 & 0.119 & 6.093 & 4.429 & 25.532 & 0.712 \\
            &  & RF (ICCV'23) & 16.383 & 0.755 & 0.146 & 0.133 & 6.233 & 4.612 & 22.506 & 0.733 \\
            &  & CSEC (CVPR'24) & 15.250 & 0.820 & 0.167 & 0.170 & 5.836 & 5.191 & 23.996 & 0.488 \\
            &  & Ours & \textbf{22.007} & \textbf{0.907} & \textbf{0.090} & \textbf{0.111} & \textbf{5.684} & \textbf{4.100} & \textbf{21.058} & \textbf{0.773} \\
        \hline
        \multirow{8}{*}{\rotatebox{90}{BIE}} 
            & \multirow{4}{*}{BAID} 
            & LCDPNet (ECCV'22) & 15.254 & 0.753 & 0.129 & 0.122 & 5.690 & 3.881 & 22.732 & 0.757 \\
            &  & RF (ICCV'23) & 15.914 & 0.800 & 0.109 & 0.112 & 5.797 & 3.820 & 21.184 & 0.785 \\
            &  & CSEC (CVPR'24) & 18.476 & 0.870 & 0.119 & 0.130 & 5.075 & 3.972 & 20.578 & 0.594 \\
            &  & Ours & \textbf{19.058} & \textbf{0.876} & \textbf{0.095} & \textbf{0.106} & \textbf{5.178} & \textbf{3.539} & \textbf{20.664} & \textbf{0.817} \\
            \cline{2-11}
            & \multirow{4}{*}{Backlit} 
            & LCDPNet (ECCV'22) & - & - & - & - & 4.886 & 3.435 & 17.781 & 0.728 \\
            &  & RF (ICCV'23) & - & - & - & - & 5.145 & 3.517 & 16.814 & 0.753 \\
            &  & CSEC (CVPR'24) & - & - & - & - & 4.732 & 3.505 & 16.833 & 0.545 \\
            &  & Ours & - & - & - & - & \textbf{4.627} & \textbf{3.123} & \textbf{16.177} & \textbf{0.782} \\
        \hline
        \multirow{8}{*}{\rotatebox{90}{L2HT}} 
            & \multirow{4}{*}{HDRReal} 
            & LCDPNet (ECCV'22) & 14.261 & 0.521 & 0.359 & 0.259 & 8.353 & 5.644 & 25.931 & 0.648 \\
            &  & RF (ICCV'23) & 13.138 & 0.499 & 0.386 & 0.282 & 9.976 & 6.547 & 27.306 & 0.689 \\
            &  & CSEC (CVPR'24) & 14.766 & 0.610 & 0.343 & 0.260 & 7.130 & 5.418 & 23.421 & 0.566 \\
            &  & Ours & \textbf{16.818} & \textbf{0.721} & \textbf{0.265} & \textbf{0.201} & \textbf{5.455} & \textbf{3.615} & \textbf{23.034} & \textbf{0.764} \\
            \cline{2-11}
            & \multirow{4}{*}{HDREye} 
            & LCDPNet (ECCV'22) & 15.497 & 0.643 & 0.247 & 0.162 & 6.079 & 4.063 & 24.472 & 0.721 \\
            &  & RF (ICCV'23) & 16.072 & 0.682 & 0.229 & 0.155 & 6.440 & 4.023 & 24.932 & 0.729 \\
            &  & CSEC (CVPR'24) & 16.001 & 0.636 & 0.234 & 0.155 & 5.289 & 3.757 & 23.533 & 0.660 \\
            &  & Ours & \textbf{16.154} & \textbf{0.687} & \textbf{0.216} & \textbf{0.147} & \textbf{3.481} & \textbf{3.049} & \textbf{23.497} & \textbf{0.732} \\
        \hline
    \end{tabular}
\end{table*}

\begin{figure*}
    \centering
    \includegraphics[width=1\linewidth]{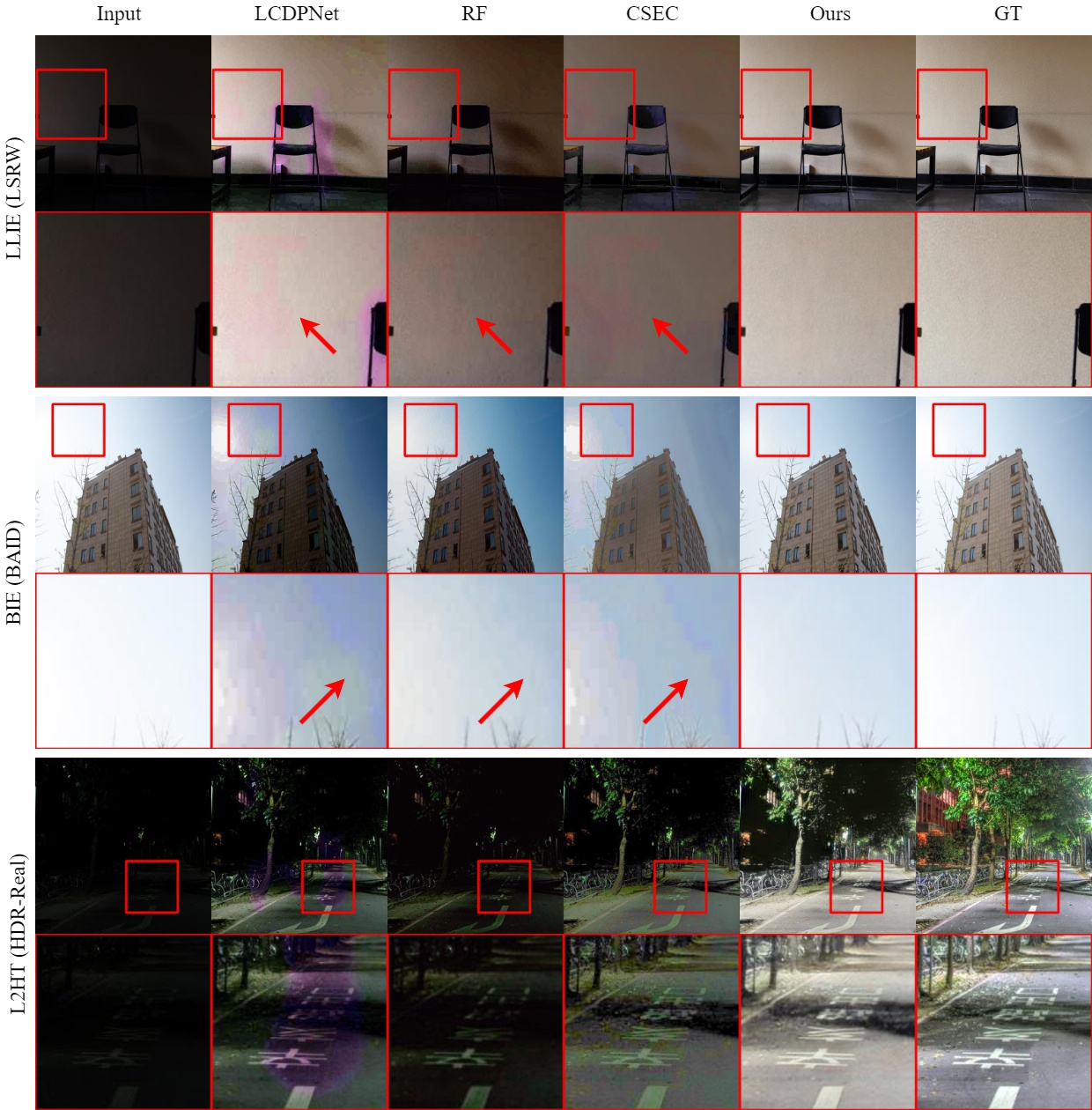}
    \caption{Visual comparisons of cross-task evaluation. The regions marked with red boxes and arrows indicate areas with artifacts introduced by the task-specific models.}
    \label{fig:cross_task_generalization}
\end{figure*}

\subsection{Generalization Performance}
\label{sec:Generalization Performance}

Existing contrast enhancement methods are typically designed for specific tasks and lack generalization across multiple tasks (see Fig.~\ref{fig:cross_task}). Our UNICE aims to provide a universal solution, and we evaluate its generalization performance from two aspects.
(1) The first aspect is cross-task generalization performance. We benchmark UNICE against task-specific models across four contrast-related tasks: LLIE, EC, BIE, and L2HT. (2) The second aspect is cross-dataset generalization performance, where UNICE is compared with state-of-the-art methods for each specific task (see Fig.~\ref{fig:cross_dataset}).
Task-specific datasets and methods are listed in Tab.~\ref{tab:datasets_methods}. To ensure fair comparisons, we adopt different training strategies based on the requirements of individual methods. For supervised methods with paired learning, we train them using input-GT pairs from our dataset. For unpaired learning methods such as CLIP-LIT~\cite{liang2023CLIP-LIT}, we shuffle the ground truth images to create unpaired data. For methods that only require input data, like SCI~\cite{ma2022sci} or ZDCE~\cite{Zero-DCE++}, we use only the input images for training.

\subsubsection{Cross-task Generalization}

Considering that exposure correction has the broadest coverage among these tasks and that the MSEC dataset, specifically designed for this task, contains the largest amount of data, we adopt LCDPNet and CSEC as baselines. Both methods are trained on MSEC. Additionally, we retrain RF~\cite{cai2023retinexformer} on MSEC due to its superior performance in low-light image enhancement (LLIE). Quantitative results are shown in Tab.~\ref{tab:cross_task}. We see that previous task-specific methods suffer from performance drop when applied to other tasks, whereas our model significantly enhances the task-generalization capability, outperforming existing approaches by a large margin.

Fig.~\ref{fig:cross_task_generalization} presents visual comparisons. Due to limited data, task-specific models often introduce artifacts caused by overfitting to small datasets. For instance, in the case of LLIE on LSRW, color distortion and blocking artifacts appear—unrealistic color blocks emerge, and slight color misalignments in the original image become more noticeable after enhancement. In the case of BIE on BAID, the existing methods exhibit color banding, characterized by abrupt transitions between shades in areas with soft gradients. In the case of L2HT on HDR-Real, task-specific models produce a greenish-colored cast. In contrast, our model avoids these issues.

\subsubsection{Cross-dataset Generalization}

We assess cross-dataset generalization by training methods on one dataset from a contrast-related task and testing them on another dataset from the same task. Each competing method is trained on a dataset distinct from both the test sets and our collected data. For methods already trained on the relevant dataset by the original authors, we use their pre-trained models. To ensure fairness, we also retrain all baseline methods on our constructed dataset, except for training-free approaches like CoLIE~\cite{chobola2024colie}. The datasets and methods are summarized in Tab.~\ref{tab:datasets_methods}, with detailed experimental configurations for each task described below.

\noindent \textbf{LLIE task.} Evaluation is conducted on the UHD-LL~\cite{Li2023uhdll} and LSRW~\cite{hai2023r2rnet} datasets. UNICE is compared with CoLIE~\cite{chobola2024colie}, SCI~\cite{ma2022sci}, RF~\cite{cai2023retinexformer}, FLLIE~\cite{wang2023fourllie}, and SD-T~\cite{parmar2024pix2pix_turbo}.

\noindent \textbf{EC task.} Experiments are performed on the SICE~\cite{cai2018sice} and MSEC~\cite{afifi2021msec} datasets. Comparisons are made with CoLIE~\cite{chobola2024colie}, LCDP~\cite{wang2022lcdp}, RF~\cite{cai2023retinexformer}, CSEC~\cite{li_2024_cvpr_csec}, and SD-T~\cite{parmar2024pix2pix_turbo}.

\noindent \textbf{BIE task.} Currently, only two publicly available datasets exist: BAID~\cite{akai2023backlit} (paired with GT) and Backlit300~\cite{liang2023CLIP-LIT} (unpaired without GT). We select four supervised methods for experiments on BAID: CoLIE~\cite{chobola2024colie}, RF~\cite{cai2023retinexformer}, CSEC~\cite{li_2024_cvpr_csec}, and CLIP-LIT~\cite{liang2023CLIP-LIT}. For Backlit300, four unsupervised methods are chosen: CoLIE~\cite{chobola2024colie}, ZDCE~\cite{Zero-DCE++}, SCI~\cite{ma2022sci}, and CLIP-LIT~\cite{liang2023CLIP-LIT}. For unpaired learning on Backlit dataset, we follow CLIP-LIT~\cite{liang2023CLIP-LIT}, using DIV2k~\cite{DIV2k} images as high-quality references. When training on our dataset, input images and pseudo-GTs are shuffled to create unpaired data.

\noindent \textbf{L2HT task.} The HDREye~\cite{liu2020reverse} and HDRReal~\cite{liu2020reverse} datasets are employed. Testing methods include SHDR~\cite{liu2020reverse} and CEVR~\cite{chen2023CEVR}. Images are converted to high-bit-depth and tonemapped using Reinhard's method~\cite{reinhard2005}, following the approach employed by these papers. Since we focus on 8-bit images, 8-bit enhancement methods related to this task are also included for comparison: CoLIE~\cite{chobola2024colie}, LCDP~\cite{wang2022lcdp}, RF~\cite{cai2023retinexformer}, and CSEC~\cite{li_2024_cvpr_csec}.

\begin{figure*}[t]
    \centering
    \makebox[0.13\linewidth]{Input\hspace{3em}}
    \makebox[0.13\linewidth]{GT\hspace{2em}}
    \makebox[0.13\linewidth]{CoLIE~\cite{chobola2024colie}}
    \makebox[0.13\linewidth]{LCDPNet~\cite{wang2022lcdp}}
    \makebox[0.13\linewidth]{RF~\cite{cai2023retinexformer}}
    \makebox[0.13\linewidth]{CSEC~\cite{li_2024_cvpr_csec}}
    \makebox[0.13\linewidth]{\hspace{2em}Ours} \\
    
    \includegraphics[width=1\linewidth]{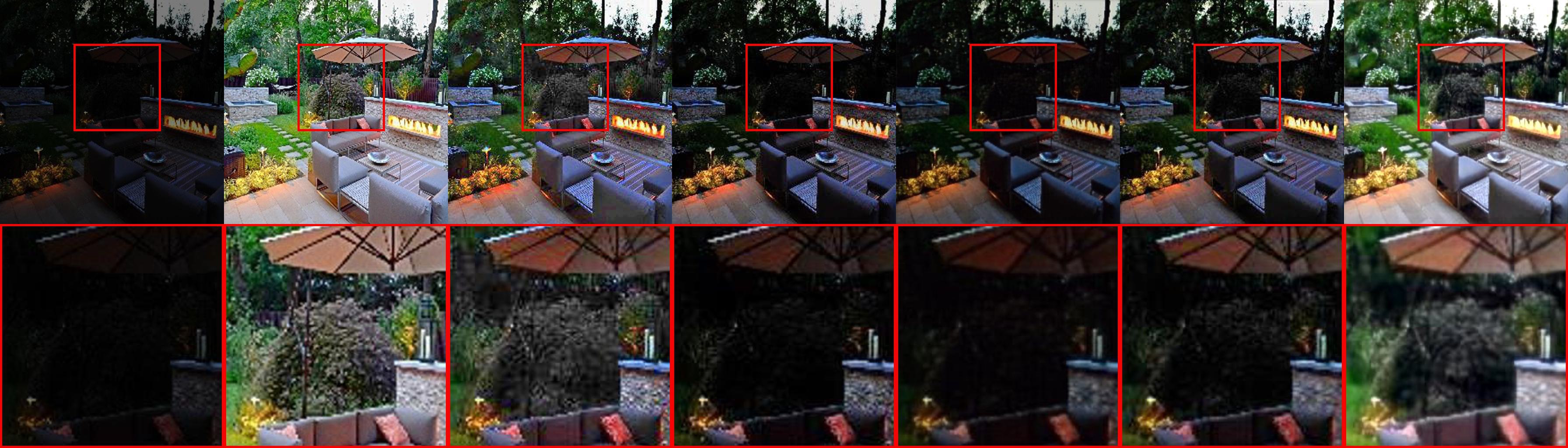} \\
    \vspace{2mm}
    \includegraphics[width=1\linewidth]{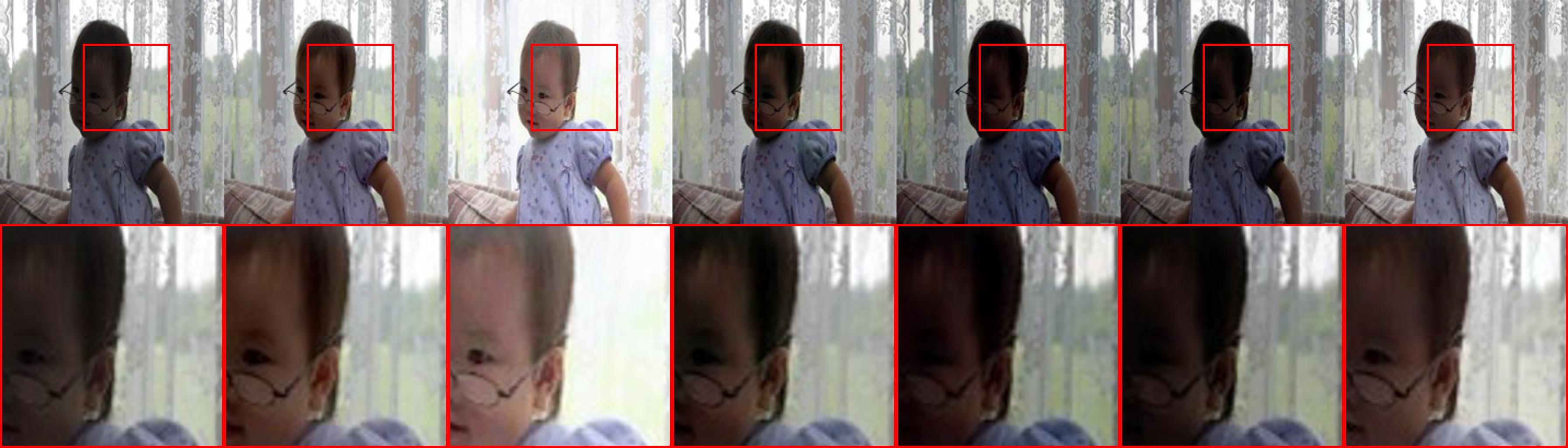} \\
    \caption{Visual comparisons of cross-dataset evaluation on the exposure correction (EC) task. The top row shows results of models trained on MSEC~\cite{afifi2021msec} but tested on SICE~\cite{cai2018sice}, while to bottom row shows the results of models trained on SICE but tested on MSEC~\cite{afifi2021msec}.Best viewed by zooming in.}

    \label{fig:method_cmp}
\end{figure*}

Tab.~\ref{tab:cross_dataset_LLIE_EC} shows the cross-dataset evaluation results for LLIE and EC tasks. UNICE demonstrates superior generalization performance over existing methods, achieving the best FR-IQA and NR-IQA scores in most cases. Models trained on our dataset show improved performance compared to task-specific datasets, validating our data generation pipeline's effectiveness. Additionally, UNICE outperforms SD-T, indicating its advantages stem primarily from our MES synthesis and fusion pipeline. Tab.~\ref{tab:cross_dataset_BIE_EC} reports the results for BIE and L2HT tasks. Similar conclusions can be made: UNICE exhibits better generalization due to its reduced reliance on human annotations. Fig.~\ref{fig:method_cmp} presents some visual comparisons on the EC task, where our proposed UNICE model demonstrates much better enhanced visual quality over its competitors. In the top row, we see that the competing models fail to enhance the under-exposed images, leaving them noticeably dark, indicating limited generalization performance. In the bottom row, most models either render the foreground too dark or the background over-bright, whereas UNICE maintains a balanced exposure, revealing the details of both the foreground and background. More visual comparisons are in the \textbf{supplementary file}.


\begin{table*}[t]
\small
    \caption{Cross-dataset evaluation on LLIE and EC tasks. ``↓(↑)'' indicates that a lower (higher) value is better. The best results are in \textbf{bold}. ``U., L., S., M.'' represent the UHD-LL, LSRW, SICE, and MSEC datasets, respectively.}
    \label{tab:cross_dataset_LLIE_EC}
    \centering
    \begin{tabular}{cclcccccccc}
        \hline
         Task &  & Method / Trainset & PSNR↑ & SSIM↑ & LPIPS↓ & DISTS↓ & NIQE↓ & PI↓ & BRIS.↓ & ARN.↑ \\
        \hline
        \multirow{16}{*}{\rotatebox{90}{\centering LLIE}}
        & \multirow{8}{*}{\rotatebox{90}{\centering UHD-LL (U.)}} 
        & CoLIE (ECCV'24) / - & 17.768 & 0.881 & 0.117 & 0.132 & 5.949 & 4.258 & 26.002 & 0.615 \\ 
        & & SCI (CVPR'22) / L. & 16.224 & 0.834 & 0.157 & 0.135 & 6.500 & 4.695 & 26.735 & 0.698 \\ 
        & & FLLIE (MM'23) / L. & 18.152 & 0.890 & 0.131 & 0.145 & 6.052 & 4.912 & 23.001 & 0.496 \\ 
        & & FLLIE (MM'23) / Ours & 19.894 & 0.873 & 0.109 & 0.125 & 5.831 & 4.385 & 22.075 & 0.607 \\ 
        & & RF (ICCV'23) / L. & 19.740 & 0.898 & 0.116 & 0.131 & 7.549 & 6.056 & 33.758 & 0.554 \\ 
        & & RF (ICCV'23) / Ours & 20.586 & 0.902 & 0.106 & 0.119 & 6.430 & 4.882 & 28.678 & 0.642 \\ 
        & & SD-T / Ours & 21.325 & \textbf{0.910} & 0.098 & 0.116 & 6.118 & 4.374 & 22.632 & 0.762 \\ 
        & & Ours / Ours & \textbf{22.007} & 0.907 & \textbf{0.090} & \textbf{0.111} & \textbf{5.684} & \textbf{4.100} & \textbf{21.058} & \textbf{0.773} \\  
        \cline{2-11}
        & \multirow{8}{*}{\rotatebox{90}{\centering LSRW (L.)}} 
        & CoLIE (ECCV'24) / - & 18.247 & 0.683 & 0.198 & 0.163 & 6.100 & 4.351 & 32.966 & 0.630 \\ 
        & & SCI (CVPR'22) / U. & 13.093 & 0.455 & 0.273 & 0.198 & 7.638 & 6.007 & 33.757 & 0.671 \\ 
        & & FLLIE (MM'23) / U. & 16.856 & 0.643 & 0.194 & 0.164 & 6.135 & 4.912 & 25.908 & 0.646 \\ 
        & & FLLIE (MM'23) / Ours & 18.382 & 0.673 & 0.178 & 0.150 & 5.942 & 4.447 & 22.308 & 0.684 \\ 
        & & RF (ICCV'23) / U. & 17.367 & 0.685 & 0.165 & 0.143 & 6.115 & 4.595 & 29.258 & 0.727 \\ 
        & & RF (ICCV'23) / Ours & 18.180 & 0.705 & 0.158 & 0.135 & 5.994 & 4.412 & 21.849 & 0.735 \\ 
        & & SD-T / Ours & 18.993 & 0.711 & 0.155 & 0.132 & 5.904 & 4.274 & 19.379 & 0.739 \\ 
        & & Ours / Ours & \textbf{19.399} & \textbf{0.718} & \textbf{0.153} & \textbf{0.129} & \textbf{5.813} & \textbf{4.137} & \textbf{16.909} & \textbf{0.741} \\  
        \hline
        \multirow{18}{*}{\rotatebox{90}{\centering EC}}
        & \multirow{8}{*}{\rotatebox{90}{\centering SICE (S.)}} 
        & CoLIE (ECCV'24) / - & 13.365 & 0.636 & 0.269 & 0.191 & 8.791 & 5.112 & 41.382 & 0.453 \\ 
        & & LCDPNet (ECCV'22) / M. & 14.820 & 0.593 & \textbf{0.247} & 0.192 & 6.997 & 4.175 & 32.158 & 0.569 \\ 
        & & LCDPNet (ECCV'22) / Ours & 15.896 & 0.624 & 0.312 & 0.203 & 6.045 & 3.804 & 27.392 & 0.643 \\ 
        & & RF (ICCV'23) / M. & 14.468 & 0.608 & 0.281 & 0.186 & 7.464 & 4.883 & 21.514 & 0.596 \\ 
        & & RF (ICCV'23) / Ours & 16.293 & 0.630 & 0.311 & 0.206 & 6.642 & 4.087 & 20.752 & 0.654 \\ 
        & & CSEC (CVPR'24) / M. & 14.665 & 0.589 & 0.274 & 0.196 & 6.133 & 3.815 & 24.358 & 0.631 \\ 
        & & CSEC (CVPR'24) / Ours. & 16.371 & 0.611 & 0.323 & 0.206 & 5.844 & 3.66 &22.712 & 0.655
        \\ 
        & & SD-T / Ours & 16.720 & 0.641 & 0.362 & 0.255 & 6.065 & 4.437 & 24.410 & 0.679 \\ 
        & & Ours / Ours & \textbf{17.509} & \textbf{0.644} & 0.355 & \textbf{0.220} & \textbf{5.410} & \textbf{3.557} & \textbf{20.244} & \textbf{0.692} \\ 
        \cline{2-11}
        & \multirow{8}{*}{\rotatebox{90}{\centering MSEC (M.)}} 
        & CoLIE (ECCV'24) / - & 11.537 & 0.710 & 0.219 & 0.166 & 5.369 & 3.709 & 20.960 & 0.570 \\ 
        & & LCDPNet (ECCV'22) / S. & 16.872 & 0.812 & 0.165 & 0.155 & 5.381 & 3.864 & 19.403 & 0.592 \\ 
        & & LCDPNet (ECCV'22) / Ours & 18.617 & 0.851 & 0.148 & 0.130 & 5.307 & 3.731 & 19.834 & 0.678 \\ 
        & & RF (ICCV'23) / S. & 15.846 & 0.808 & 0.154 & 0.149 & 5.229 & 3.679 & 20.145 & 0.655 \\ 
        & & RF (ICCV'23) / Ours & 18.207 & 0.836 & 0.141 & 0.128 & 5.240 & 3.620 & 19.736 & 0.716 \\ 
        & & CSEC (CVPR'24) / S. & 17.001 & 0.832 & 0.135 & 0.129 & 5.141 & 3.628 & 19.984 & 0.677 \\ 
        & & CSEC (CVPR'24) / Ours & 18.113 & 0.859 & 0.130 & 0.123 & 5.211 & 3.589 & 19.639 & 0.729 \\ 
        & & SD-T / Ours & 19.141 & 0.774 & 0.144 & 0.122 & 5.699 & 4.060 & 20.758 & 0.778 \\ 
        & & Ours / Ours & \textbf{19.781} & \textbf{0.877} & \textbf{0.122} & \textbf{0.114} & \textbf{5.257} & \textbf{3.531} & \textbf{19.122} & \textbf{0.807} \\ 
        \hline
    \end{tabular}
\end{table*}

\begin{table*}[ht]
\small
    \caption{Cross-dataset evaluation on BIE and L2HT tasks. ``↓(↑)'' indicates that a lower (higher) value is better. The best results are in \textbf{bold}. Note that the Backlit dataset does not have GT, therefore FR-IQA metrics cannot be calculated, denoted by symbol `-'. "B. L. E. R." stands for the BAID, Backlit, HDR-Eye, and HDR-Real datasets, respectively.}
    \label{tab:cross_dataset_BIE_EC}
    \centering
    \begin{tabular}{cclcccccccc}
        \hline
         &  & Method / Trainset & PSNR↑ & SSIM↑ & LPIPS↓ & DISTS↓ & NIQE↓ & PI↓ & BRIS.↓ & ARN.↑ \\ 
        \hline
        \multirow{18}{*}{\rotatebox{90}{\centering BIE}}
        & \multirow{9}{*}{\rotatebox{90}{\centering BAID (B.)}} 
        & CoLIE (ECCV'24) / - & 18.263 & 0.847 & 0.117 & 0.108 & 5.665 & 3.760 & 26.401 & 0.669 \\ 
        & & ZDCE++ (PAMI'21) / L. & 11.601 & 0.454 & 0.283 & 0.183 & 6.561 & 4.490 & 27.692 & 0.532 \\
        & & ZDCE++ (PAMI'21) / Ours. & 12.124 & 0.649 & 0.281 & 0.178 & 6.432 & 4.201 & 26.019 & 0.548 \\ 
        & & SCI (CVPR'22) / L. & 16.797 & 0.698 & 0.228 & 0.150 & 5.671 & 3.848 & 25.467 & 0.651 \\ 
        & & SCI (CVPR'22) / Ours & 17.139 & 0.737 & 0.213 & 0.139 & 5.473 & 3.815 & 24.249 & 0.693 \\
        & & CLIP-LIT (ICCV'23) / L. & 17.215 & 0.756 & 0.153 & 0.142 & 5.582 & 3.714 & 25.416 & 0.712 \\
        & & CLIP-LIT (ICCV'23) / Ours & 17.539 & 0.799 & 0.122 & 0.131 & 5.433 & 3.635 & 24.124 & 0.733 \\ 
        & &  SD-T / Ours & 18.472 & 0.830 & 0.113 & 0.118 & 5.295 & 3.564 & 21.319 & 0.746 \\
        & & Ours / Ours & \textbf{19.058} & \textbf{0.876} & \textbf{0.095} & \textbf{0.106} & \textbf{5.178} & \textbf{3.539} & \textbf{20.664} & \textbf{0.817} \\ 
        \cline{2-11}
        & \multirow{9}{*}{\rotatebox{90}{\centering Backlit (L.)}} 
        & CoLIE (ECCV'24) / - & - & - & - & - & 4.635 & 3.151 & 18.428 & 0.619 \\ 
        & & CSEC (CVPR'23) / B. & - & - & - & - & 4.983 & 3.475 & 18.295 & 0.609 \\
        & & CSEC (CVPR'23) / Ours. & - & - & - & - & 4.816 & 3.443 & 17.191 & 0.644 \\
        & & RF (ICCV'23) / B. & - & - & - & - & 4.783 & 3.378 & 18.001 & 0.613 \\ 
        & & RF (ICCV'23) / Ours. & - & - & - & - & 4.701 & 3.329 & 17.334 & 0.693 \\ 
        & & CLIP-LIT (ICCV'23) / B. & - & - & - & - & 4.867 & 3.248 & 19.139 & 0.682 \\
        & & CLIP-LIT (ICCV'23) / Ours. & - & - & - & - & 4.813 & 3.319 & 18.928 & 0.701 \\
        & & SD-T / Ours & - & - & - & - & 4.806 & 3.271 & 17.139 & 0.728 \\ 
        & & Ours / Ours & - & - & - & - & \textbf{4.627} & \textbf{3.123} & \textbf{16.177} & \textbf{0.782} \\
        \hline
        \multirow{22}{*}{\rotatebox{90}{\centering L2HT}}
        & \multirow{11}{*}{\rotatebox{90}{\centering HDR-Eye (E.)}} 
        & CoLIE (ECCV'24) / - & 13.268 & 0.597 & 0.234 & 0.157 & 4.399 & 3.259 & 25.514 & 0.552 \\ 
        & & SHDR (CVPR'20) / R. & 12.943 & 0.483 & 0.463 & 0.342 & 8.761 & 6.558 & 31.414 & 0.373 \\ 
        & & SHDR (CVPR'20) / Ours & 13.254 & 0.495 & 0.434 & 0.262 & 8.992 & 6.305 & 30.283 & 0.411 \\ 
        & & CEVR (ICCV'23) / R. & 12.998 & 0.484 & 0.464 & 0.343 & 8.935 & 6.537 & 31.334 & 0.376 \\ 
        & & CEVR (ICCV'23) / Ours & 13.486 & 0.501 & 0.381 & 0.247 & 8.767 & 6.318 & 30.342 & 0.409 \\ 
        & & RF (CVPR'23) / R. & 14.265 & 0.576 & 0.337 & 0.205 & 3.743 & 5.654 & 26.886 & 0.503 \\ 
        & & RF (CVPR'23) / Ours & 14.508 & 0.649 & 0.299 & 0.178 & 3.652 & 5.427 & 25.809 & 0.599 \\ 
        & & CSEC (CVPR'24) / R. & 14.790 & 0.661 & 0.322 & 0.164 & 3.621 & 4.979 & 27.665 & 0.605 \\ 
        & & CSEC (CVPR'24) / Ours & 15.295 & 0.656 & 0.306 & 0.158 & 3.567 & 4.715 & 24.718 & 0.657 \\
        & &  SD-T / Ours & 15.404 & 0.671 & 0.231 & 0.152 & 3.528 & 3.788 & 24.357 & 0.693 \\ 
        & & Ours / Ours & \textbf{16.154} & \textbf{0.687} & \textbf{0.216} & \textbf{0.147} & \textbf{3.481} & \textbf{3.049} & \textbf{23.497} & \textbf{0.732} \\ 
        \cline{2-11}
        & \multirow{11}{*}{\rotatebox{90}{\centering HDR-Real (R.)}} 
        & CoLIE (ECCV'24) / - & 11.970 & 0.520 & 0.389 & 0.272 & 7.950 & 5.436 & 26.389 & 0.541 \\ 
        & & SHDR (CVPR'20) / E. & 12.998 & 0.484 & 0.464 & 0.343 & 8.767 & 6.318 & 30.342 & 0.409 \\ 
        & & SHDR (CVPR'20) / Ours & 13.394 & 0.515 & 0.438 & 0.324 & 8.406 & 6.044 & 29.142 & 0.449 \\ 
        & & CEVR (ICCV'23) / E. & 13.473 & 0.471 & 0.492 & 0.371 & 9.076 & 6.664 & 31.893 & 0.357 \\ 
        & & CEVR (ICCV'23) / Ours & 14.181 & 0.534 & 0.419 & 0.308 & 8.155 & 5.828 & 28.188 & 0.481 \\
        & & RF (CVPR'23) / E. & 14.335 & 0.623 & 0.364 & 0.279 & 7.498 & 5.277 & 26.178 & 0.516 \\ 
        & & RF (CVPR'23) / Ours & 15.044 & 0.642 & 0.353 & 0.261 & 7.168 & 5.026 & 25.762 & 0.558 \\ 
        & & CSEC (CVPR'24) / E. & 14.386 & 0.664 & 0.341 & 0.274 & 6.948 & 4.871 & 25.859 & 0.587 \\ 
        & & CSEC (CVPR'24) / Ours & 15.488 & 0.675 & 0.339 & 0.254 & 6.591 & 4.760 & 24.637 & 0.599 \\ 
        & &  SD-T / Ours & 16.148 & 0.701 & 0.301 & 0.223 & 5.812 & 4.218 & 23.718 & 0.670 \\
        & & Ours / Ours & \textbf{16.818} & \textbf{0.721} & \textbf{0.265} & \textbf{0.201} & \textbf{5.455} & \textbf{3.615} & \textbf{23.034} & \textbf{0.764} \\ 
        \cline{2-11}
        \hline
    \end{tabular}
\end{table*}


\subsection{User Study}
\begin{figure}[t]
    \centering
    \includegraphics[width=1.0\linewidth]{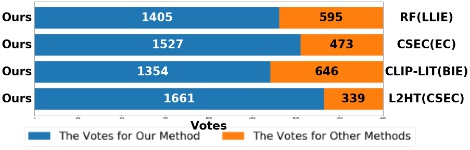}
    \caption{Voting statistics of competing methods in the user study. UNICE wins the most votes.}
    \label{fig:user_study_votes}
\end{figure}

We conducted a user study to evaluate the visual quality of contrast enhancement results obtained by different methods. For each of the four tasks (LLIE, EC, BIE, and L2HT), we select the best task-specific model (based on their quantitative metrics in the experiments) as the competitor of UNICE, \textit{i.e.}, RF~\cite{cai2023retinexformer} for LLIE, CSEC~\cite{li_2024_cvpr_csec} for EC, CLIP-LIT~\cite{liang2023CLIP-LIT} for BIE, and CSEC~\cite{li_2024_cvpr_csec} for L2HT. For each task, we randomly select 200 images from the corresponding datasets and compare the pair of enhanced images by UNICE and the competitor. Ten volunteers are invited to vote for the better image in each pair. Ultimately, as shown in Fig.~\ref{fig:user_study_votes}, UNICE received 70\%, 76\%, 68\%, and 83\% of the votes for the four tasks, respectively.

\subsection{Complexity Analysis}
We compare the model size, FLOPs and FPS of UNICE, RF~\cite{cai2023retinexformer}, LCDP~\cite{wang2022lcdp}, CSEC~\cite{li_2024_cvpr_csec} and SD-Turbo (SD-T)~\cite{parmar2024pix2pix_turbo} in Tab.~\ref{tab:Model Parameters}. All models are tested on an NVIDIA A100 40GB GPU. Compared with task-specific models RF~\cite{cai2023retinexformer}, LCDP~\cite{wang2022lcdp} and CSEC~\cite{li_2024_cvpr_csec}, UNICE has more parameters and higher FLOPs since it builds upon the pre-trained SD-Turbo. However, we focus on generalized contrast enhancement, which cannot be achieved by task-specific models. 

\begin{table}[t]
\small
\caption{Complexity comparison of representative methods at 512×512 resolution. }
\centering
\begin{tabular}{l|c|c|c|c|c}
\hline
         & LCDP & CSEC & RF & SD-T & UNICE \\ \hline
Input size & 512 & 512 & 512 & 512 & 512 \\
Param. (M) & 0.28 & 0.30 & 1.61 & 946 & 965 \\
FPS        & 13.91 & 13.83 & 12.71 & 7.08 & 5.58 \\
FLOPs (G)  & 19 & 20 & 64 & 1,648 & 3,348 \\ \hline
\end{tabular}
\label{tab:Model Parameters}
\end{table}

\subsection{Ablation Study}

\subsubsection{Pseudo-GT Generation Methods}

Our data generation pipeline ensembles MEF algorithms to generate pseudo-GT for network training. The candidate MEF methods include FMMEF~\cite{li2020fmmef}, GradientMEF~\cite{lee2018GG}, MDO~\cite{ulucan178MDO}, Mertens~\cite{mertens2009pixelmef}, and PerceptualMEF~\cite{liu2022perceptual}. We randomly blend the results and select the best outcome using NR-IQA methods. The results are shown in Tab.~\ref{tab:pseudo_GT_methods}. It is evident that the ensembled pipeline can improve the quality.

\begin{table}[t]
\small
\caption{NR-IQA metrics for various multi-exposure fusion (MEF) methods. ``↓(↑)'' indicates lower(higher) score is better. The best results are in \textbf{bold}.}
\label{tab:pseudo_GT_methods}
\centering
\begin{tabular}{l|c|c|c|c}
\hline
Method & NIQE↓ & PI↓ & BRIS.↓ & ARN.↑ \\ \hline
FMMEF~\cite{li2020fmmef} & 5.079 & 3.493 & 18.302 & 0.739 \\ 
GradientMEF~\cite{lee2018GG} & 4.939 & 3.592 & 18.273 & 0.667 \\ 
MDO~\cite{ulucan178MDO} & 5.256 & 3.869 & 20.083 & 0.644 \\ 
Mertens~\cite{mertens2009pixelmef} & 4.985 & 3.613 & 18.318 & 0.710 \\ 
PerceptualMEF~\cite{liu2022perceptual} & 5.099 & 3.920 & 19.313 & 0.666 \\ 
Ensambled & \textbf{4.910} & \textbf{3.474} & \textbf{17.912} & \textbf{0.746} \\ 
\hline
\end{tabular}
\end{table}

\subsubsection{Network Backbone Selection}

\begin{figure}
    \centering
    \includegraphics[width=1.\linewidth]{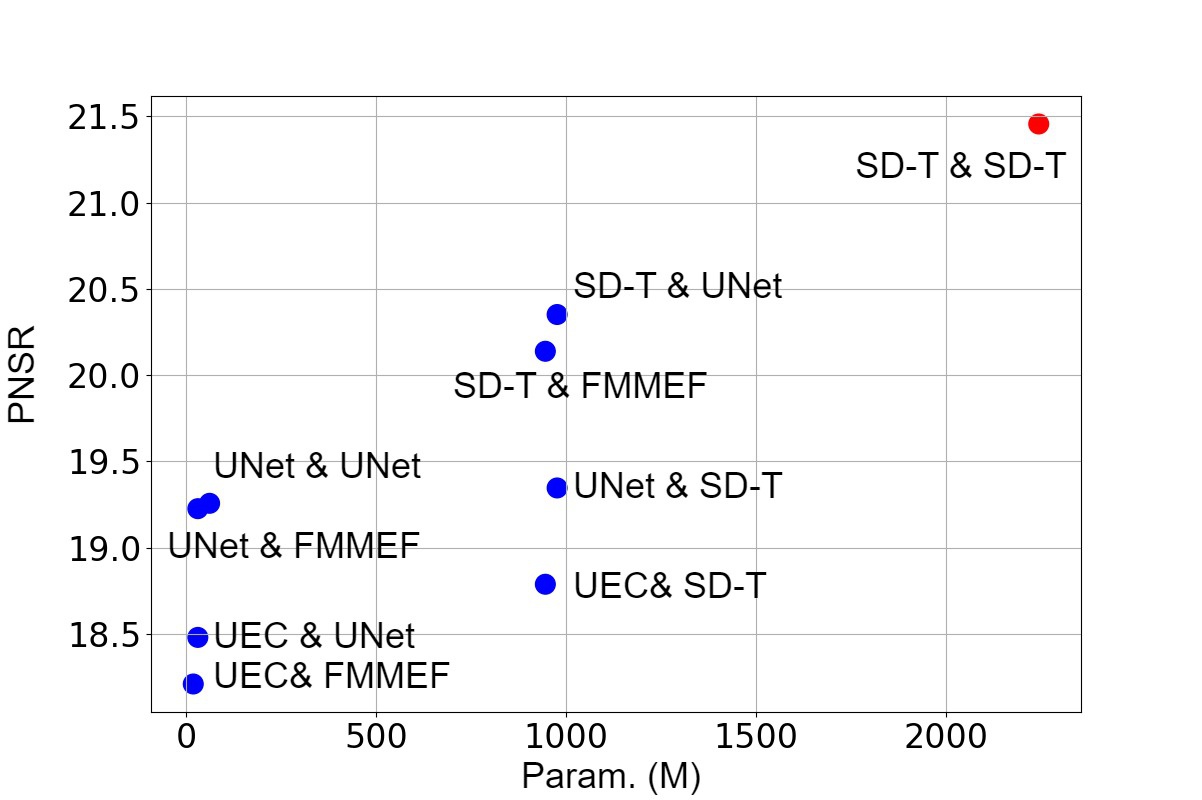}
    \caption{PSNR vs. model size for different combinations of the backbone networks in the two stages of UNICE.}
    \label{fig:various backbones}
\end{figure}

For the network architecture of UNICE, we select SD-Turbo as the backbone, which is a large model. To verify the effectiveness of our design, we also employ other backbones for comparison. For Stage 1, three network backbones are used: UEC~\cite{ruodai2024uec}, UNet~\cite{ronneberger2015unet} and SD-T~\cite{parmar2024pix2pix_turbo}. For Stage 2, there are also three options: directly using FMMEF~\cite{li2020fmmef}, UNet~\cite{ronneberger2015unet}, and SD-T~\cite{parmar2024pix2pix_turbo}. By combining the two stages, we obtain 9 combinations of backbones. We then evaluate them on the task-specific datasets and calculate the average performance. The results are shown in Tab.~\ref{tab:backbone_ablation}, where the SD-T model demonstrates clear advantages. We further visualize the relationship between PSNR and model size in Fig.~\ref{fig:various backbones}. It can be seen that SD-T brings much improvement over other models in both stages.

\begin{table*}[t]
\caption{Ablation on the selection of network backbones in both stages of UNICE. ``↓(↑)'' indicates lower(higher) score is better. The best results are in \textbf{bold}.}
\label{tab:backbone_ablation}
\centering
\begin{tabular}{l|l|c|c|c|c|c|c|c|c|c}
\hline
Stage1 & Stage2 & Param. &  PSNR↑ & SSIM↑ & LPIPS↓ & DISTS↓ & NIQE↓ & PI↓ & BRIS.↓ & ARN.↑ \\ \hline
UEC~\cite{ruodai2024uec} & FMMEF~\cite{li2020fmmef} & 19K & 18.210 & 0.763 & 0.172 & 0.156 & 6.081 & 4.568 & 29.103 & 0.578 \\ 
UEC~\cite{ruodai2024uec} & UNet~\cite{ronneberger2015unet} & 31M & 18.480 & 0.717 & 0.185 & 0.158 & 6.326 & 4.617 & 33.090 & 0.618 \\ 
UEC~\cite{ruodai2024uec} & SD-T~\cite{parmar2024pix2pix_turbo} & 946M & 18.788 & 0.704 & 0.159 & 0.135 & 5.926 & 4.335 & 20.005 & 0.730 \\ 
UNet~\cite{ronneberger2015unet} & FMMEF~\cite{li2020fmmef} & 31M & 19.226 & 0.892 & 0.111 & 0.125 & 6.198 & 4.581 & 27.387 & 0.629 \\ 
UNet~\cite{ronneberger2015unet} & UNet~\cite{ronneberger2015unet} & 62M & 19.257 & 0.855 & 0.129 & 0.137 & 7.312 & 5.865 & 32.444 & 0.569 \\ 
UNet~\cite{ronneberger2015unet} & SD-T~\cite{parmar2024pix2pix_turbo} & 977M & 19.346 & 0.753 & 0.144 & 0.129 & 6.021 & 4.409 & 21.442 & 0.717 \\ 
SD-T~\cite{parmar2024pix2pix_turbo} & FMMEF~\cite{li2020fmmef} & 946M & 20.138 & 0.899 & 0.108 & 0.121 & 6.353 & 4.783 & 28.252 & 0.638 \\ 
SD-T~\cite{parmar2024pix2pix_turbo} & UNet~\cite{ronneberger2015unet} & 977M & 20.351 & 0.897 & 0.101 & 0.119 & 5.788 & 4.162 & 22.989 & 0.711 \\ 
SD-T~\cite{parmar2024pix2pix_turbo} & SD-T~\cite{parmar2024pix2pix_turbo} & 2245M & \textbf{21.455} & \textbf{0.905} & \textbf{0.091} & \textbf{0.112} & \textbf{5.695} & \textbf{4.120} & \textbf{21.131} & \textbf{0.761} \\ \hline
\end{tabular}
\end{table*}

\section{Conclusion}

In this paper, we introduced UNICE, a universal and generalized model for various contrast enhancement tasks. We collected a large-scale dataset of raw images with high dynamic range, and used it to render a set of MES and the corresponding pseudo GTs. 
With the constructed dataset and the pre-trained SD-Turbo model, we trained an MES-Net to synthesize an MES from a single sRGB image, and an MEF-Net to fuse the MES into a high-quality contrast enhanced sRGB image. Equipped with the MES-Net and MEF-Net, our UNICE model demonstrated significantly stronger generalization performance over existing contrast enhancement methods across different tasks. Its outputs can even outperform the GTs in terms of no-reference quality evaluation. UNICE also demonstrated competitive full-reference quality metrics by fine-tuning on each dataset. It provided a new solution to achieve generalized and high quality image contrast and dynamic range enhancement. 



\bibliographystyle{IEEEtran}
\bibliography{main}

\clearpage 
\onecolumn 
\begin{center}
    \LARGE\bfseries 
    Supplementary Material for ``UNICE: Training A Universal Image Contrast Enhancer''
\end{center}
\vspace{1.5cm} 
\setcounter{page}{1}

\noindent In the supplementary file, we provide the following materials:

\begin{enumerate}
    \item More visual comparisons with GT images across various datasets and tasks (referring to Sec. 4.2 of the main paper);
    \item More results on cross-dataset experiments (referring to Sec. 4.3 of the main paper).
\end{enumerate}


\section{More Visual Comparisons with GT}

\begin{figure*}[htbp]
    \centering
    \makebox[0pt][r]{\raisebox{6.5cm}{\rotatebox{90}{Input}}\hspace{0.1cm}}%
    \makebox[0pt][r]{\raisebox{4cm}{\rotatebox{90}{GT}}\hspace{0.1cm}}%
    \makebox[0pt][r]{\raisebox{1cm}{\rotatebox{90}{Ours}}\hspace{0.1cm}}%
    \begin{subfigure}{0.16\textwidth}
        \includegraphics[width=\textwidth]{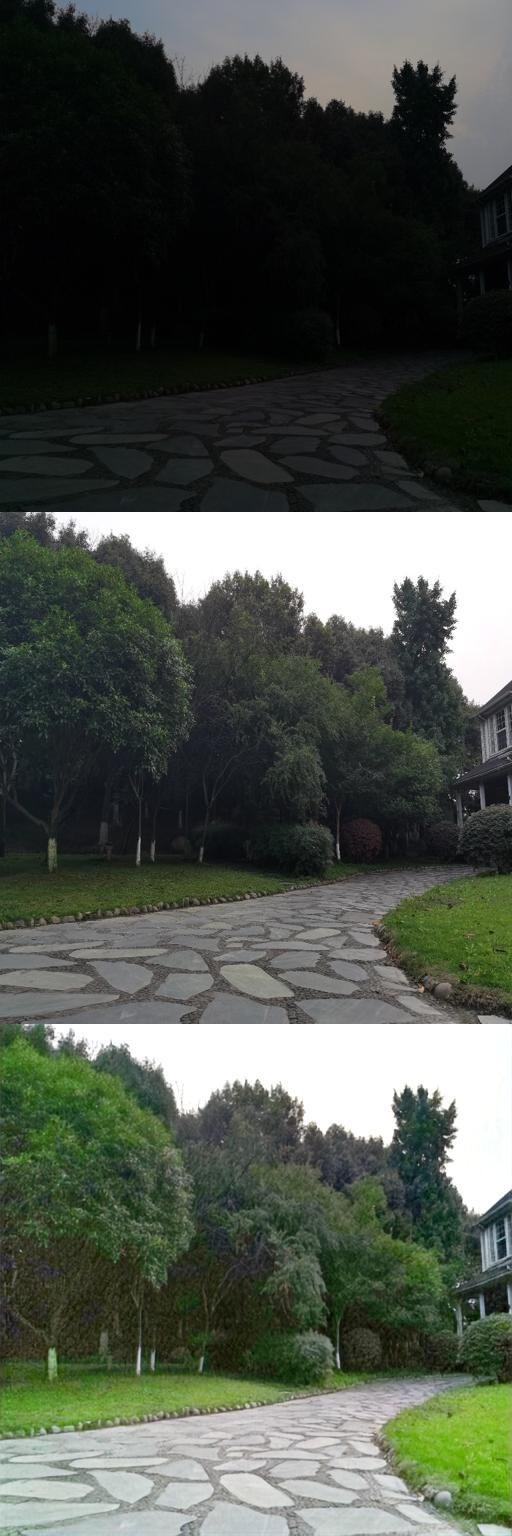}
    \end{subfigure}
    \begin{subfigure}{0.16\textwidth}
        \includegraphics[width=\textwidth]{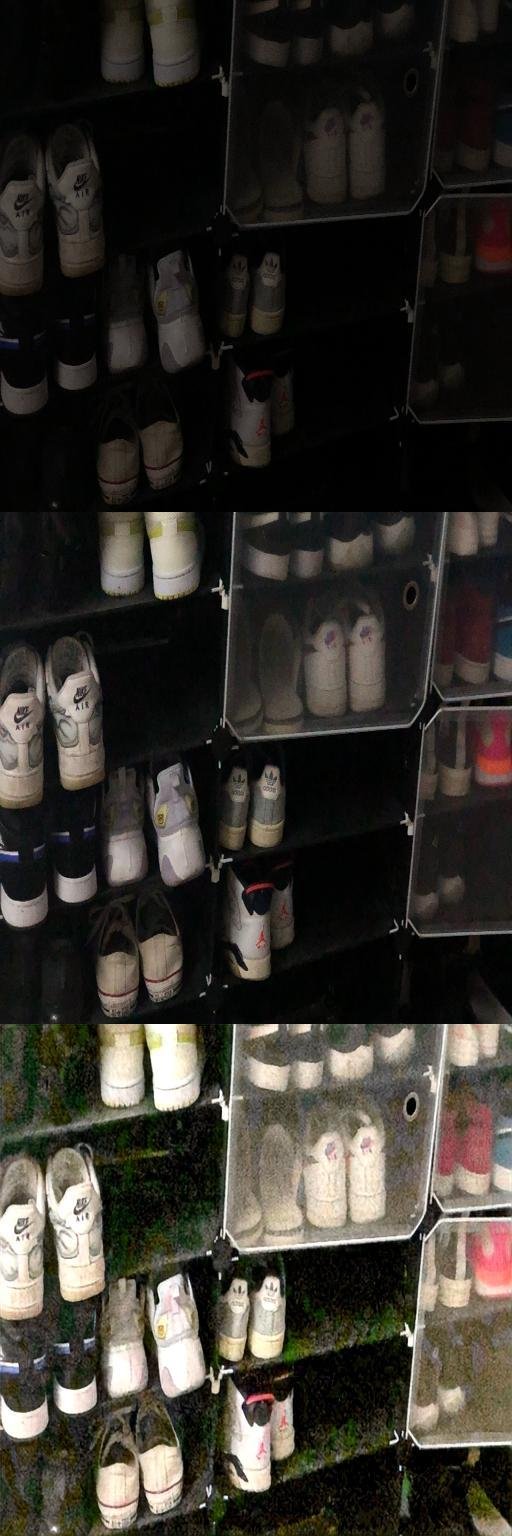}
    \end{subfigure}
    \begin{subfigure}{0.16\textwidth}
        \includegraphics[width=\textwidth]{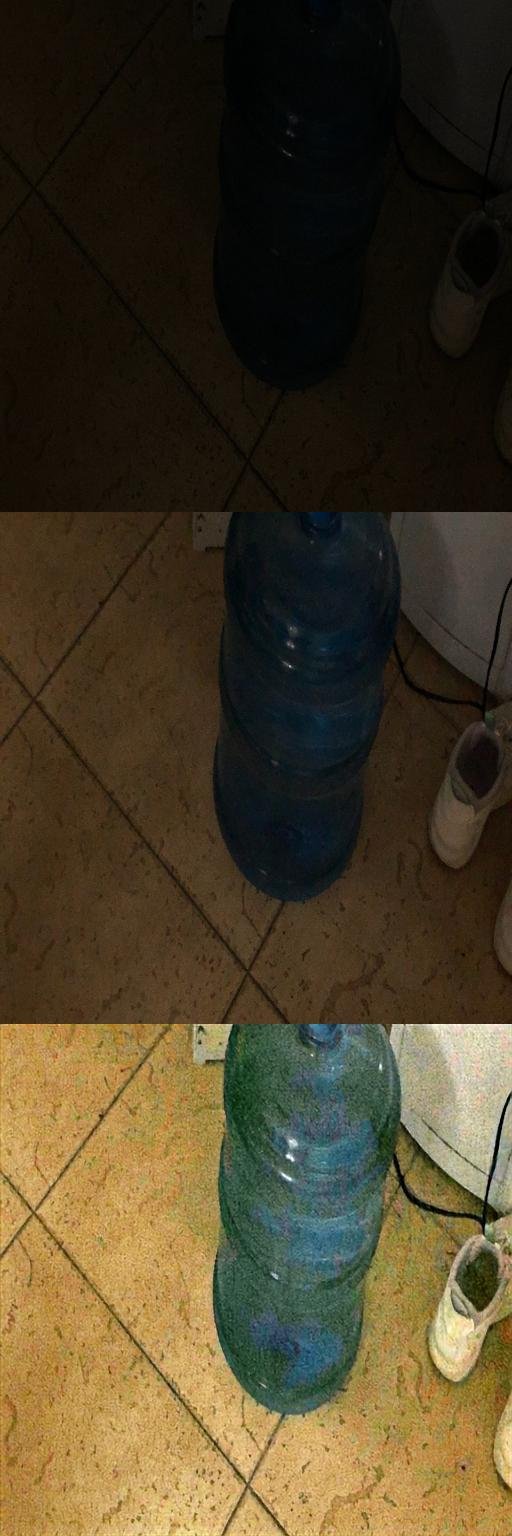}
    \end{subfigure}
    \begin{subfigure}{0.16\textwidth}
        \includegraphics[width=\textwidth]{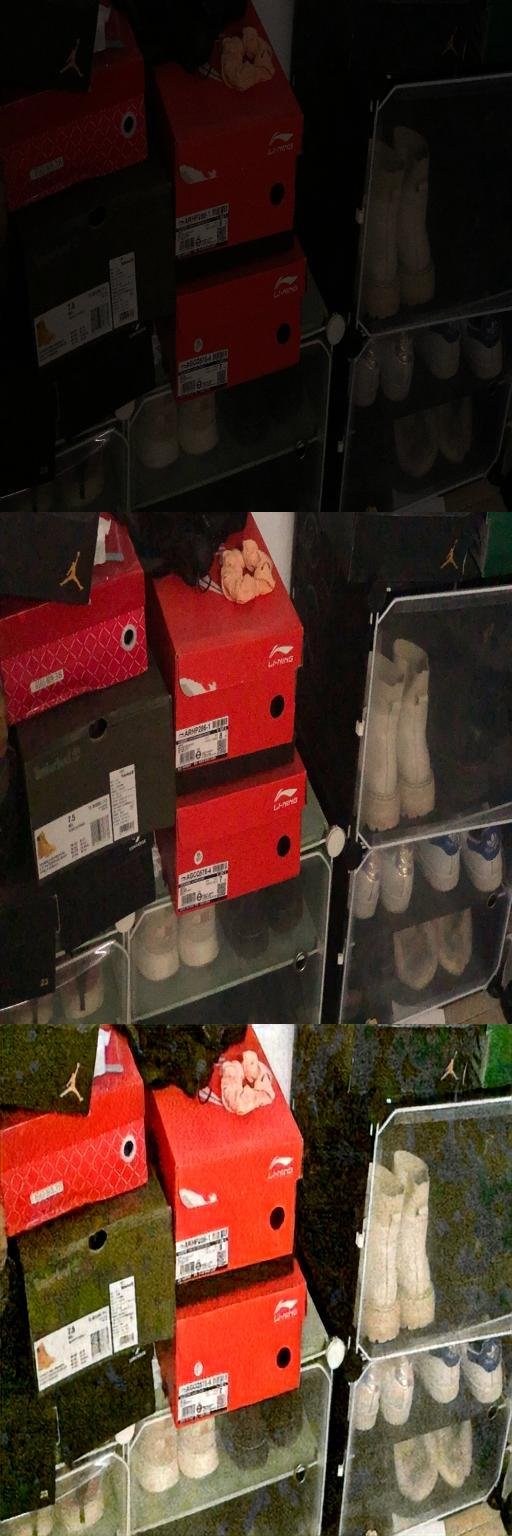}
    \end{subfigure}
    \begin{subfigure}{0.16\textwidth}
        \includegraphics[width=\textwidth]{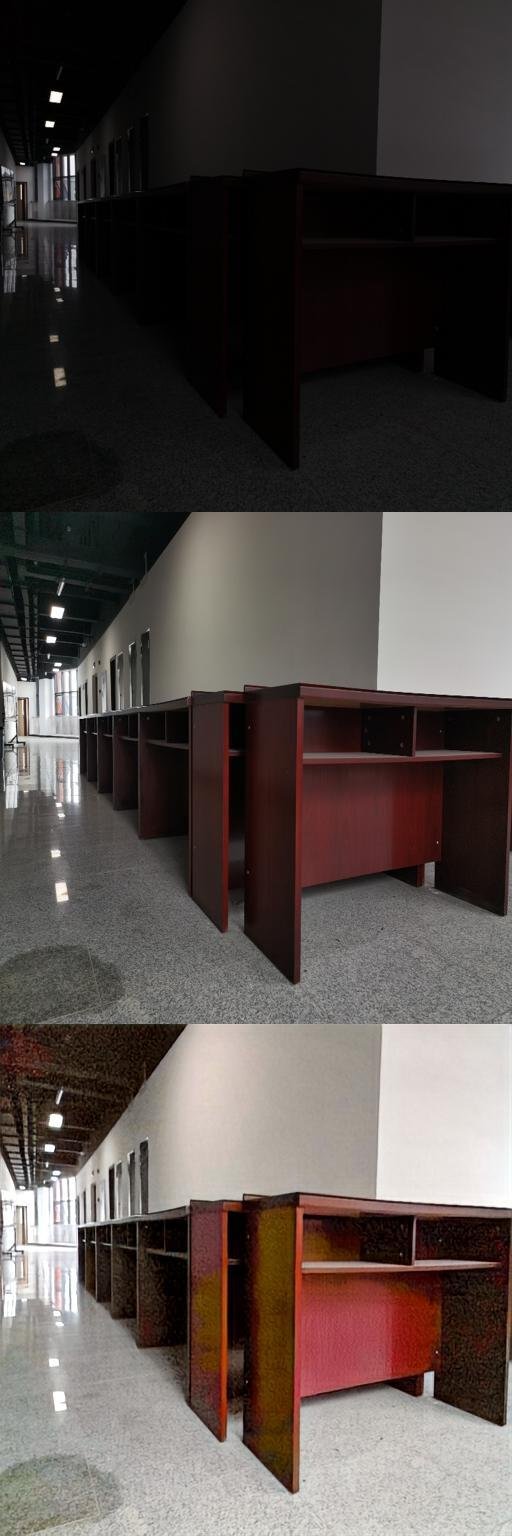}
    \end{subfigure}
    \begin{subfigure}{0.16\textwidth}
        \includegraphics[width=\textwidth]{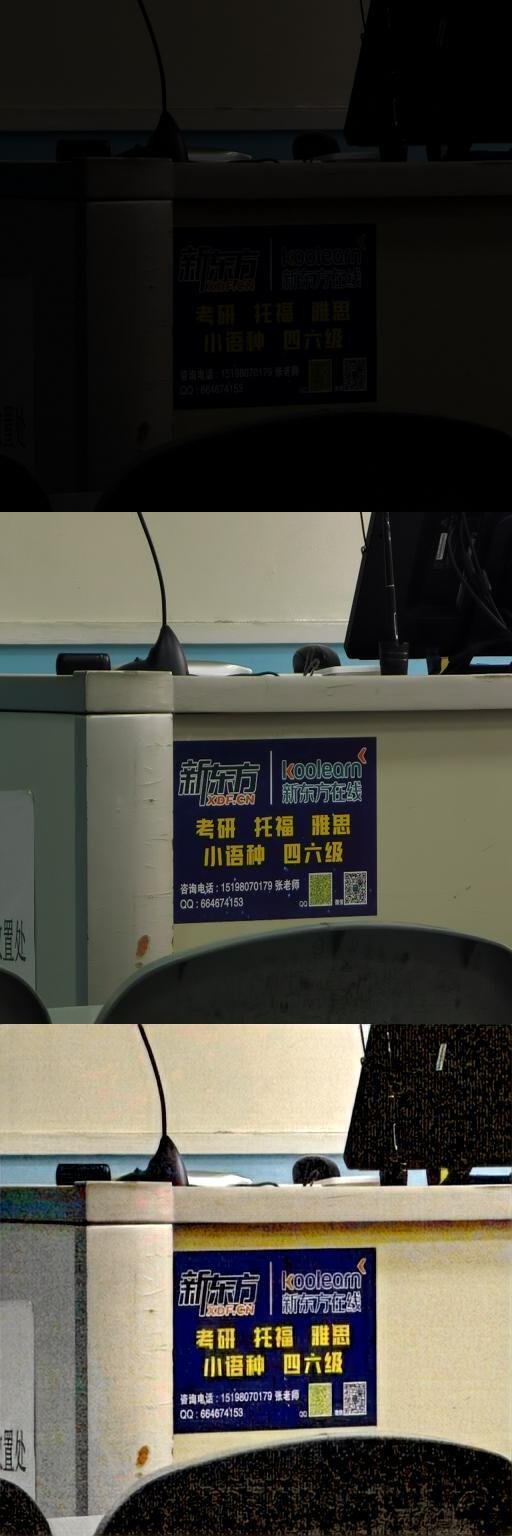}
    \end{subfigure}
    \caption{Visual comparison with GT on the LSRW~\cite{hai2023r2rnet} dataset. }
    \label{fig:GT_CMP_LSRW}
\end{figure*}

\begin{figure*}[htbp]
    \centering
    \makebox[0pt][r]{\raisebox{6.5cm}{\rotatebox{90}{Input}}\hspace{0.1cm}}%
    \makebox[0pt][r]{\raisebox{4cm}{\rotatebox{90}{GT}}\hspace{0.1cm}}%
    \makebox[0pt][r]{\raisebox{1cm}{\rotatebox{90}{Ours}}\hspace{0.1cm}}%
    \begin{subfigure}{0.16\textwidth}
        \includegraphics[width=\textwidth]{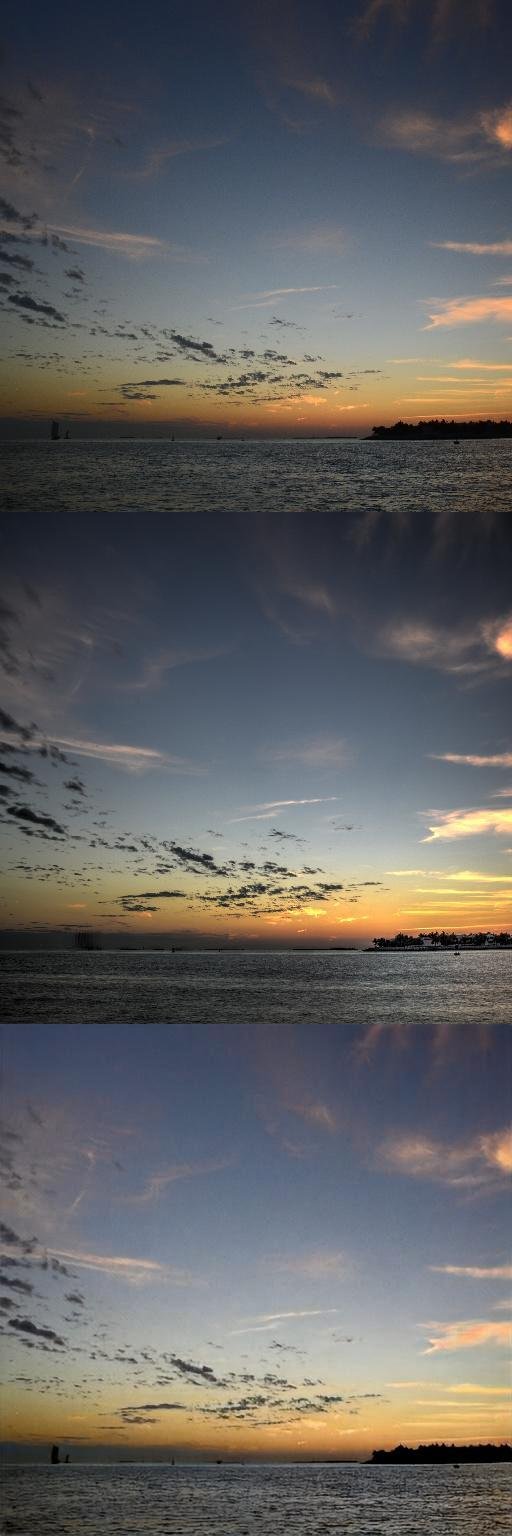}
    \end{subfigure}
    \begin{subfigure}{0.16\textwidth}
        \includegraphics[width=\textwidth]{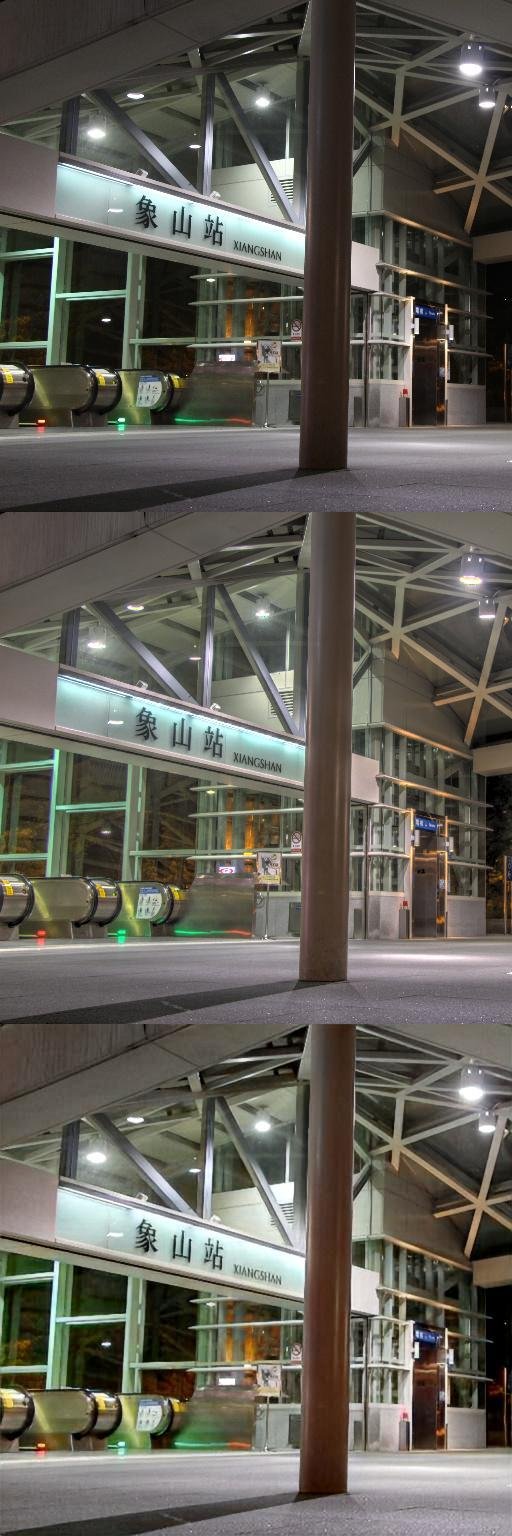}
    \end{subfigure}
    \begin{subfigure}{0.16\textwidth}
        \includegraphics[width=\textwidth]{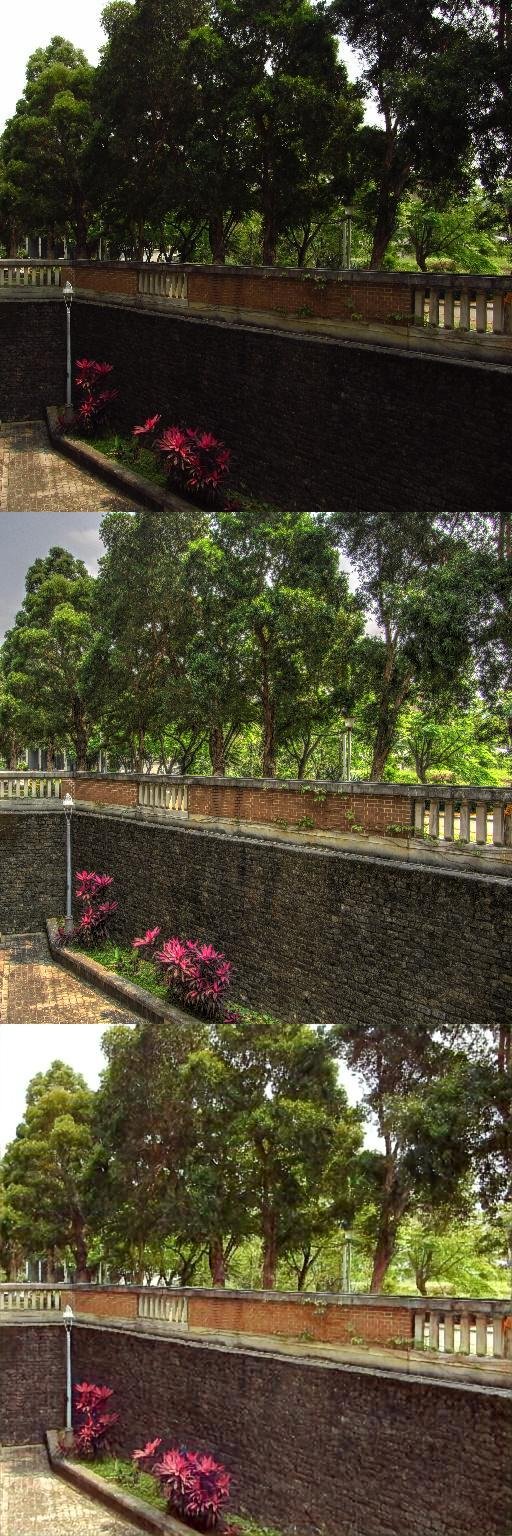}
    \end{subfigure}
    \begin{subfigure}{0.16\textwidth}
        \includegraphics[width=\textwidth]{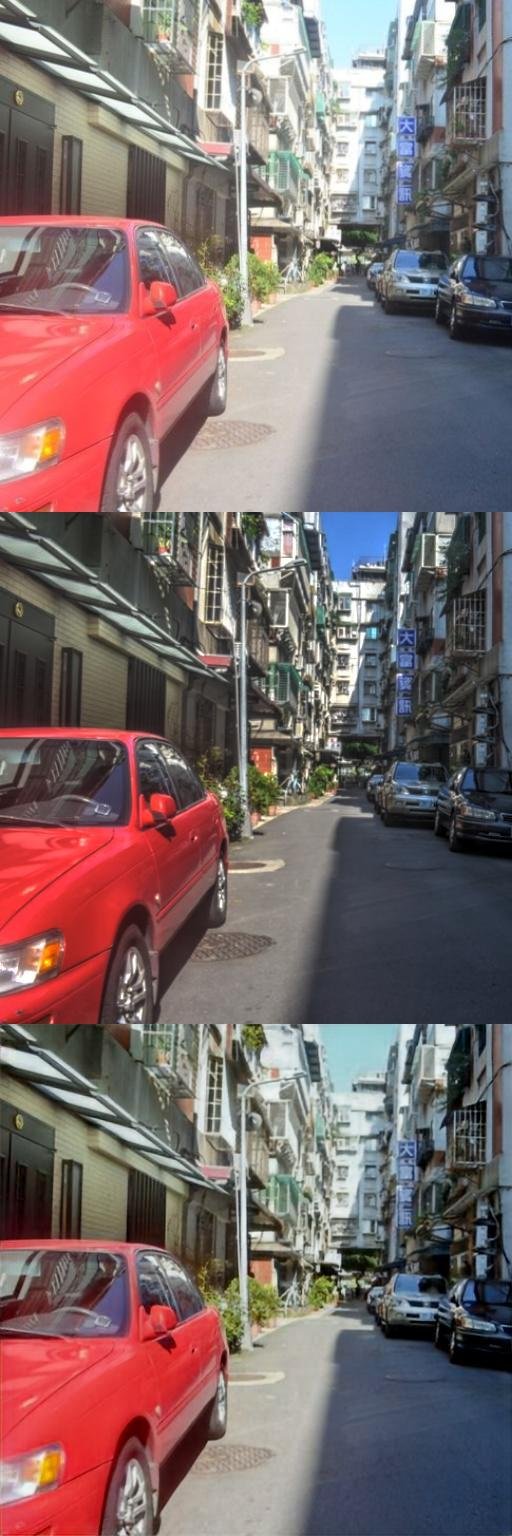}
    \end{subfigure}
    \begin{subfigure}{0.16\textwidth}
        \includegraphics[width=\textwidth]{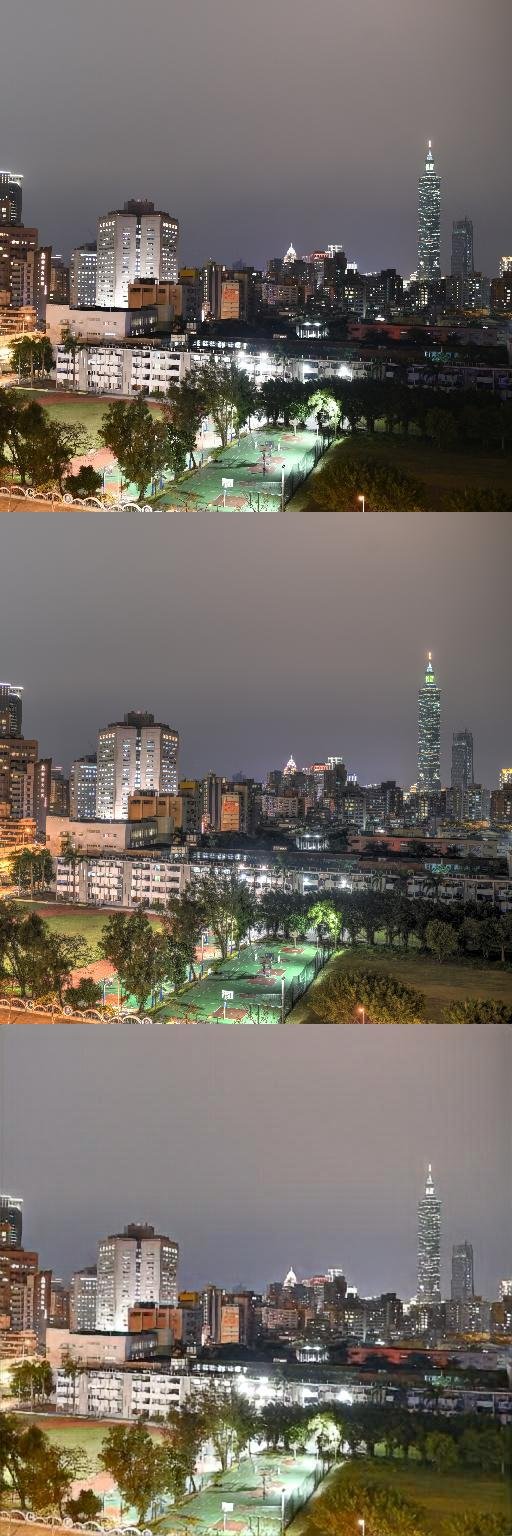}
    \end{subfigure}
    \begin{subfigure}{0.16\textwidth}
        \includegraphics[width=\textwidth]{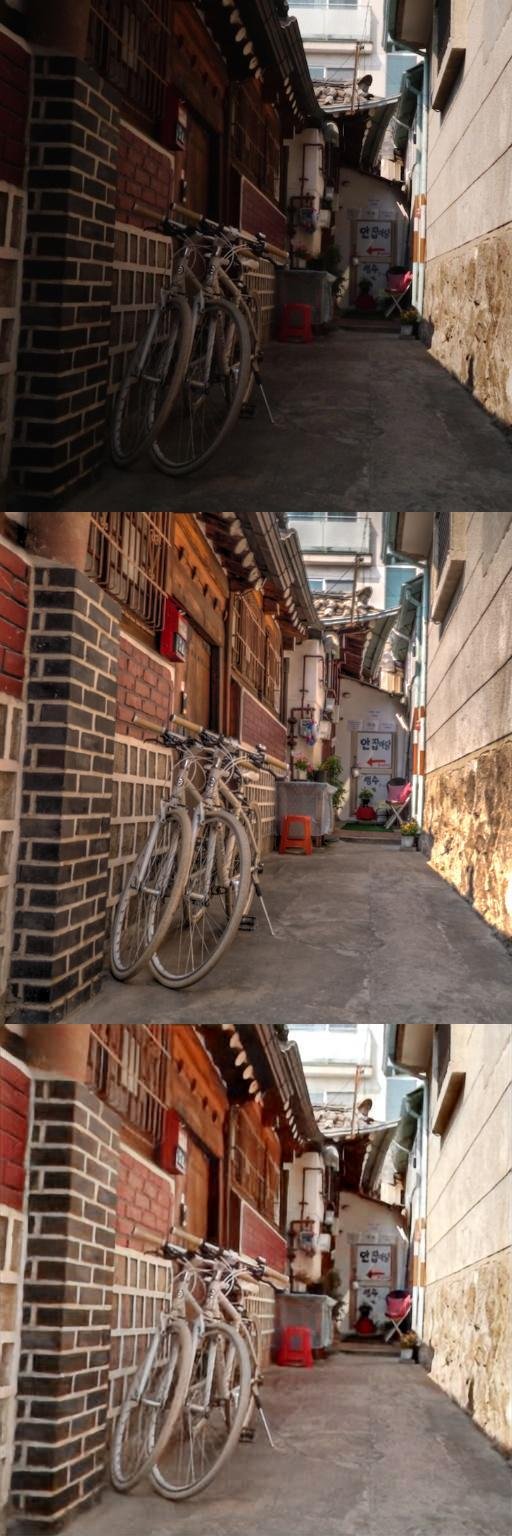}
    \end{subfigure}
    \caption{Visual comparison with GT on the HDRReal dataset~\cite{liu2020reverse}. }
    \label{fig:GT_CMP_HDRReal}
\end{figure*}

In Sec. 4.2 of the main paper, we have demonstrated that UNICE, despite being trained without human annotated data, can deliver high-quality contrast enhancement results on various contrast enhancement tasks. In this supplementary file, we provide additional visual results, which are summarized in Tab.~\ref{tab:visual_results_GT} and presented in Fig.~\ref{fig:GT_CMP_LSRW}$\sim$Fig.~\ref{fig:GT_CMP_HDRReal}. (Note that the Backlit~\cite{liang2023CLIP-LIT} dataset has no GT, therefore no visual comparison can be provided.) While the GT quality of these datasets is generally higher than the input, they are not perfect. For instance, many details in the LSRW~\cite{hai2023r2rnet}, UHD-LL~\cite{Li2023uhdll}, MSEC~\cite{afifi2021msec}, HDREye~\cite{liu2020reverse} and HDRReal~\cite{liu2020reverse} datasets remain dark and unclear. In the BAID~\cite{lv2022backlitnet} dataset, some images exhibit local overexposure, such as the loss of texture in the clouds. The SICE~\cite{cai2018sice} dataset contains images that are excessively enhanced by multi-exposure fusion algorithms, resulting in halo artifacts and unrealistic appearance. In contrast, UNICE, which is trained on large scale of data and leverages the generative prior of pre-trained SD-Turbo model, can perform better in many scenarios.

\begin{table*}[h]
\centering
\caption{Summary of visual comparisons with GT on different datasets and tasks. Note that the Backlit~\cite{liang2023CLIP-LIT} dataset has no GT, therefore no visual comparison can be provided.}
\label{tab:visual_results_GT}
\begin{tabular}{c|c|c|c|c|c|c|c}
\hline
Task & \multicolumn{2}{c|}{LLIE} & \multicolumn{2}{c|}{EC} & BIE & \multicolumn{2}{c}{L2HT} \\
\hline
Dataset & LSRW~\cite{hai2023r2rnet} & UHD-LL~\cite{Li2023uhdll} & SICE~\cite{cai2018sice} & MSEC~\cite{afifi2021msec} & BAID~\cite{lv2022backlitnet} & HDR-Eye~\cite{liu2020reverse} & HDR-Real~\cite{liu2020reverse} \\
\hline
Figure & Fig.~\ref{fig:GT_CMP_LSRW} & Fig.~\ref{fig:GT_CMP_UHD_LL} & Fig.~\ref{fig:GT_CMP_SICE} & Fig.~\ref{fig:GT_CMP_MSEC} & Fig.~\ref{fig:GT_CMP_BAID} & Fig.~\ref{fig:GT_CMP_HDREye} & Fig.~\ref{fig:GT_CMP_HDRReal} \\
\hline
\end{tabular}
\end{table*}

\section{More Cross-dataset Experiments}

\begin{figure*}
    \centering
    \makebox[0pt][r]{\raisebox{18.5cm}{\rotatebox{90}{Input}}\hspace{0.1cm}}%
    \makebox[0pt][r]{\raisebox{16cm}{\rotatebox{90}{GT}}\hspace{0.1cm}}%
    \makebox[0pt][r]{\raisebox{12.5cm}{\rotatebox{90}{CoLIE~\cite{chobola2024colie}}}\hspace{0.1cm}}%
    \makebox[0pt][r]{\raisebox{9.5cm}{\rotatebox{90}{FLLIE~\cite{wang2023fourllie}}}\hspace{0.1cm}}%
    \makebox[0pt][r]{\raisebox{7cm}{\rotatebox{90}{RF~\cite{cai2023retinexformer}}}\hspace{0.1cm}}%
    \makebox[0pt][r]{\raisebox{4cm}{\rotatebox{90}{SCI~\cite{ma2022sci}}}\hspace{0.1cm}}%
    \makebox[0pt][r]{\raisebox{1cm}{\rotatebox{90}{Ours}}\hspace{0.1cm}}%
    \includegraphics[width=\textwidth]{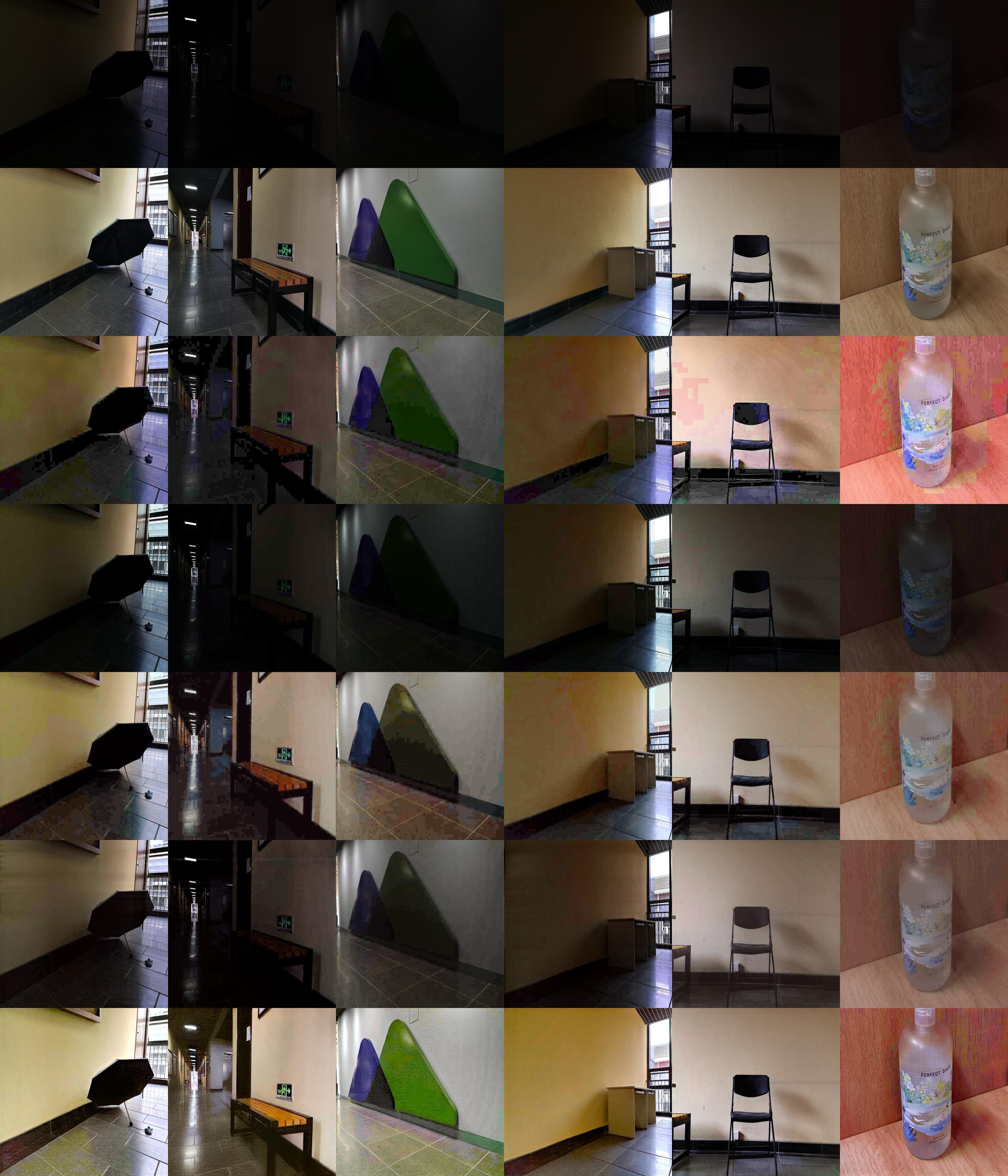}
    \caption{Visual comparison of cross-dataset validation on LSRW~\cite{hai2023r2rnet} dataset.}
    \label{fig:method_cmp_LSRW}
\end{figure*}

\begin{figure*}
    \centering
    \makebox[0pt][r]{\raisebox{18.5cm}{\rotatebox{90}{Input}}\hspace{0.1cm}}%
    \makebox[0pt][r]{\raisebox{16cm}{\rotatebox{90}{GT}}\hspace{0.1cm}}%
    \makebox[0pt][r]{\raisebox{12.5cm}{\rotatebox{90}{CoLIE~\cite{chobola2024colie}}}\hspace{0.1cm}}%
    \makebox[0pt][r]{\raisebox{9.5cm}{\rotatebox{90}{FLLIE~\cite{wang2023fourllie}}}\hspace{0.1cm}}%
    \makebox[0pt][r]{\raisebox{7cm}{\rotatebox{90}{RF~\cite{cai2023retinexformer}}}\hspace{0.1cm}}%
    \makebox[0pt][r]{\raisebox{4cm}{\rotatebox{90}{SCI~\cite{ma2022sci}}}\hspace{0.1cm}}%
    \makebox[0pt][r]{\raisebox{1cm}{\rotatebox{90}{Ours}}\hspace{0.1cm}}%
    \includegraphics[width=\textwidth]{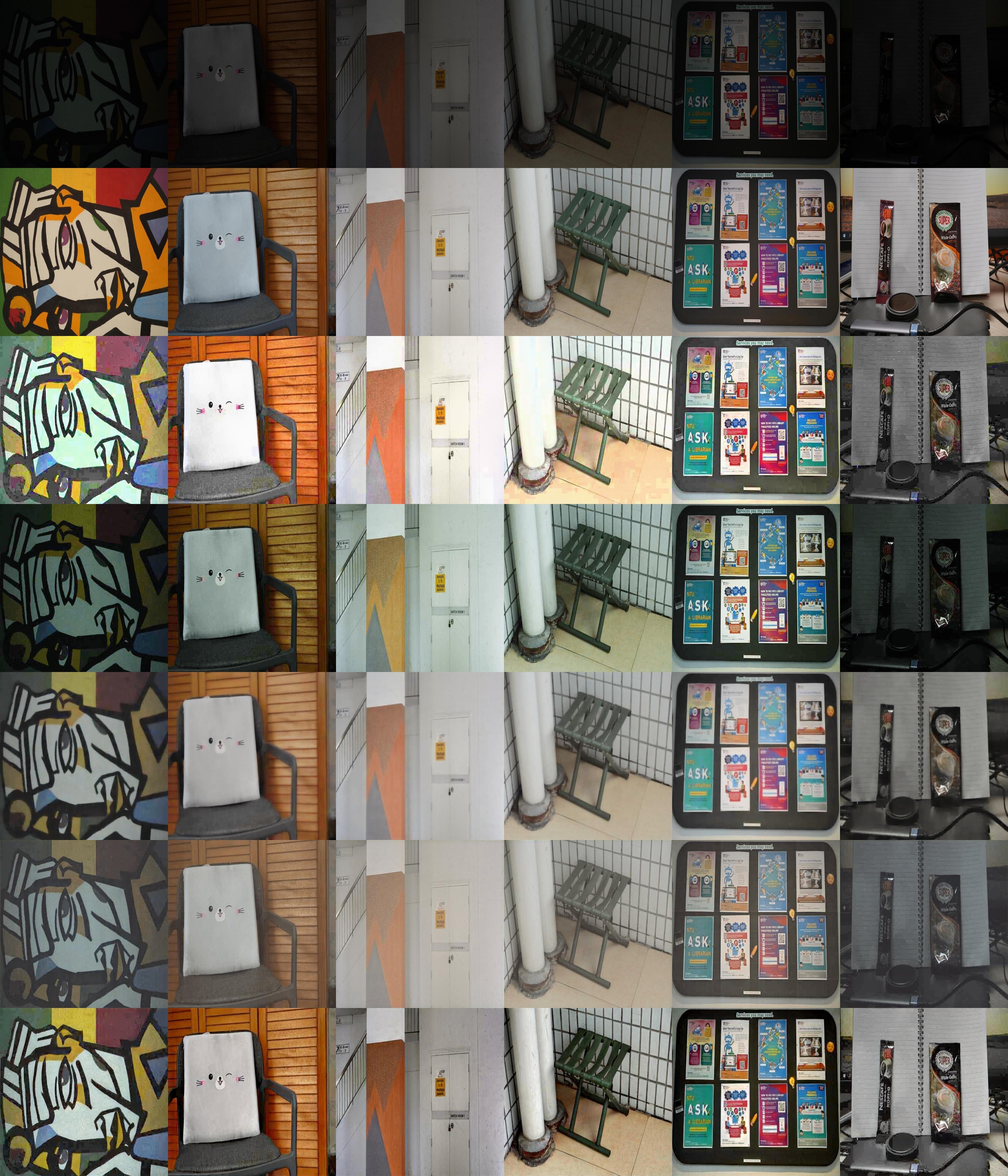}
    \caption{Visual comparison of cross-dataset validation on UHD-LL~\cite{Li2023uhdll} dataset.}
    \label{fig:method_cmp_UHDLL}
\end{figure*}

\begin{figure*}
    \centering
    \makebox[0pt][r]{\raisebox{18.5cm}{\rotatebox{90}{Input}}\hspace{0.1cm}}%
    \makebox[0pt][r]{\raisebox{16cm}{\rotatebox{90}{GT}}\hspace{0.1cm}}%
    \makebox[0pt][r]{\raisebox{12.5cm}{\rotatebox{90}{CoLIE~\cite{chobola2024colie}}}\hspace{0.1cm}}%
    \makebox[0pt][r]{\raisebox{9.5cm}{\rotatebox{90}{LCDP~\cite{wang2022lcdp}}}\hspace{0.1cm}}%
    \makebox[0pt][r]{\raisebox{7cm}{\rotatebox{90}{RF~\cite{cai2023retinexformer}}}\hspace{0.1cm}}%
    \makebox[0pt][r]{\raisebox{3.5cm}{\rotatebox{90}{CSEC~\cite{li_2024_cvpr_csec}}}\hspace{0.1cm}}%
    \makebox[0pt][r]{\raisebox{1cm}{\rotatebox{90}{Ours}}\hspace{0.1cm}}%
    \includegraphics[width=\textwidth]{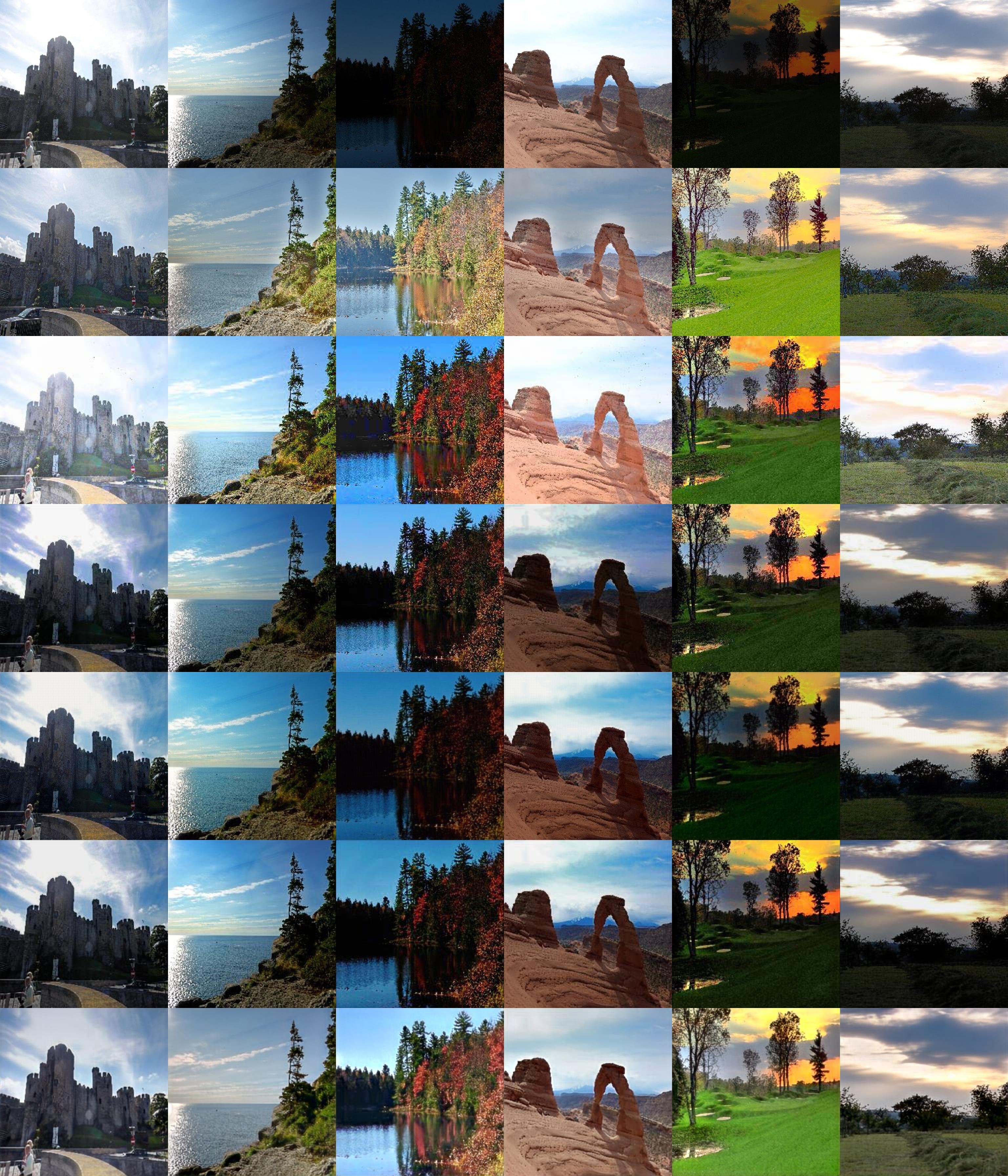}
    \caption{Visual comparison of cross-dataset validation on SICE~\cite{cai2018sice} dataset.}
    \label{fig:method_cmp_SICE}
\end{figure*}

\begin{figure*}
    \centering
    \makebox[0pt][r]{\raisebox{18.5cm}{\rotatebox{90}{Input}}\hspace{0.1cm}}%
    \makebox[0pt][r]{\raisebox{16cm}{\rotatebox{90}{GT}}\hspace{0.1cm}}%
    \makebox[0pt][r]{\raisebox{12.5cm}{\rotatebox{90}{CoLIE~\cite{chobola2024colie}}}\hspace{0.1cm}}%
    \makebox[0pt][r]{\raisebox{9.5cm}{\rotatebox{90}{LCDP~\cite{wang2022lcdp}}}\hspace{0.1cm}}%
    \makebox[0pt][r]{\raisebox{7cm}{\rotatebox{90}{RF~\cite{cai2023retinexformer}}}\hspace{0.1cm}}%
    \makebox[0pt][r]{\raisebox{3.5cm}{\rotatebox{90}{CSEC~\cite{li_2024_cvpr_csec}}}\hspace{0.1cm}}%
    \makebox[0pt][r]{\raisebox{1cm}{\rotatebox{90}{Ours}}\hspace{0.1cm}}%
    \includegraphics[width=\textwidth]{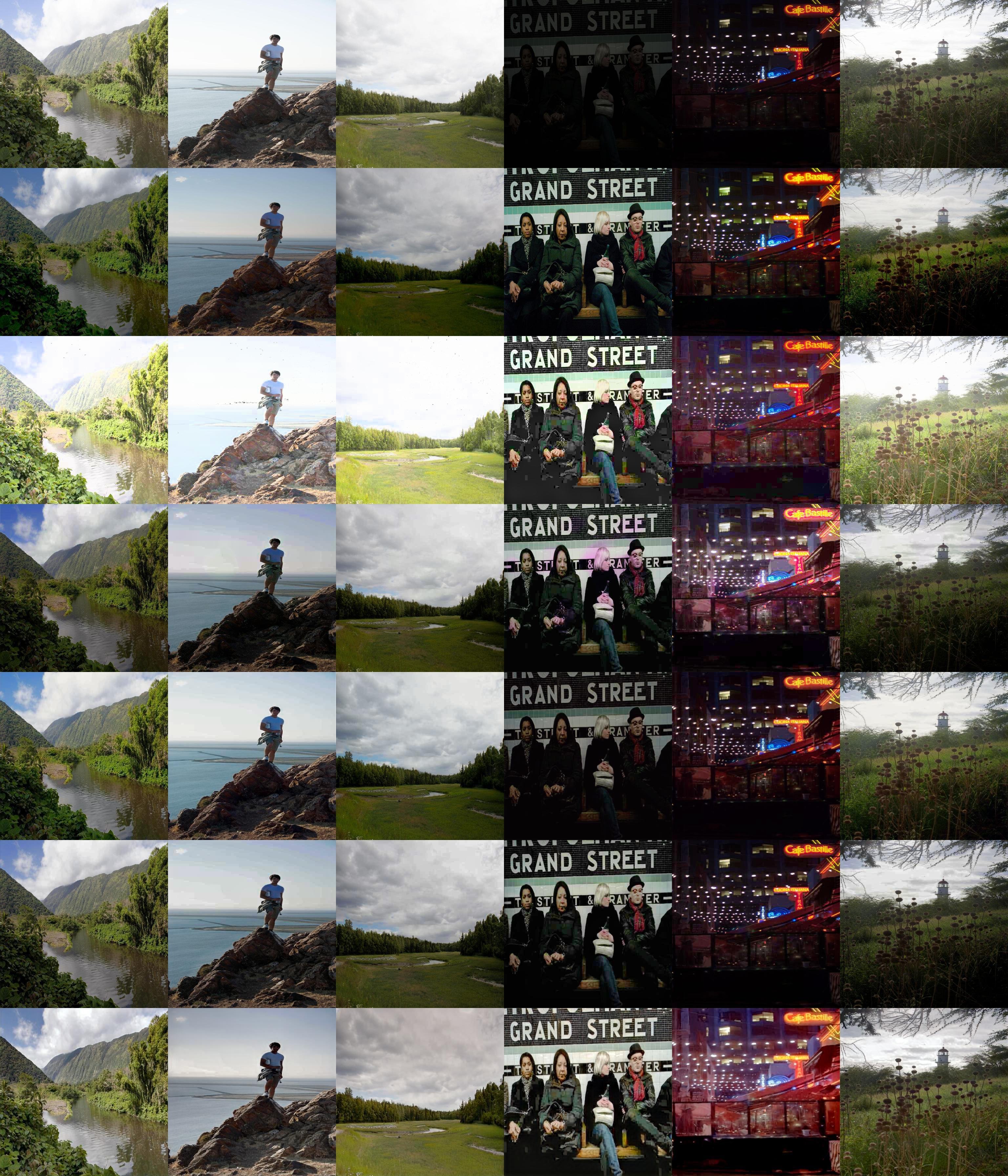}
    \caption{Visual comparison of cross-dataset validation on MSEC~\cite{afifi2021msec} dataset.}
    \label{fig:method_cmp_MSEC}
\end{figure*}

\begin{figure*}
    \centering
    \makebox[0pt][r]{\raisebox{18.5cm}{\rotatebox{90}{Input}}\hspace{0.1cm}}%
    \makebox[0pt][r]{\raisebox{16cm}{\rotatebox{90}{GT}}\hspace{0.1cm}}%
    \makebox[0pt][r]{\raisebox{12.5cm}{\rotatebox{90}{CoLIE~\cite{chobola2024colie}}}\hspace{0.1cm}}%
    \makebox[0pt][r]{\raisebox{9.5cm}{\rotatebox{90}{ZDCE++~\cite{Zero-DCE++}}}\hspace{0.1cm}}%
    \makebox[0pt][r]{\raisebox{7cm}{\rotatebox{90}{SCI~\cite{ma2022sci}}}\hspace{0.1cm}}%
    \makebox[0pt][r]{\raisebox{3.5cm}{\rotatebox{90}{CLIP-LIT~\cite{liang2023CLIP-LIT}}}\hspace{0.1cm}}%
    \makebox[0pt][r]{\raisebox{1cm}{\rotatebox{90}{Ours}}\hspace{0.1cm}}%
    \includegraphics[width=\textwidth]{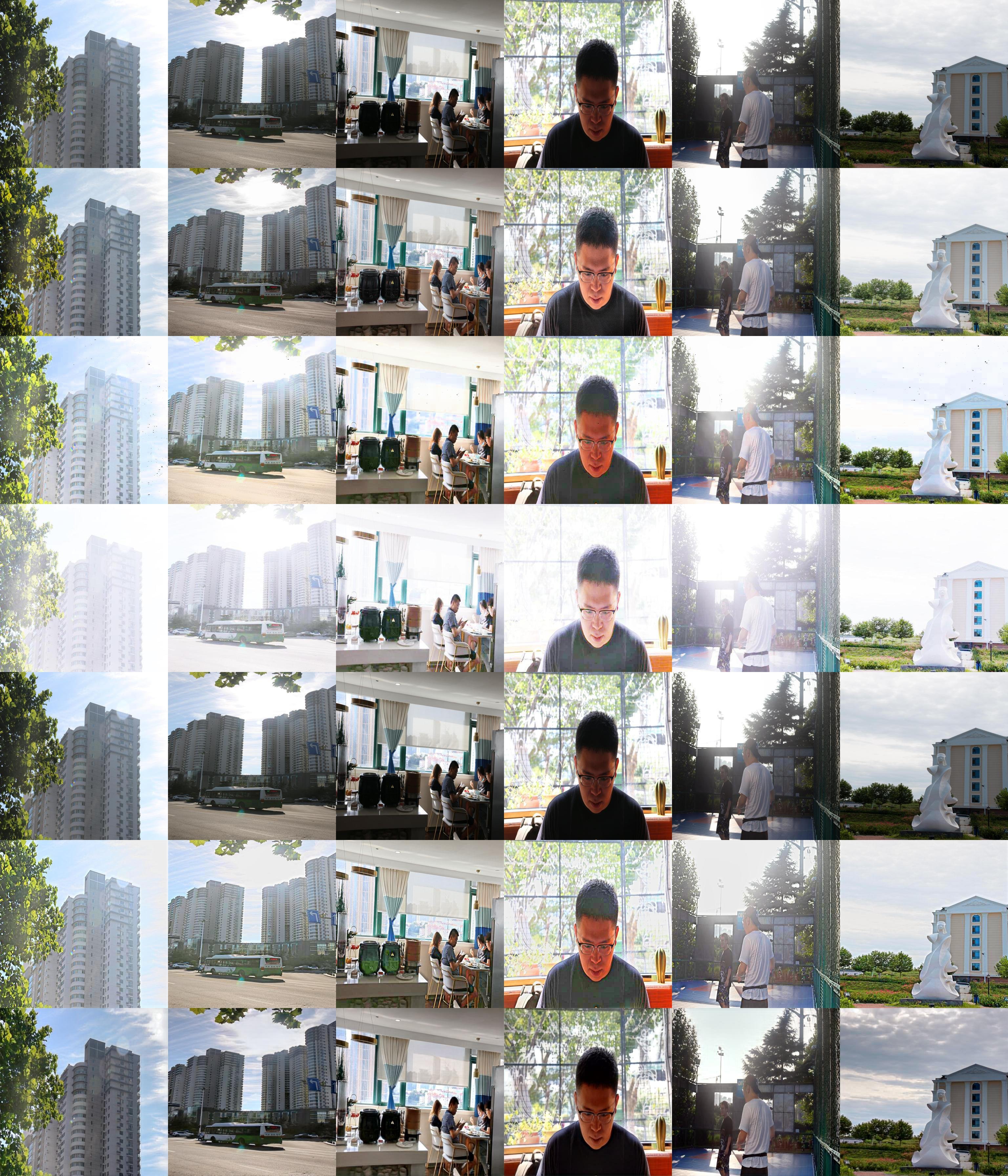}
    \caption{Visual comparison of cross-dataset validation on BAID~\cite{lv2022backlitnet} dataset.}
    \label{fig:method_cmp_BAID}
\end{figure*}

\begin{figure*}
    \centering
    \makebox[0pt][r]{\raisebox{15.5cm}{\rotatebox{90}{Input}}\hspace{0.1cm}}%
    \makebox[0pt][r]{\raisebox{12.5cm}{\rotatebox{90}{CoLIE~\cite{chobola2024colie}}}\hspace{0.1cm}}%
    \makebox[0pt][r]{\raisebox{10cm}{\rotatebox{90}{RF~\cite{cai2023retinexformer}}}\hspace{0.1cm}}%
    \makebox[0pt][r]{\raisebox{6.5cm}{\rotatebox{90}{CSEC~\cite{li_2024_cvpr_csec}}}\hspace{0.1cm}}%
    \makebox[0pt][r]{\raisebox{3.5cm}{\rotatebox{90}{CLIP-LIT~\cite{liang2023CLIP-LIT}}}\hspace{0.1cm}}%
    \makebox[0pt][r]{\raisebox{1cm}{\rotatebox{90}{Ours}}\hspace{0.1cm}}%
    \includegraphics[width=1\linewidth]{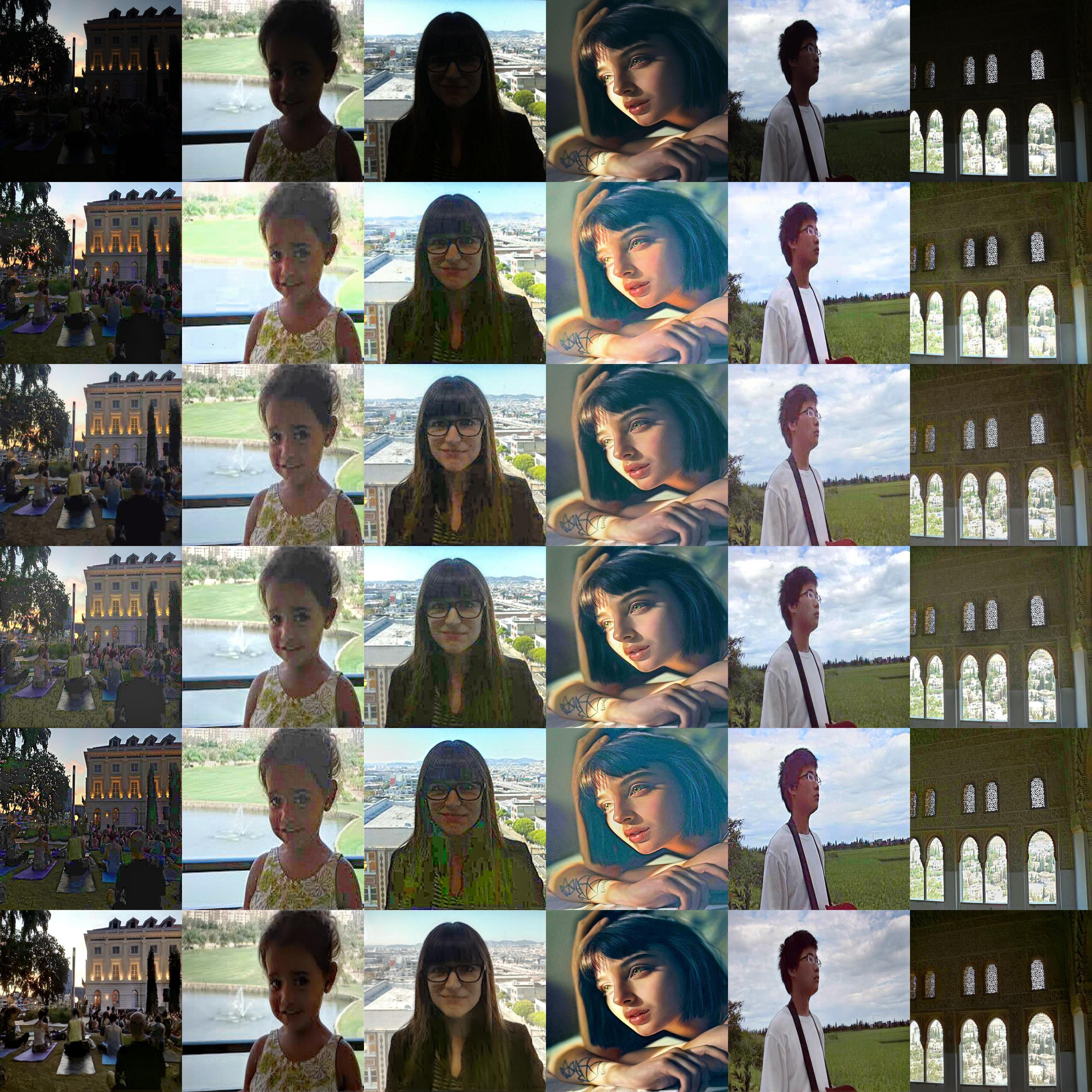}
    \caption{Visual comparison of cross-dataset validation on Backlit~\cite{liang2023CLIP-LIT} dataset.}
    \label{fig:method_cmp_Backlit}
\end{figure*}

\begin{figure*}
    \centering
    \makebox[0pt][r]{\raisebox{18.5cm}{\rotatebox{90}{Input}}\hspace{0.1cm}}%
    \makebox[0pt][r]{\raisebox{16cm}{\rotatebox{90}{GT}}\hspace{0.1cm}}%
    \makebox[0pt][r]{\raisebox{12.5cm}{\rotatebox{90}{CoLIE~\cite{chobola2024colie}}}\hspace{0.1cm}}%
    \makebox[0pt][r]{\raisebox{9.5cm}{\rotatebox{90}{LCDP~\cite{wang2022lcdp}}}\hspace{0.1cm}}%
    \makebox[0pt][r]{\raisebox{7cm}{\rotatebox{90}{RF~\cite{cai2023retinexformer}}}\hspace{0.1cm}}%
    \makebox[0pt][r]{\raisebox{3.5cm}{\rotatebox{90}{CSEC~\cite{li_2024_cvpr_csec}}}\hspace{0.1cm}}%
    \makebox[0pt][r]{\raisebox{1cm}{\rotatebox{90}{Ours}}\hspace{0.1cm}}%
    \includegraphics[width=\textwidth]{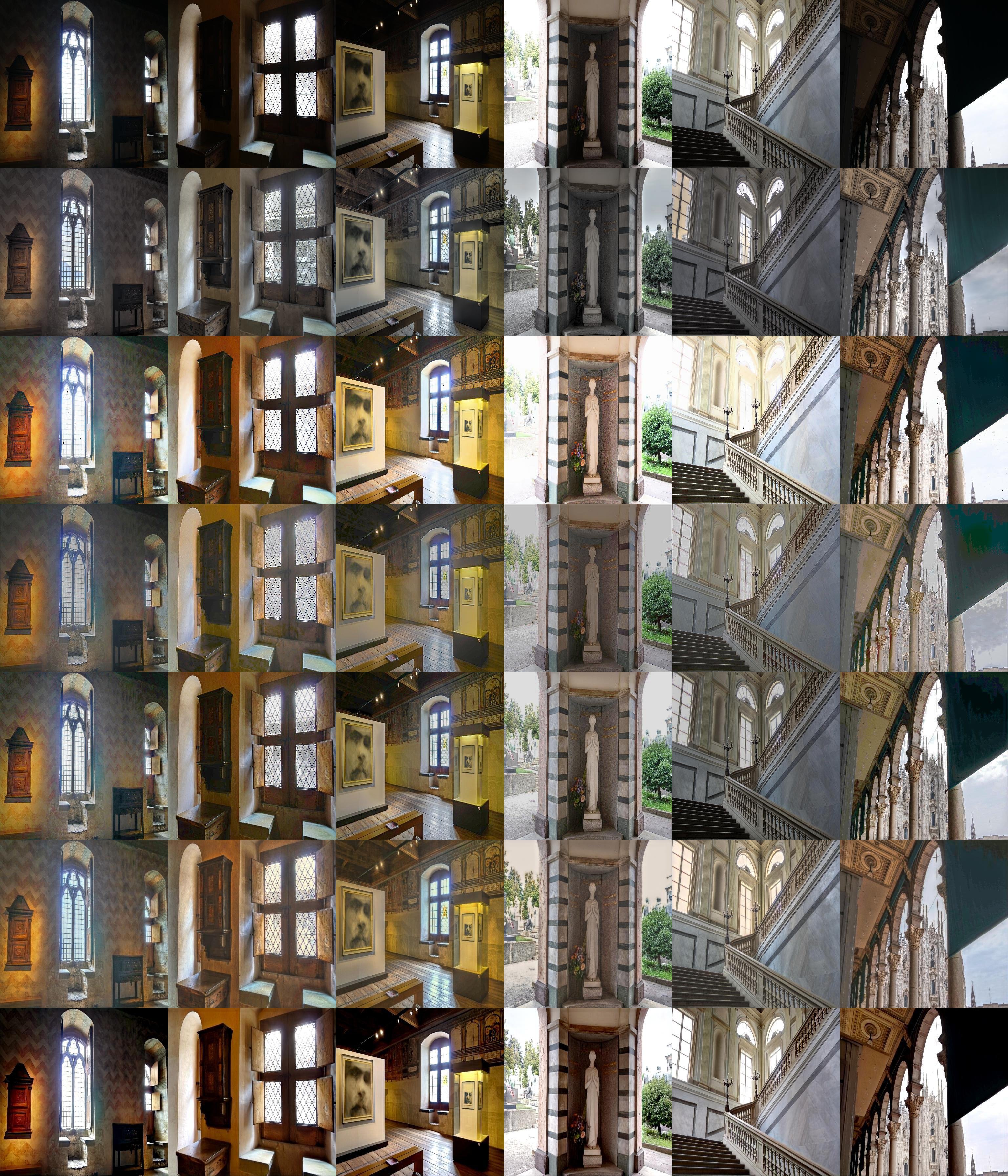}
    \caption{Visual comparison of cross-dataset validation on HDR-Eye~\cite{liu2020reverse} dataset.}
    \label{fig:method_cmp_HDREye}
\end{figure*}

\begin{figure*}
    \centering
    \makebox[0pt][r]{\raisebox{18.5cm}{\rotatebox{90}{Input}}\hspace{0.1cm}}%
    \makebox[0pt][r]{\raisebox{16cm}{\rotatebox{90}{GT}}\hspace{0.1cm}}%
    \makebox[0pt][r]{\raisebox{12.5cm}{\rotatebox{90}{CoLIE~\cite{chobola2024colie}}}\hspace{0.1cm}}%
    \makebox[0pt][r]{\raisebox{9.5cm}{\rotatebox{90}{LCDP~\cite{wang2022lcdp}}}\hspace{0.1cm}}%
    \makebox[0pt][r]{\raisebox{7cm}{\rotatebox{90}{RF~\cite{cai2023retinexformer}}}\hspace{0.1cm}}%
    \makebox[0pt][r]{\raisebox{3.5cm}{\rotatebox{90}{CSEC~\cite{li_2024_cvpr_csec}}}\hspace{0.1cm}}%
    \makebox[0pt][r]{\raisebox{1cm}{\rotatebox{90}{Ours}}\hspace{0.1cm}}%
    \includegraphics[width=\textwidth]{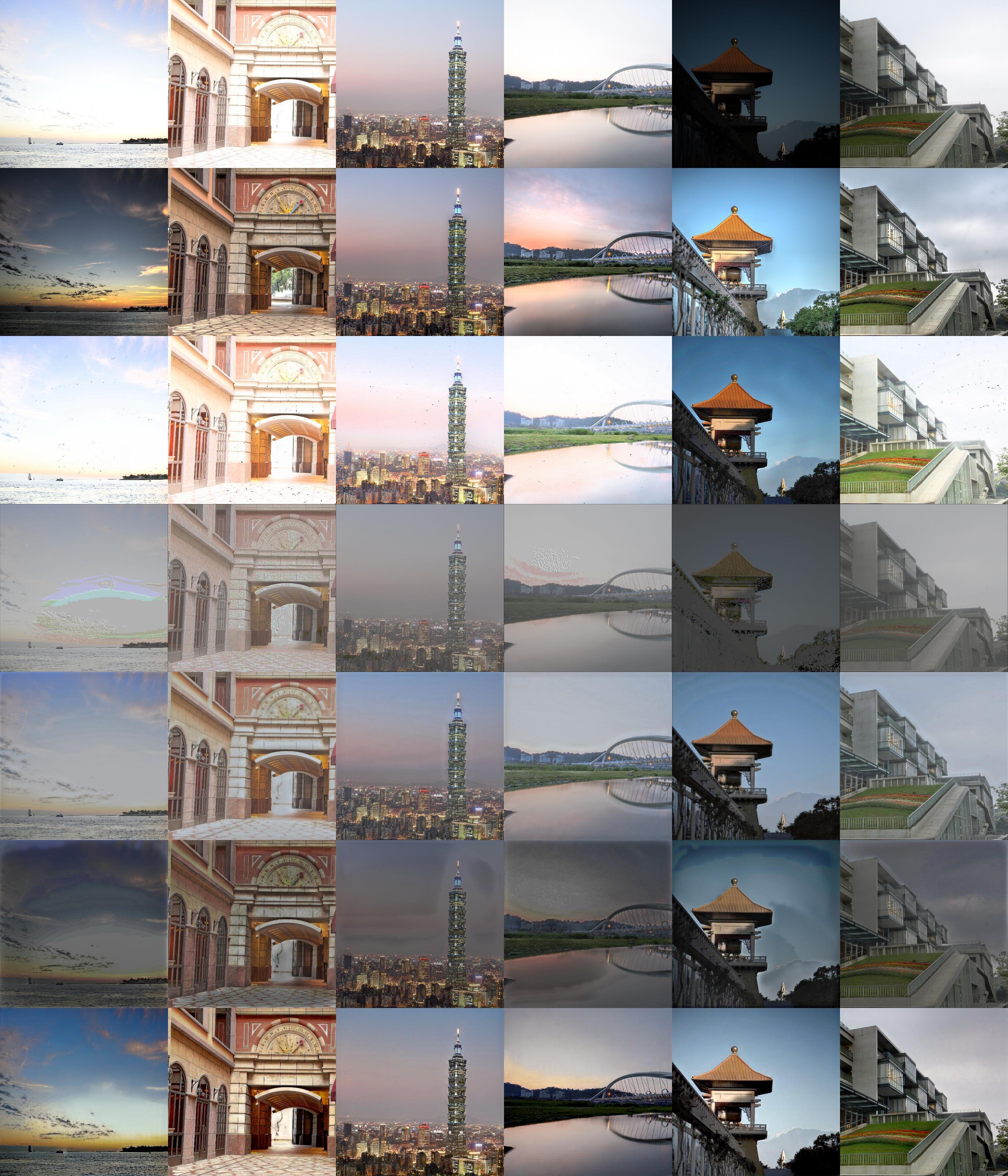}
    \caption{Visual comparison of cross-dataset validation on HDR-Real~\cite{liu2020reverse} dataset.}
    \label{fig:method_cmp_HDRReal}
\end{figure*}

In Section~4.3 of the main paper, we presented cross-dataset evaluation results for four tasks: Low-Light Image Enhancement (LLIE), Exposure Correction (EC), Backlit Image Enhancement (BIE), and Low Dynamic Range to High Dynamic Range Transformation (L2HT). In this supplementary material, we provide additional qualitative results to further support our findings. These visual comparisons are summarized in Table~\ref{tab:Generalization Performance} for the reader's convenience.
From the extended visual comparisons for the LLIE and EC tasks shown in Figs.~\ref{fig:method_cmp_LSRW}, \ref{fig:method_cmp_UHDLL}, \ref{fig:method_cmp_SICE}, and \ref{fig:method_cmp_MSEC}, we observe consistent conclusions with those drawn in the main paper. Specifically, UNICE demonstrates significantly better generalization performance, primarily due to its independence from human annotations.
For the BIE and L2HT tasks, visual comparisons are provided in Figs.~\ref{fig:method_cmp_BAID}, \ref{fig:method_cmp_Backlit}, \ref{fig:method_cmp_HDREye}, and \ref{fig:method_cmp_HDRReal}. In Figure~\ref{fig:method_cmp_BAID}, it is evident that many existing methods suffer from overexposure, particularly in background regions. In contrast, UNICE effectively maintains balanced exposure across different areas of the image.
Similarly, Figs.~\ref{fig:method_cmp_Backlit} and \ref{fig:method_cmp_HDREye} reveal that several methods tend to excessively brighten the image, resulting in the loss of true blacks and a reduced dynamic range. UNICE, however, enhances overall brightness while preserving dark regions, thereby achieving a superior dynamic range.
The advantages of our method become even more pronounced in Fig.~\ref{fig:method_cmp_HDRReal}. This is attributed to the relatively small scale of the HDR-Eye~\cite{liu2020reverse} training dataset used in this evaluation. While other methods exhibit noticeably degraded performance under these conditions, our approach consistently delivers significantly better results.

\begin{table*}[t]
\centering
\caption{Summary of visual results on cross-dataset experiments across different tasks.}
\label{tab:Generalization Performance}
\begin{tabular}{c|l|l|c}
\hline
Task & Train Set & Test Set & Figure \\ 
\hline
\multirow{2}{*}{LLIE} 
& UHD-LL~\cite{Li2023uhdll} & LSRW~\cite{hai2023r2rnet} & Fig.~\ref{fig:method_cmp_LSRW} \\ 
& LSRW~\cite{hai2023r2rnet} & UHD-LL~\cite{Li2023uhdll} & Fig.~\ref{fig:method_cmp_UHDLL} \\ 
\hline
\multirow{2}{*}{EC} 
& MSEC~\cite{afifi2021msec} & SICE~\cite{cai2018sice} & Fig.~\ref{fig:method_cmp_SICE} \\ 
& SICE~\cite{cai2018sice} & MSEC~\cite{afifi2021msec} & Fig.~\ref{fig:method_cmp_MSEC} \\ 
\hline
\multirow{2}{*}{BIE} 
& Backlit~\cite{hai2023r2rnet} & BAID~\cite{lv2022backlitnet} & Fig.~\ref{fig:method_cmp_BAID} \\ 
& BAID~\cite{lv2022backlitnet} & Backlit~\cite{liang2023CLIP-LIT} & Fig.~\ref{fig:method_cmp_Backlit} \\ 
\hline
\multirow{2}{*}{L2HT} 
& HDR-Real~\cite{liu2020reverse} & HDR-Eye~\cite{liu2020reverse} & Fig.~\ref{fig:method_cmp_HDREye} \\ 
& HDR-Eye~\cite{liu2020reverse} & HDR-Real~\cite{liu2020reverse} & Fig.~\ref{fig:method_cmp_HDRReal} \\ 
\hline
\end{tabular}
\end{table*}

\begin{figure*}[htbp]
    \centering
    \makebox[0pt][r]{\raisebox{6.5cm}{\rotatebox{90}{Input}}\hspace{0.1cm}}%
    \makebox[0pt][r]{\raisebox{4cm}{\rotatebox{90}{GT}}\hspace{0.1cm}}%
    \makebox[0pt][r]{\raisebox{1cm}{\rotatebox{90}{Ours}}\hspace{0.1cm}}%
    \begin{subfigure}{0.16\textwidth}
        \includegraphics[width=\textwidth]{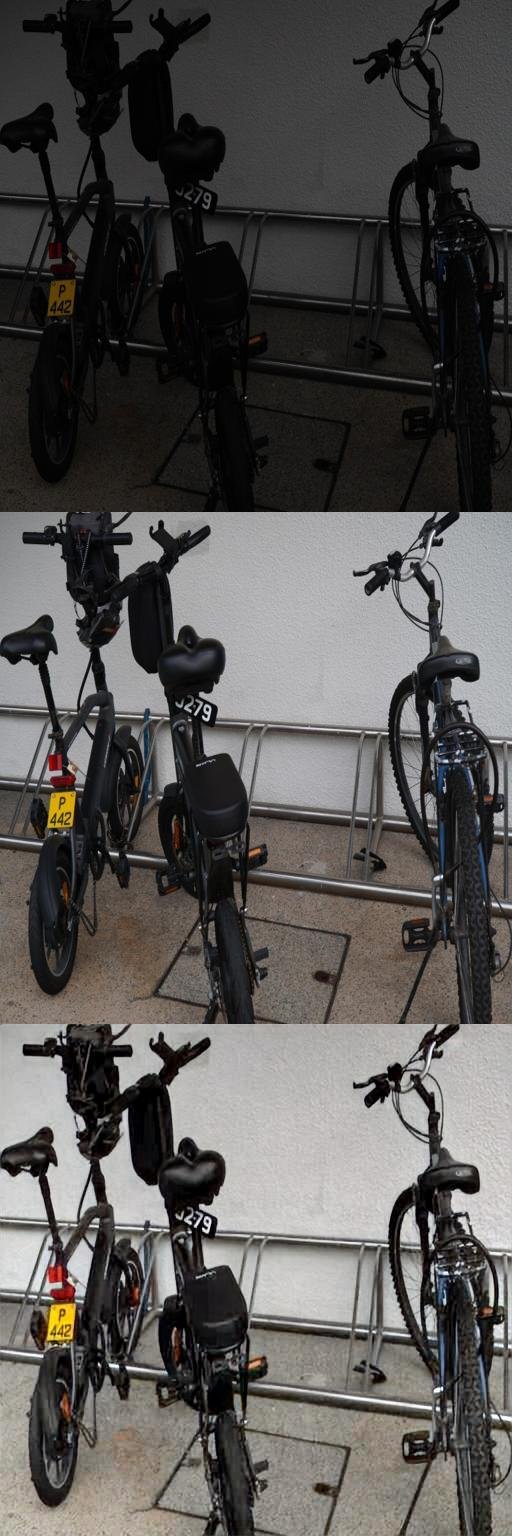}
    \end{subfigure}
    \begin{subfigure}{0.16\textwidth}
        \includegraphics[width=\textwidth]{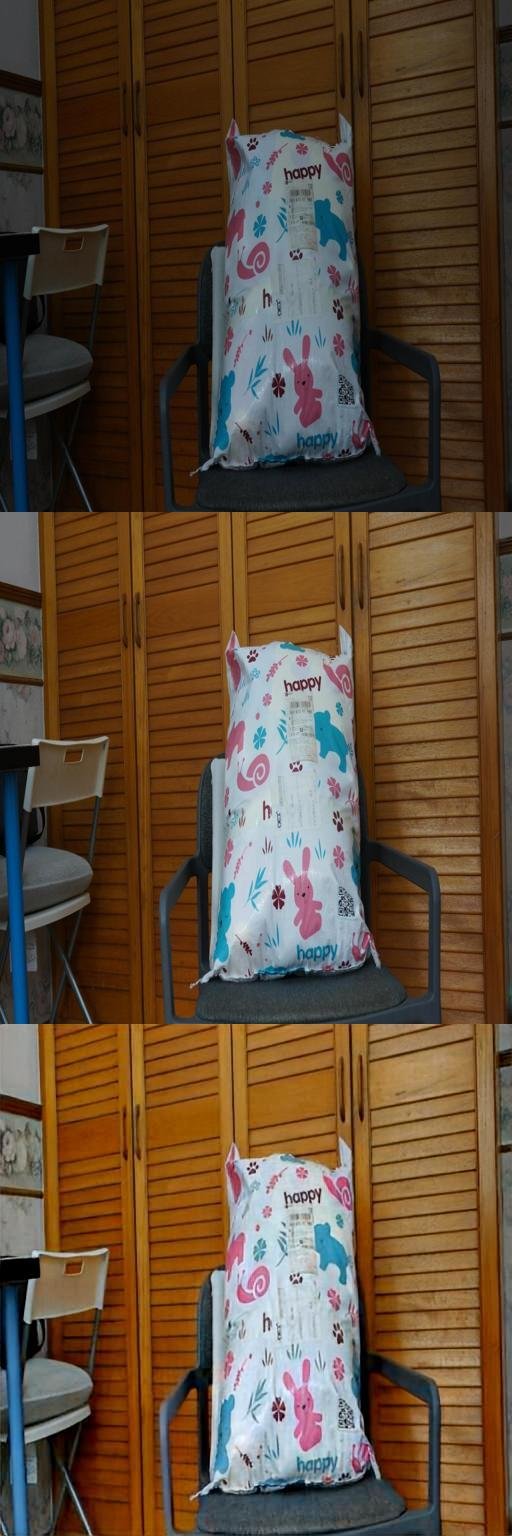}
    \end{subfigure}
    \begin{subfigure}{0.16\textwidth}
        \includegraphics[width=\textwidth]{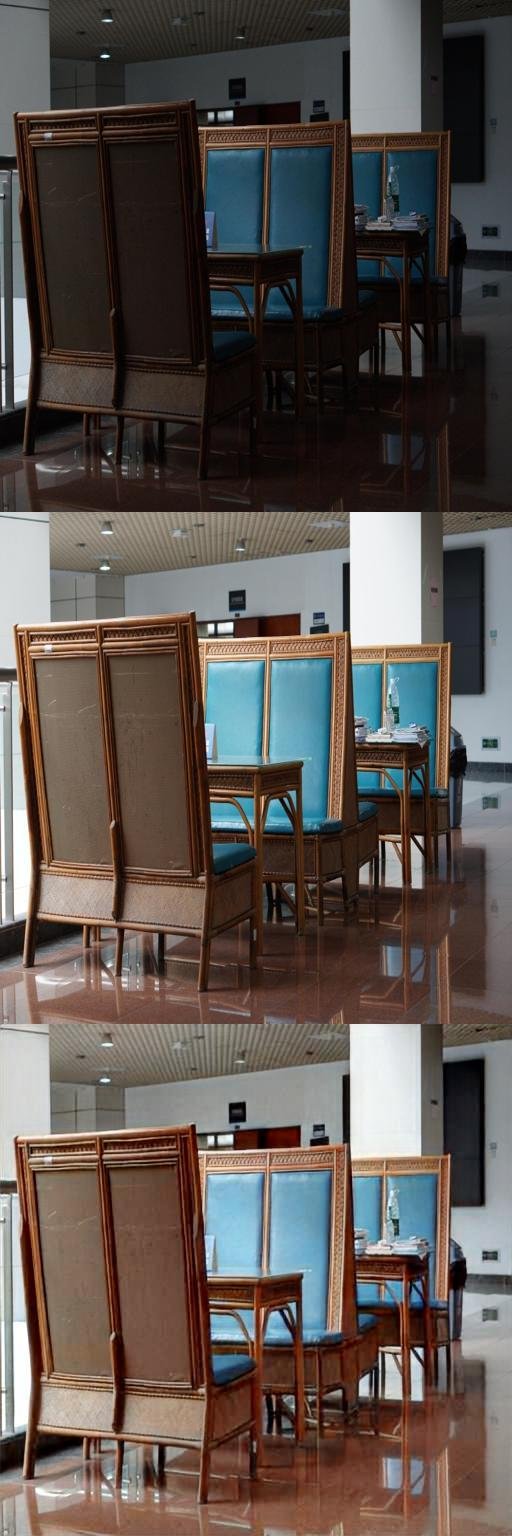}
    \end{subfigure}
    \begin{subfigure}{0.16\textwidth}
        \includegraphics[width=\textwidth]{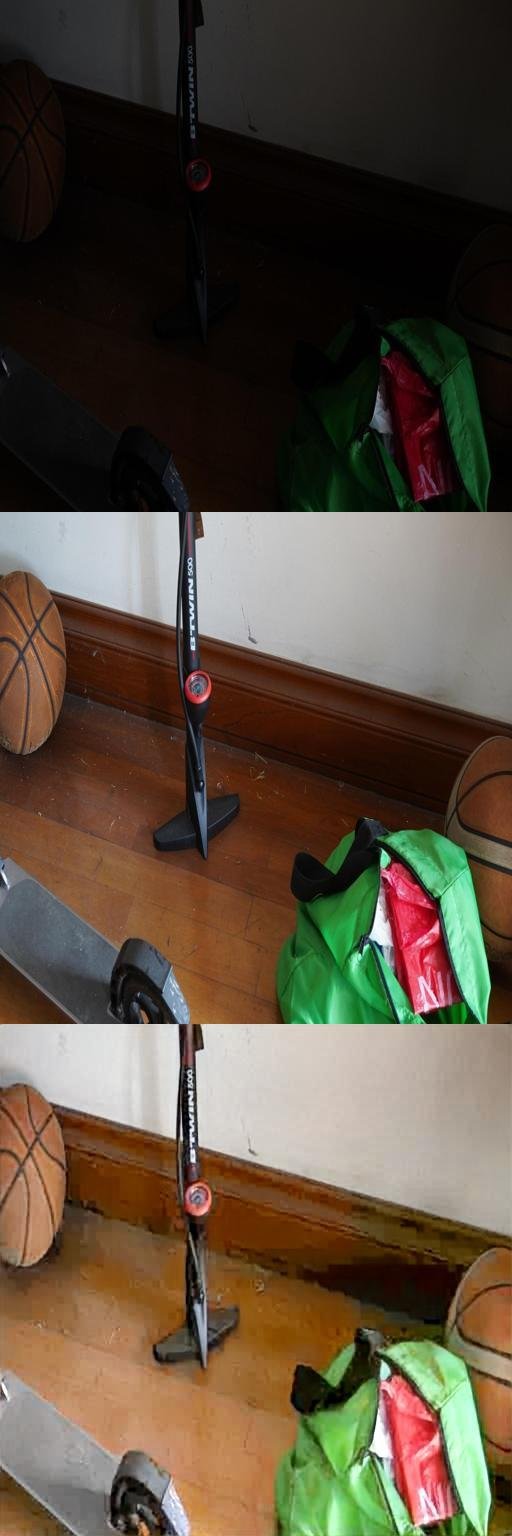}
    \end{subfigure}
    \begin{subfigure}{0.16\textwidth}
        \includegraphics[width=\textwidth]{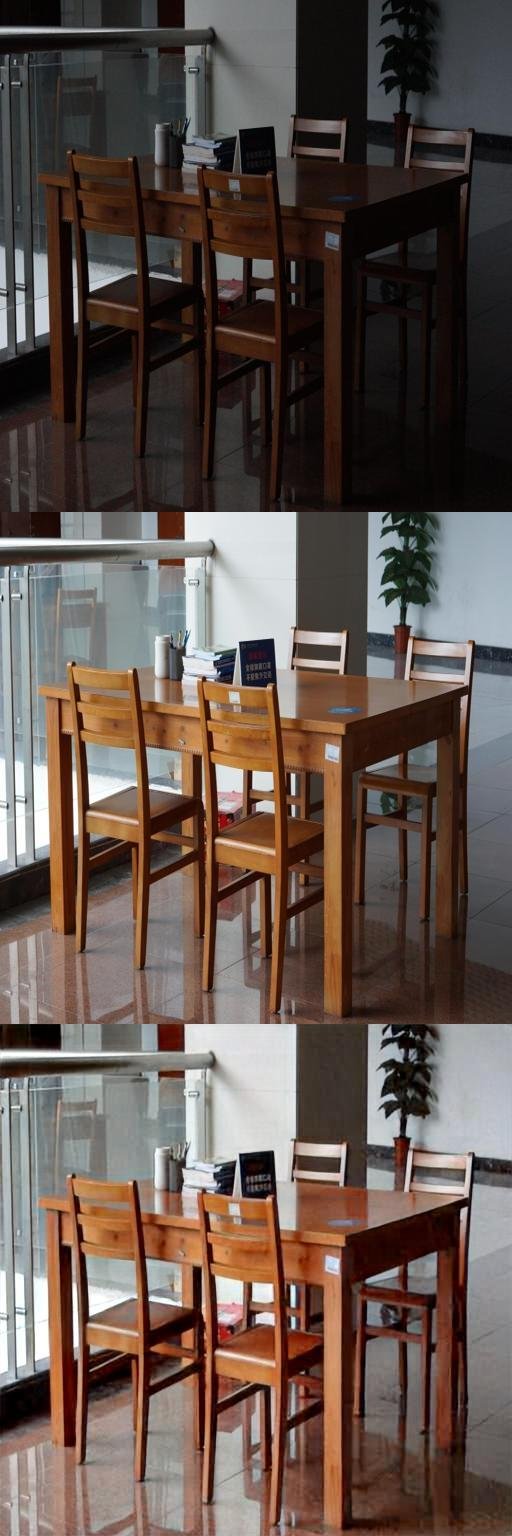}
    \end{subfigure}
    \begin{subfigure}{0.16\textwidth}
        \includegraphics[width=\textwidth]{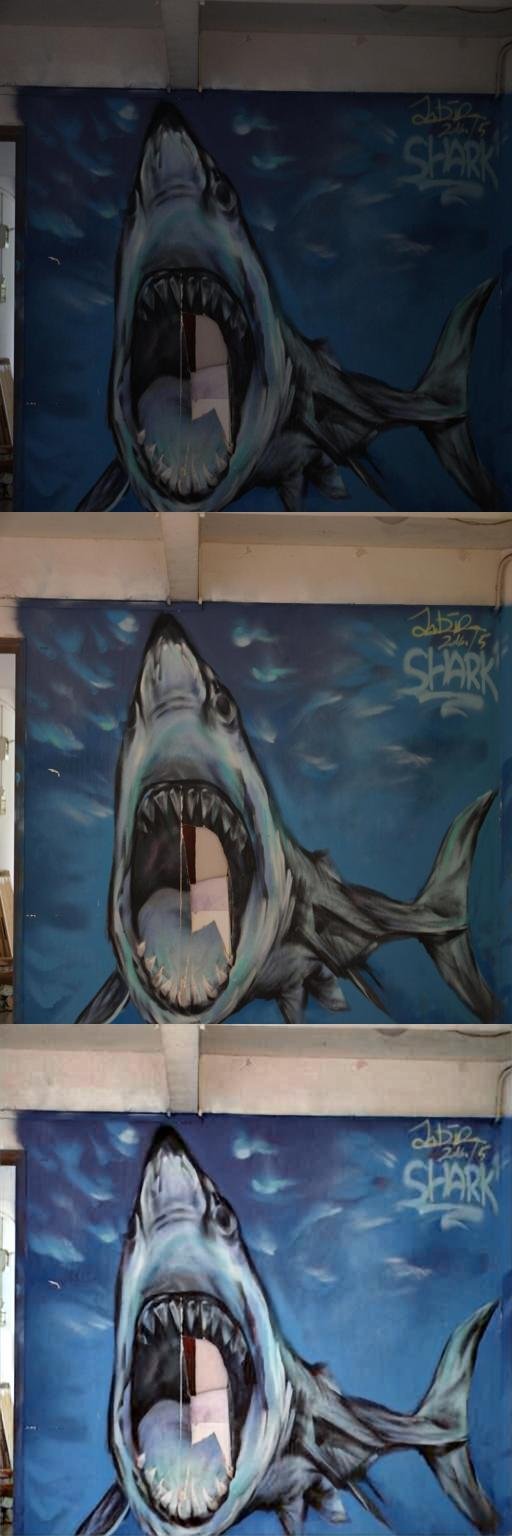}
    \end{subfigure}
    \caption{Visual comparison with GT on the UHD-LL~\cite{Li2023uhdll} dataset. }
    \label{fig:GT_CMP_UHD_LL}
\end{figure*}

\begin{figure*}[htbp]
    \centering
    \makebox[0pt][r]{\raisebox{6.5cm}{\rotatebox{90}{Input}}\hspace{0.1cm}}%
    \makebox[0pt][r]{\raisebox{4cm}{\rotatebox{90}{GT}}\hspace{0.1cm}}%
    \makebox[0pt][r]{\raisebox{1cm}{\rotatebox{90}{Ours}}\hspace{0.1cm}}%
    \begin{subfigure}{0.16\textwidth}
        \includegraphics[width=\textwidth]{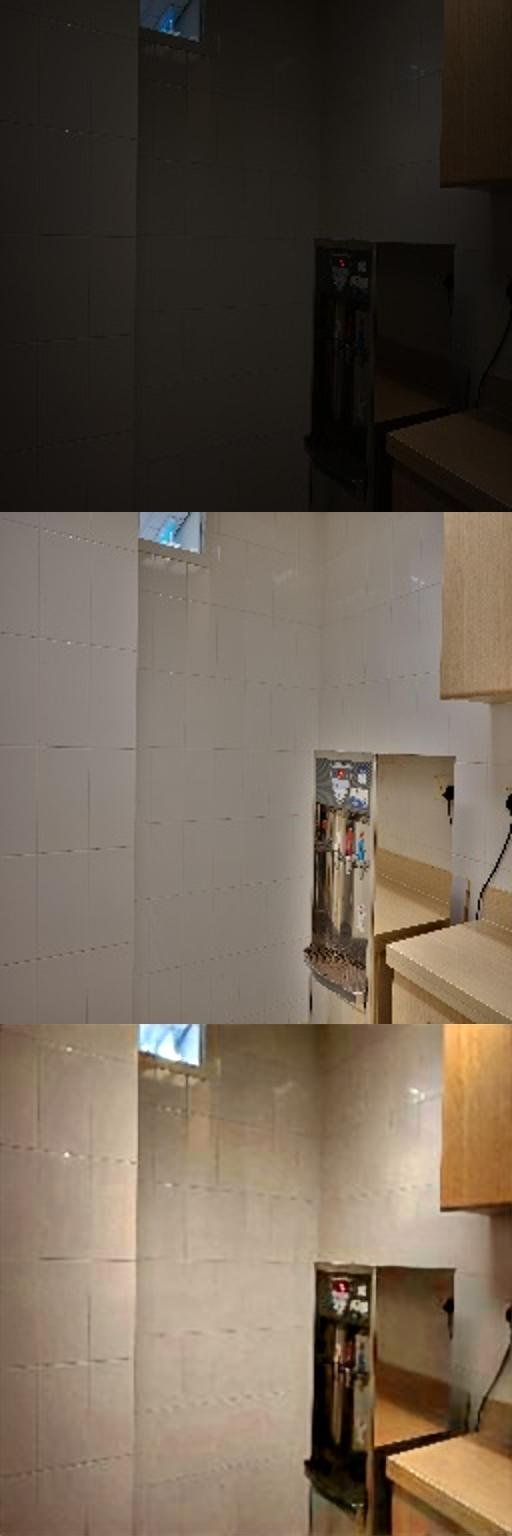}
    \end{subfigure}
    \begin{subfigure}{0.16\textwidth}
        \includegraphics[width=\textwidth]{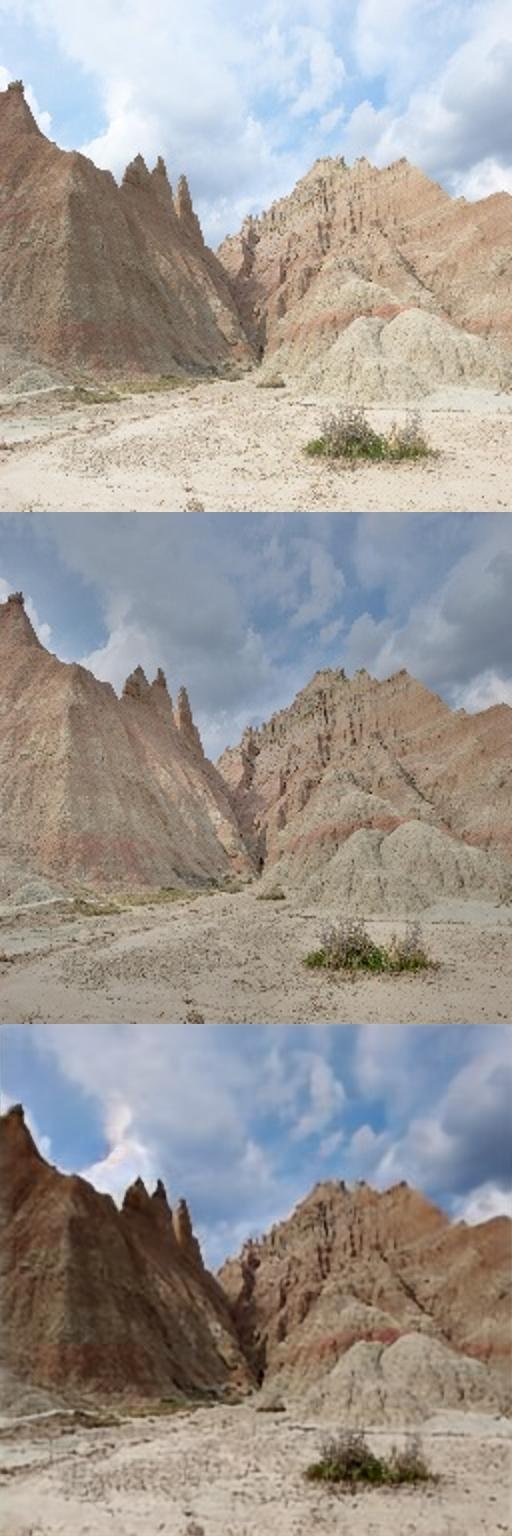}
    \end{subfigure}
    \begin{subfigure}{0.16\textwidth}
        \includegraphics[width=\textwidth]{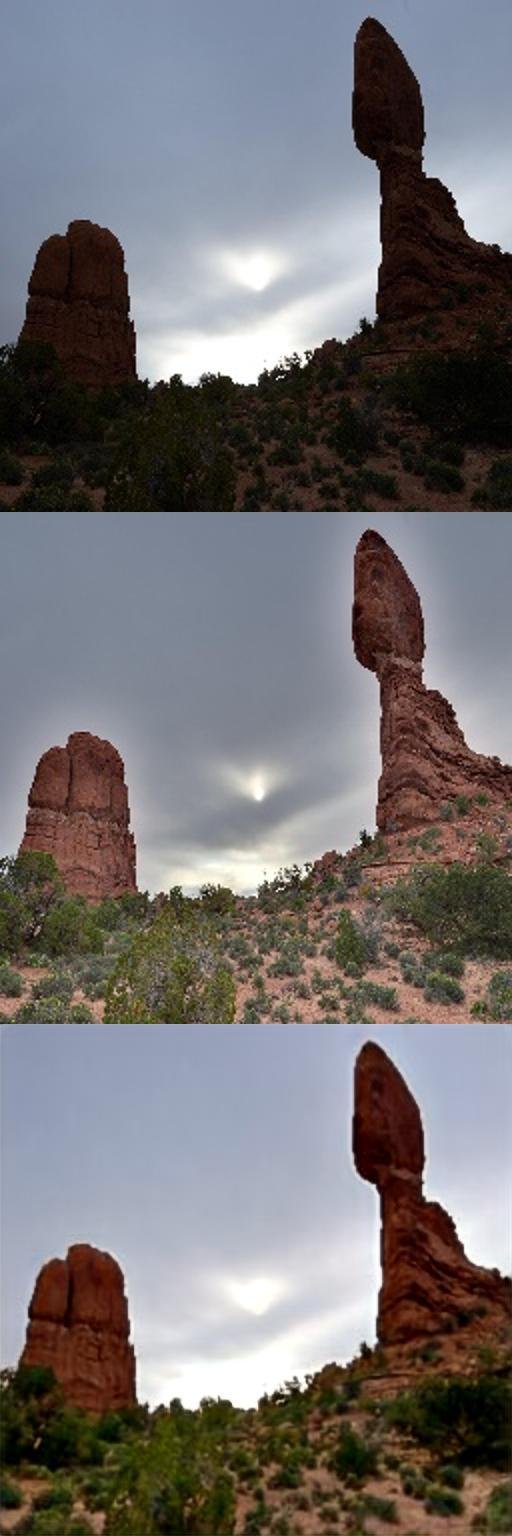}
    \end{subfigure}
    \begin{subfigure}{0.16\textwidth}
        \includegraphics[width=\textwidth]{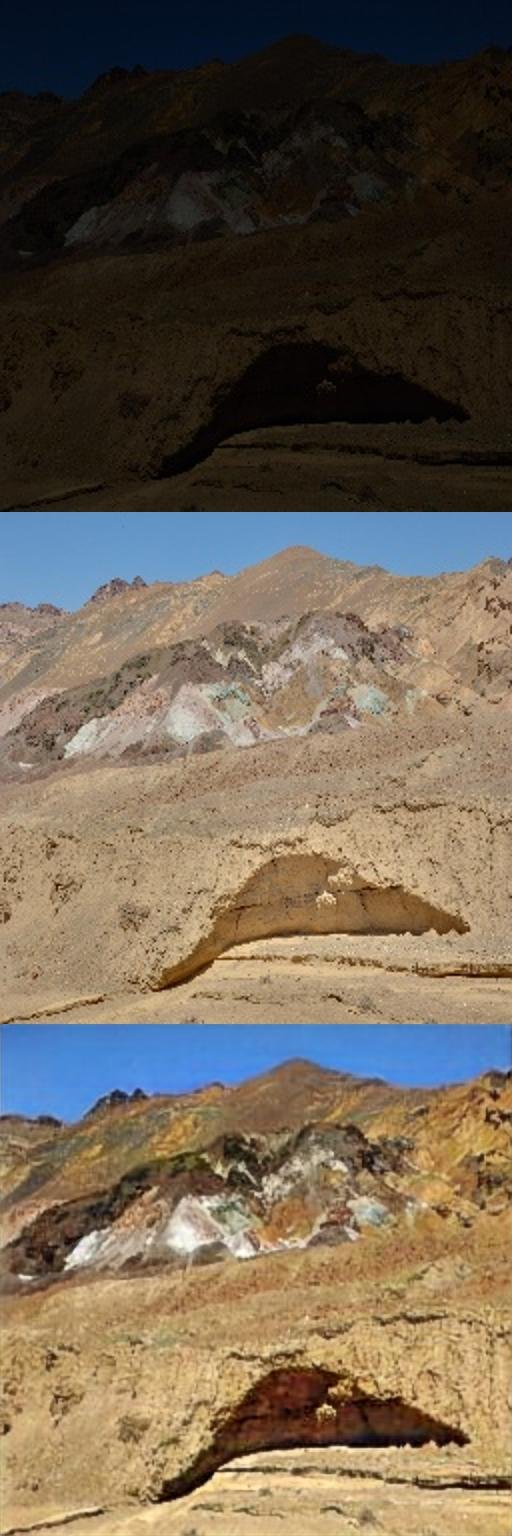}
    \end{subfigure}
    \begin{subfigure}{0.16\textwidth}
        \includegraphics[width=\textwidth]{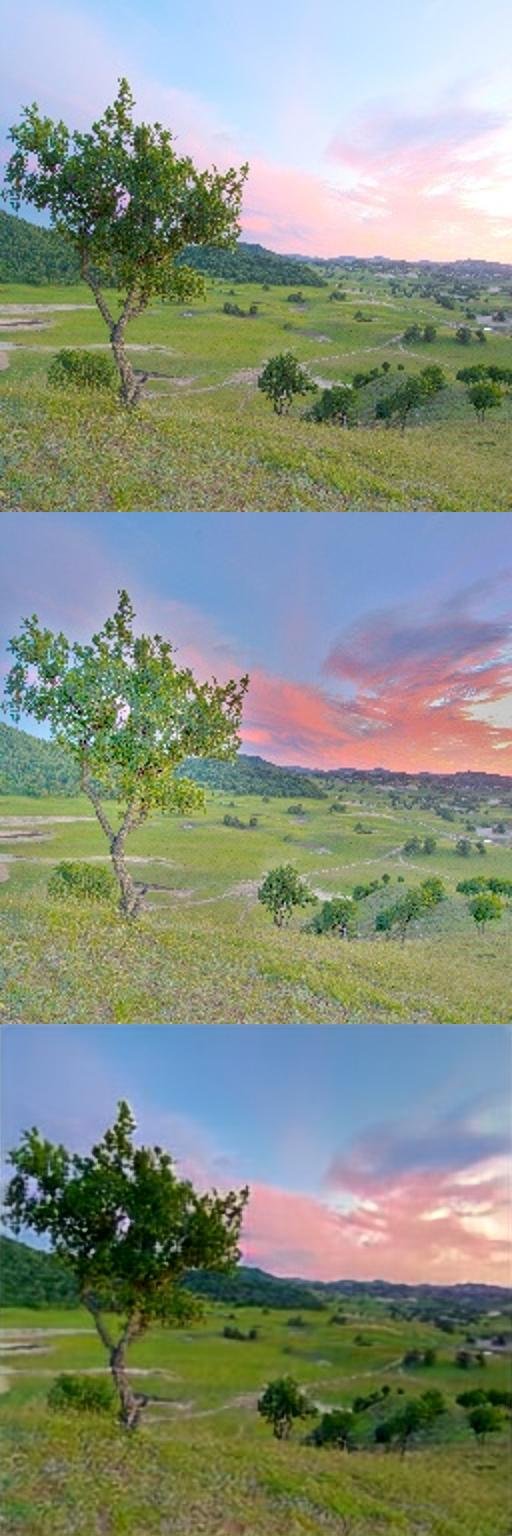}
    \end{subfigure}
    \begin{subfigure}{0.16\textwidth}
        \includegraphics[width=\textwidth]{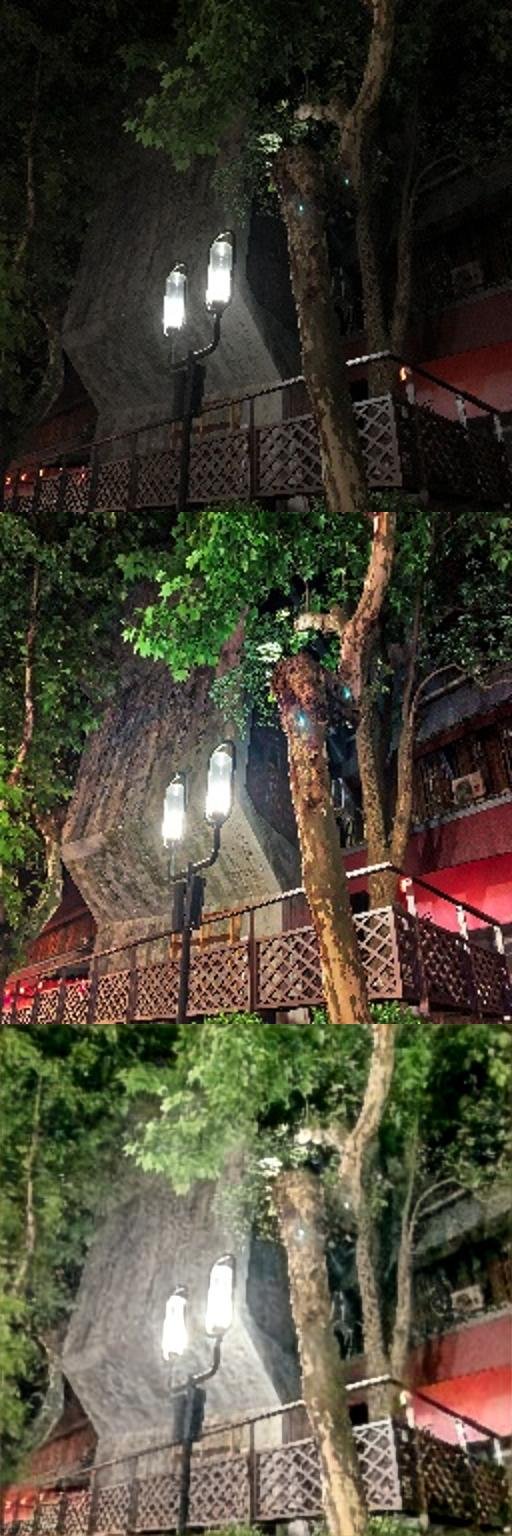}
    \end{subfigure}
    \caption{Visual comparison with GT on the SICE~\cite{cai2018sice} dataset. }
    \label{fig:GT_CMP_SICE}
\end{figure*}

\begin{figure*}[htbp]
    \centering
    \makebox[0pt][r]{\raisebox{6.5cm}{\rotatebox{90}{Input}}\hspace{0.1cm}}%
    \makebox[0pt][r]{\raisebox{4cm}{\rotatebox{90}{GT}}\hspace{0.1cm}}%
    \makebox[0pt][r]{\raisebox{1cm}{\rotatebox{90}{Ours}}\hspace{0.1cm}}%
    \begin{subfigure}{0.16\textwidth}
        \includegraphics[width=\textwidth]{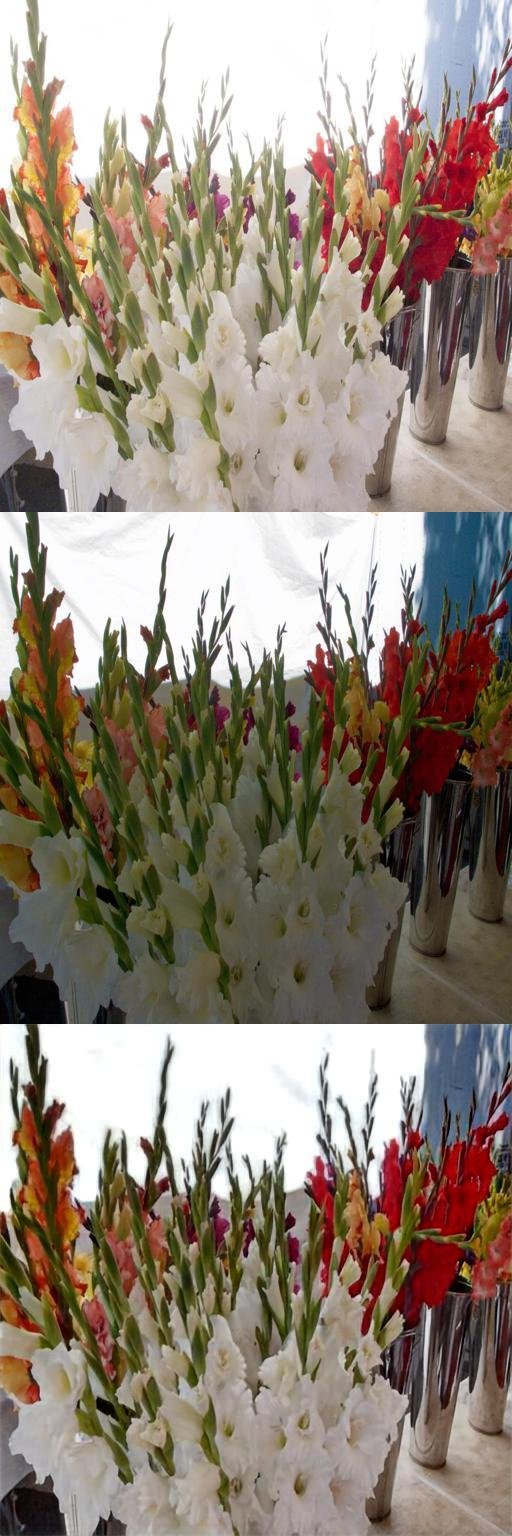}
    \end{subfigure}
    \begin{subfigure}{0.16\textwidth}
        \includegraphics[width=\textwidth]{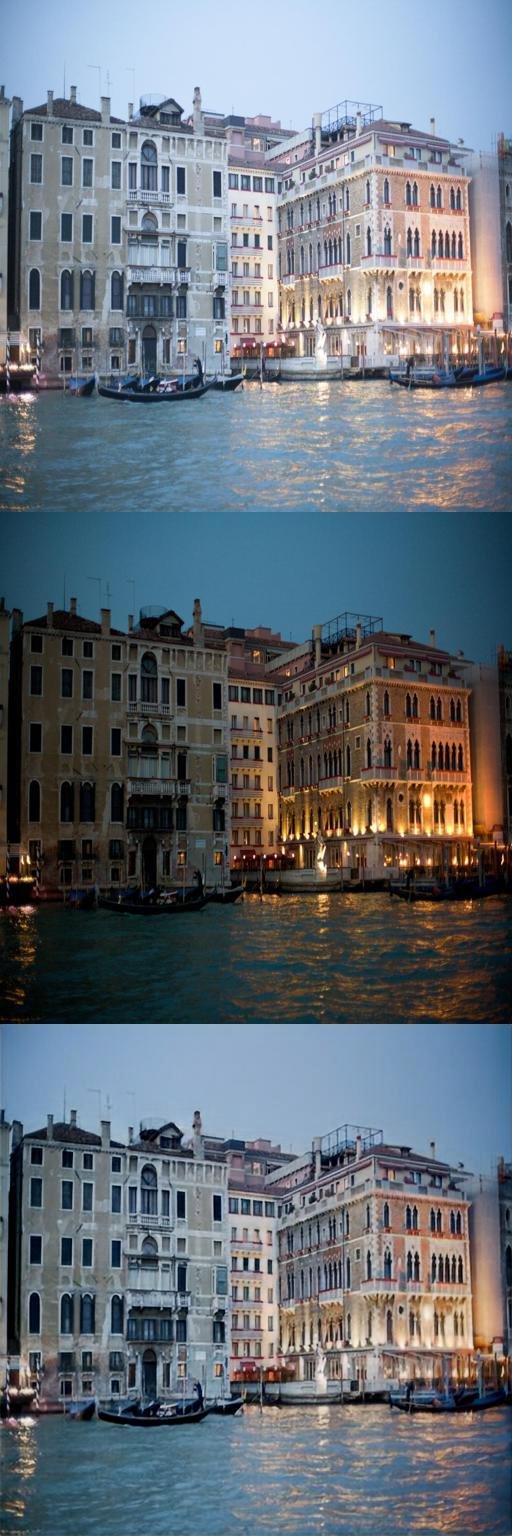}
    \end{subfigure}
    \begin{subfigure}{0.16\textwidth}
        \includegraphics[width=\textwidth]{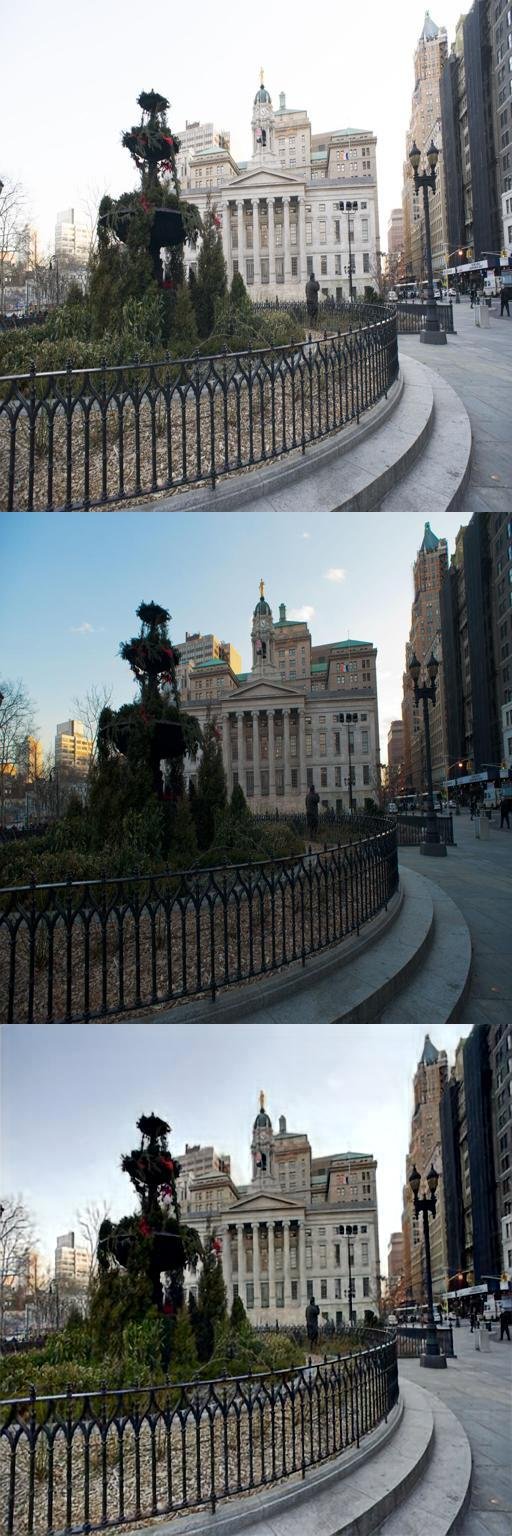}
    \end{subfigure}
    \begin{subfigure}{0.16\textwidth}
        \includegraphics[width=\textwidth]{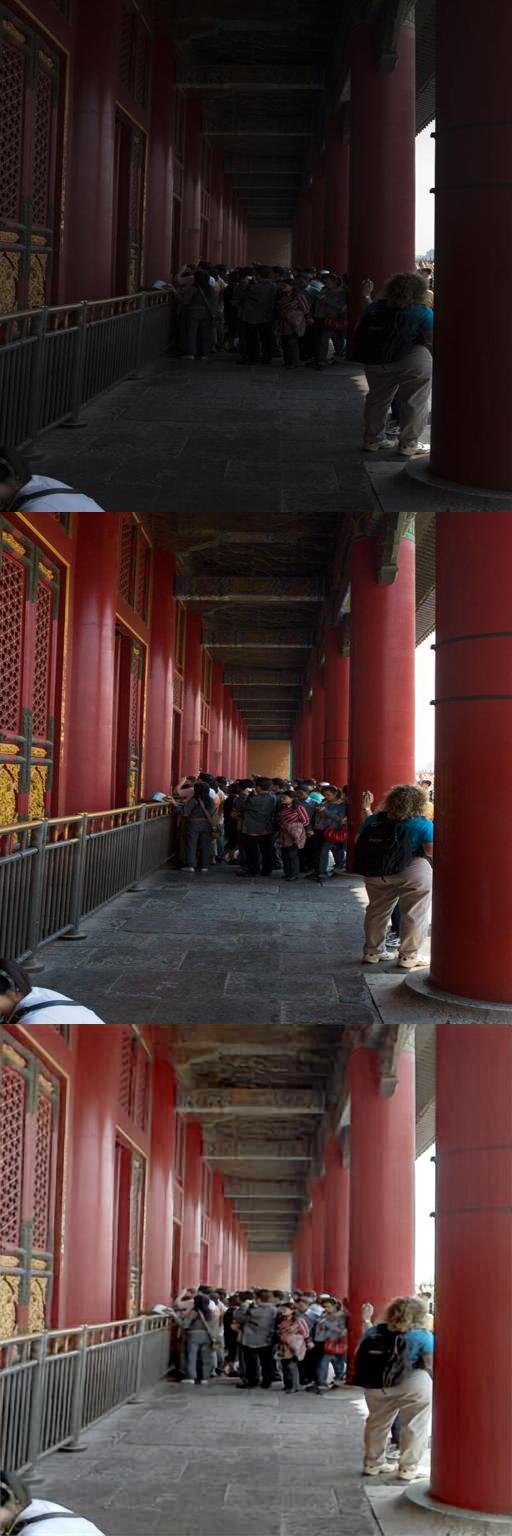}
    \end{subfigure}
    \begin{subfigure}{0.16\textwidth}
        \includegraphics[width=\textwidth]{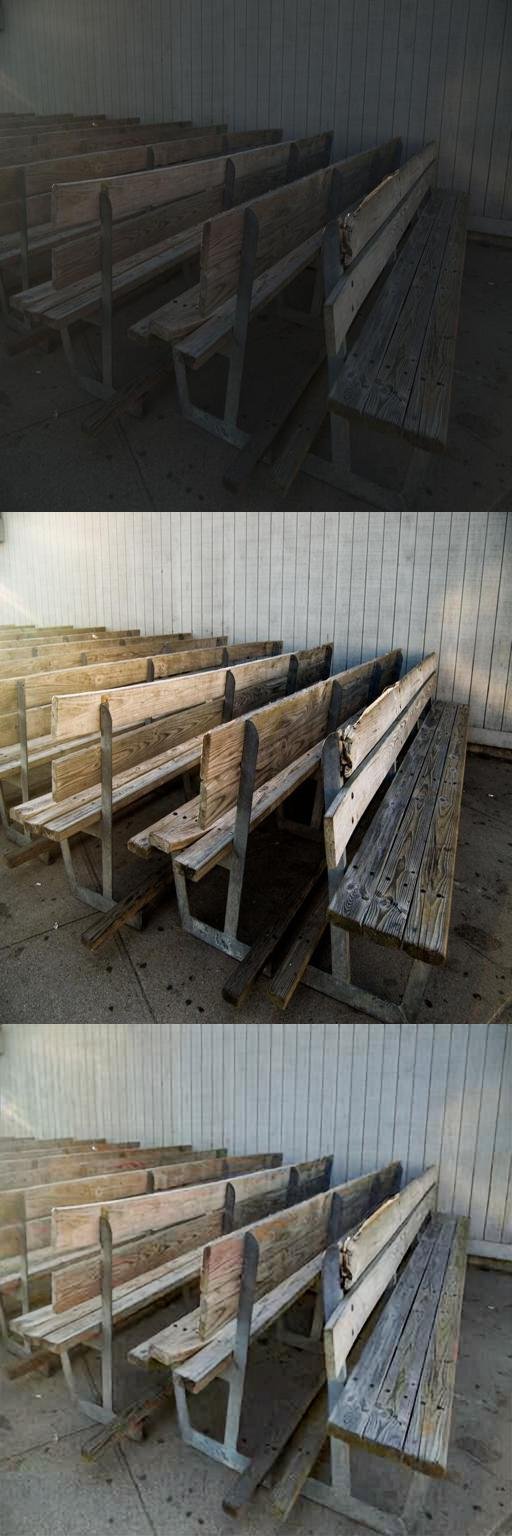}
    \end{subfigure}
    \begin{subfigure}{0.16\textwidth}
        \includegraphics[width=\textwidth]{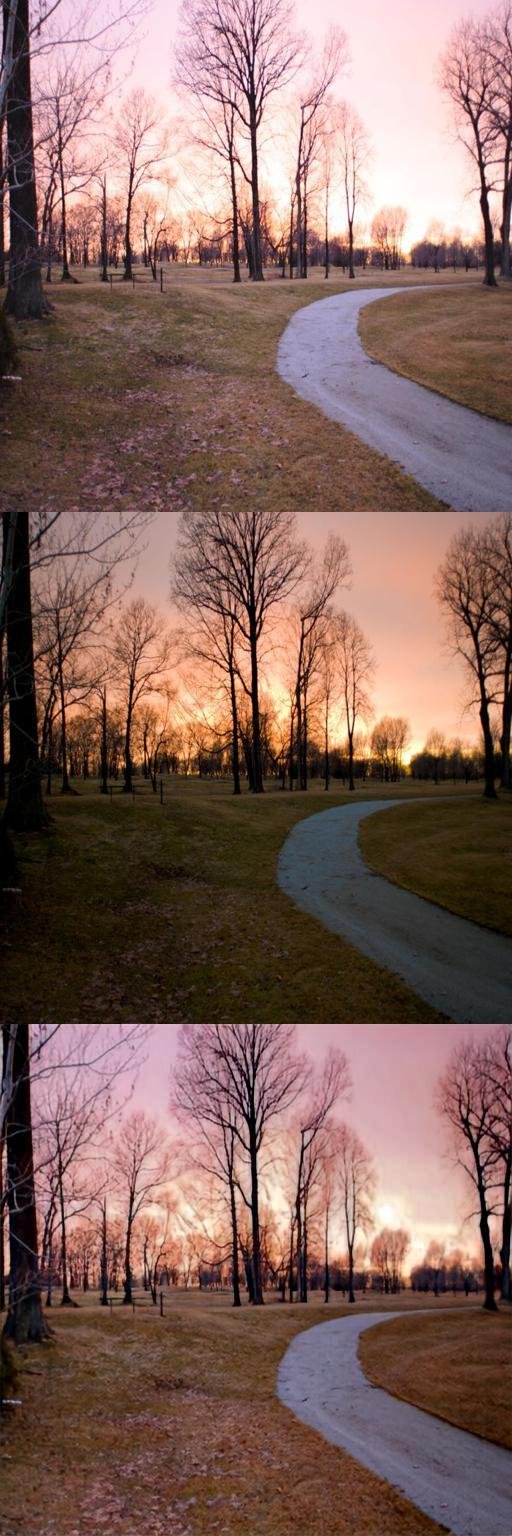}
    \end{subfigure}
    \caption{Visual comparison with GT on the MSEC~\cite{afifi2021msec} dataset. }
    \label{fig:GT_CMP_MSEC}
\end{figure*}

\begin{figure*}[htbp]
    \centering
    \makebox[0pt][r]{\raisebox{6.5cm}{\rotatebox{90}{Input}}\hspace{0.1cm}}%
    \makebox[0pt][r]{\raisebox{4cm}{\rotatebox{90}{GT}}\hspace{0.1cm}}%
    \makebox[0pt][r]{\raisebox{1cm}{\rotatebox{90}{Ours}}\hspace{0.1cm}}%
    \begin{subfigure}{0.16\textwidth}
        \includegraphics[width=\textwidth]{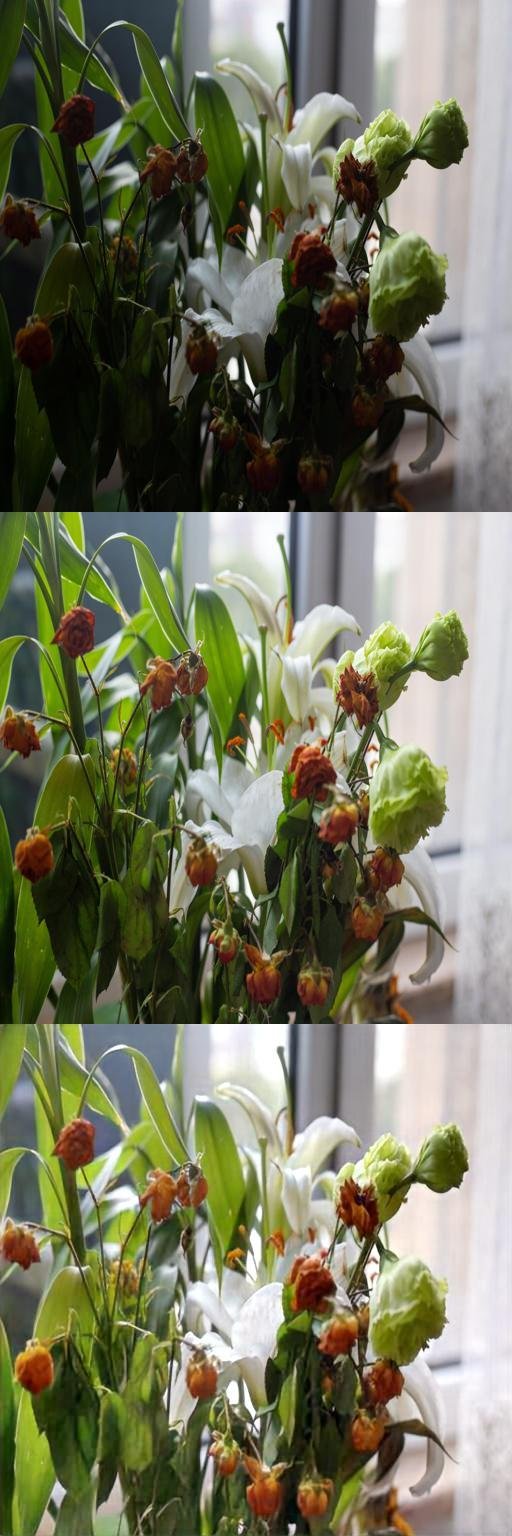}
    \end{subfigure}
    \begin{subfigure}{0.16\textwidth}
        \includegraphics[width=\textwidth]{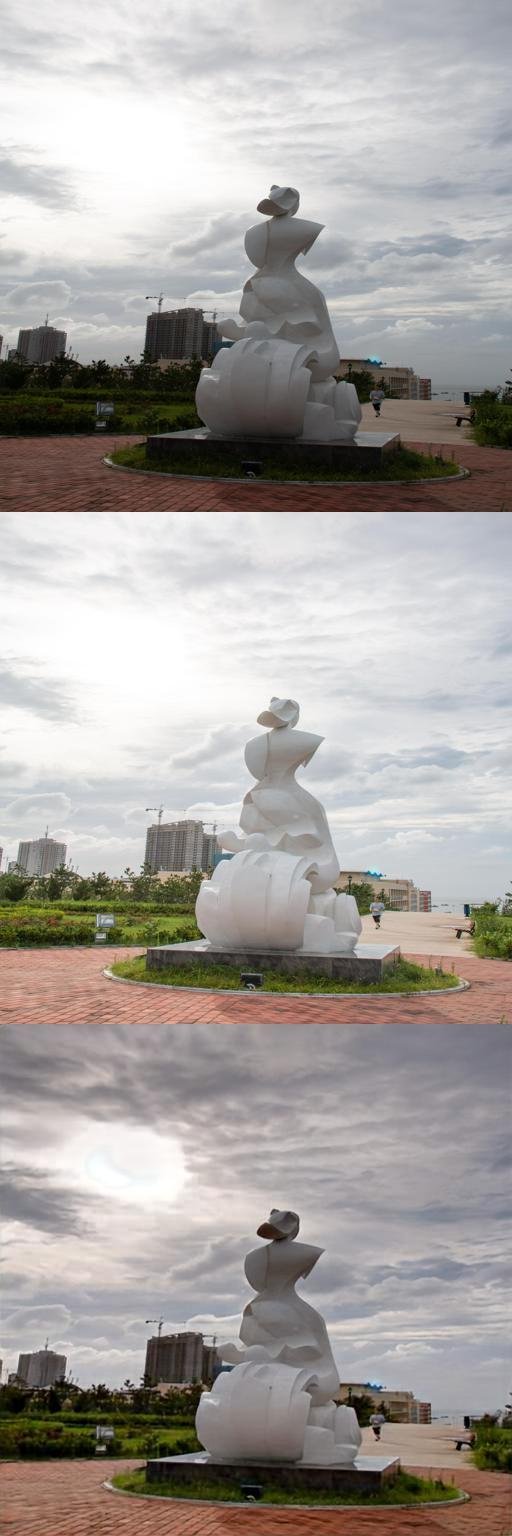}
    \end{subfigure}
    \begin{subfigure}{0.16\textwidth}
        \includegraphics[width=\textwidth]{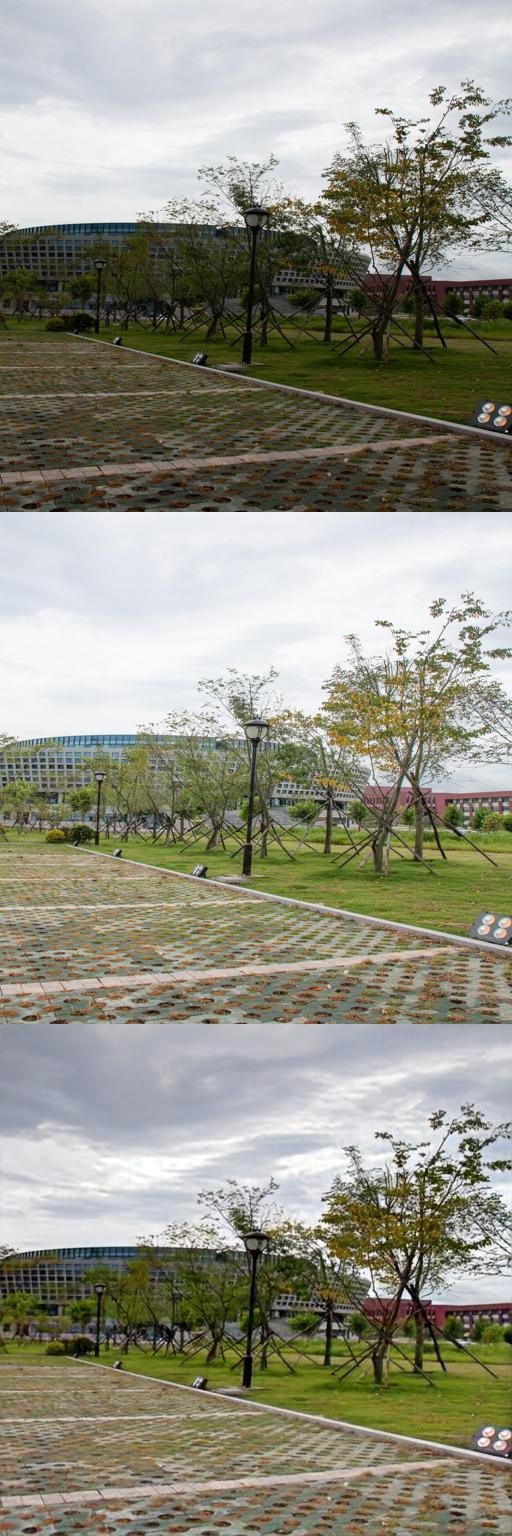}
    \end{subfigure}
    \begin{subfigure}{0.16\textwidth}
        \includegraphics[width=\textwidth]{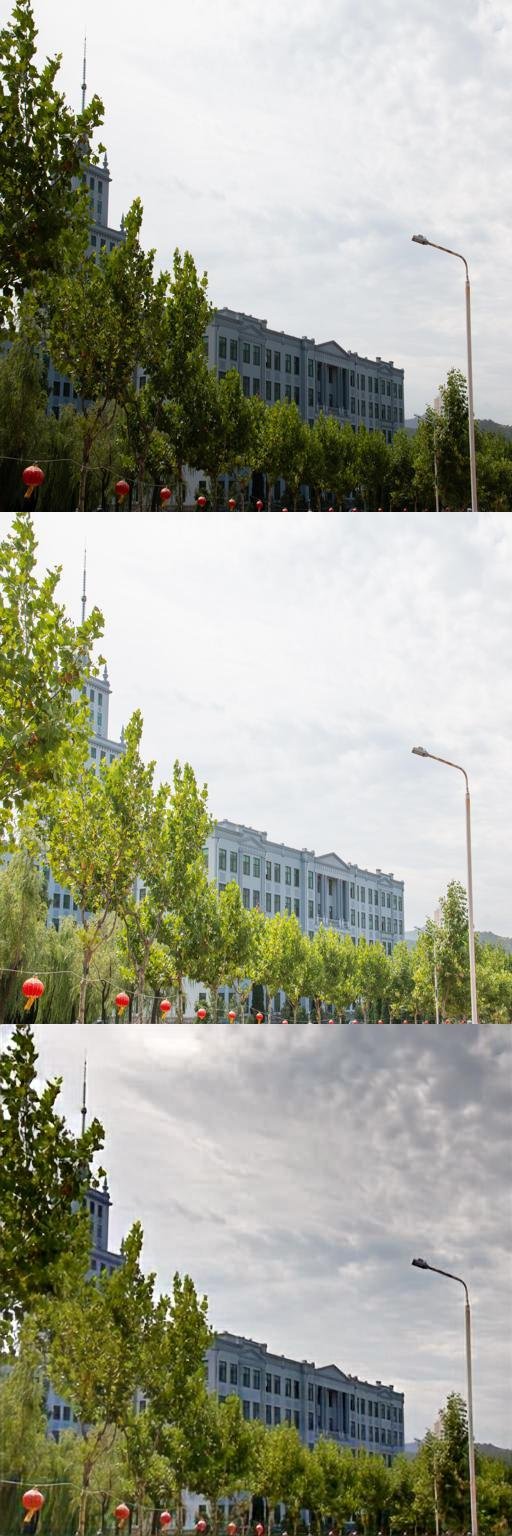}
    \end{subfigure}
    \begin{subfigure}{0.16\textwidth}
        \includegraphics[width=\textwidth]{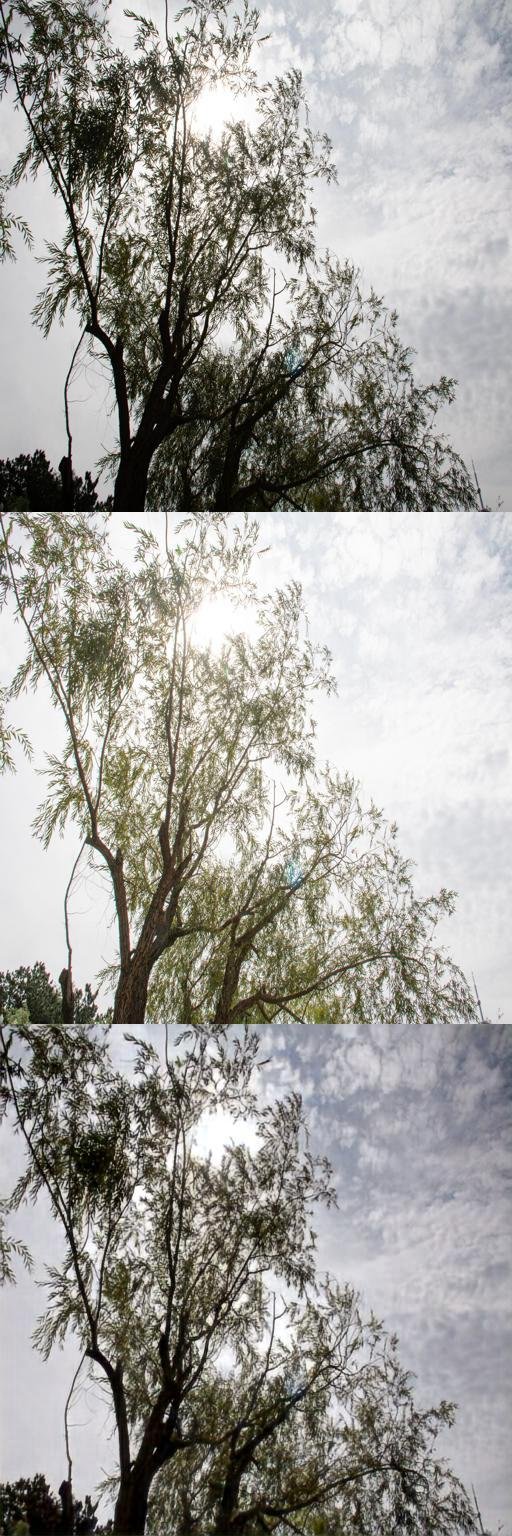}
    \end{subfigure}
    \begin{subfigure}{0.16\textwidth}
        \includegraphics[width=\textwidth]{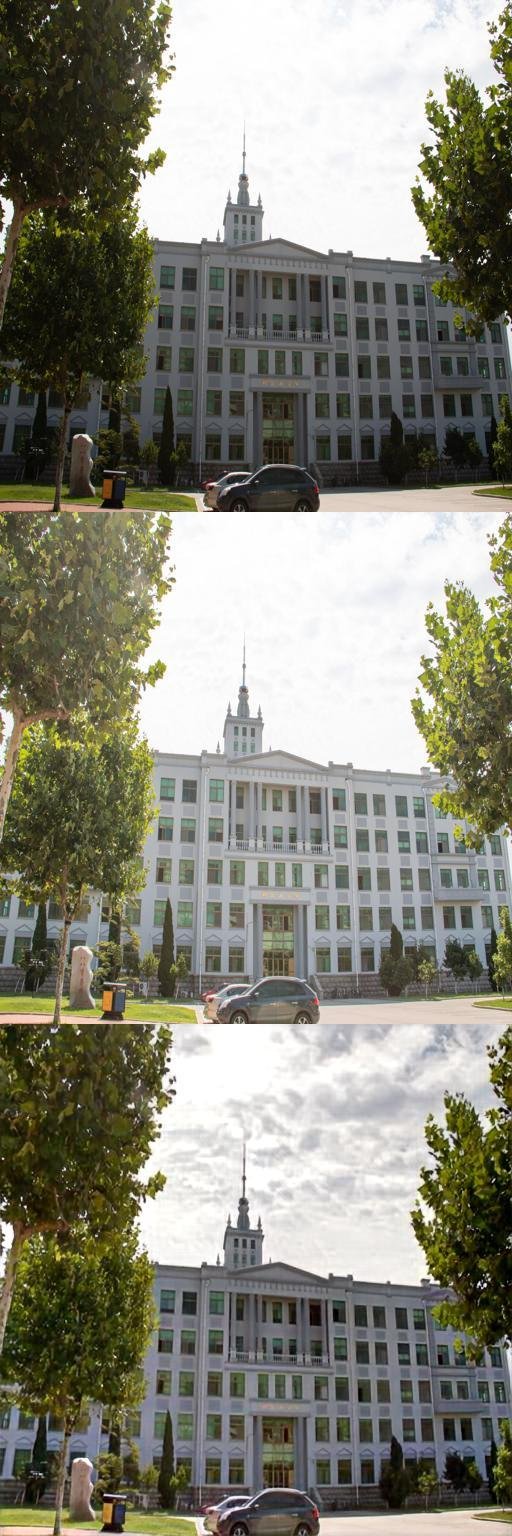}
    \end{subfigure}
    \caption{Visual comparison with GT on the BAID~\cite{lv2022backlitnet} dataset. }
    \label{fig:GT_CMP_BAID}
\end{figure*}

\begin{figure*}[htbp]
    \centering
    \makebox[0pt][r]{\raisebox{6.5cm}{\rotatebox{90}{Input}}\hspace{0.1cm}}%
    \makebox[0pt][r]{\raisebox{4cm}{\rotatebox{90}{GT}}\hspace{0.1cm}}%
    \makebox[0pt][r]{\raisebox{1cm}{\rotatebox{90}{Ours}}\hspace{0.1cm}}%
    \begin{subfigure}{0.16\textwidth}
        \includegraphics[width=\textwidth]{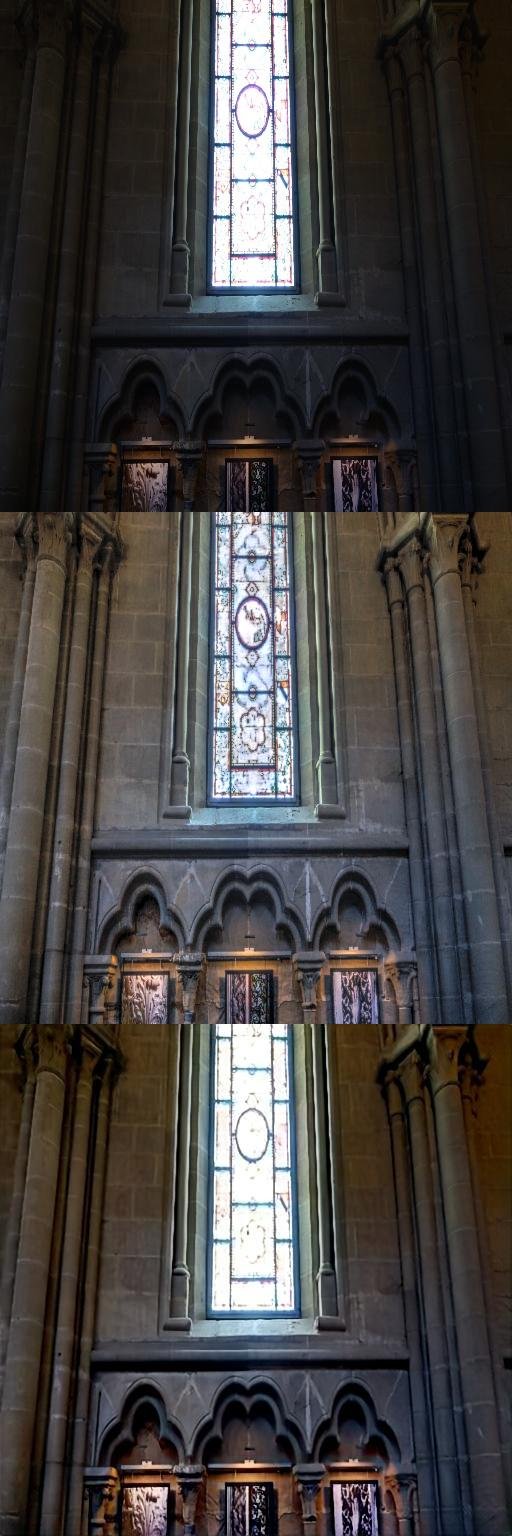}
    \end{subfigure}
    \begin{subfigure}{0.16\textwidth}
        \includegraphics[width=\textwidth]{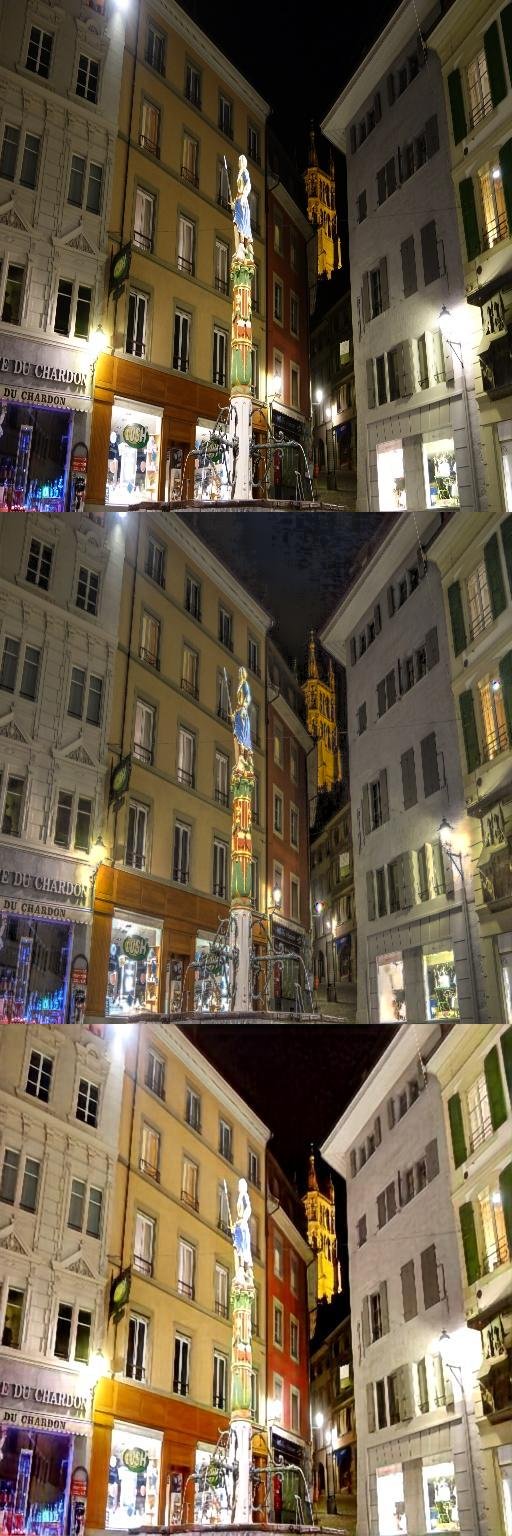}
    \end{subfigure}
    \begin{subfigure}{0.16\textwidth}
        \includegraphics[width=\textwidth]{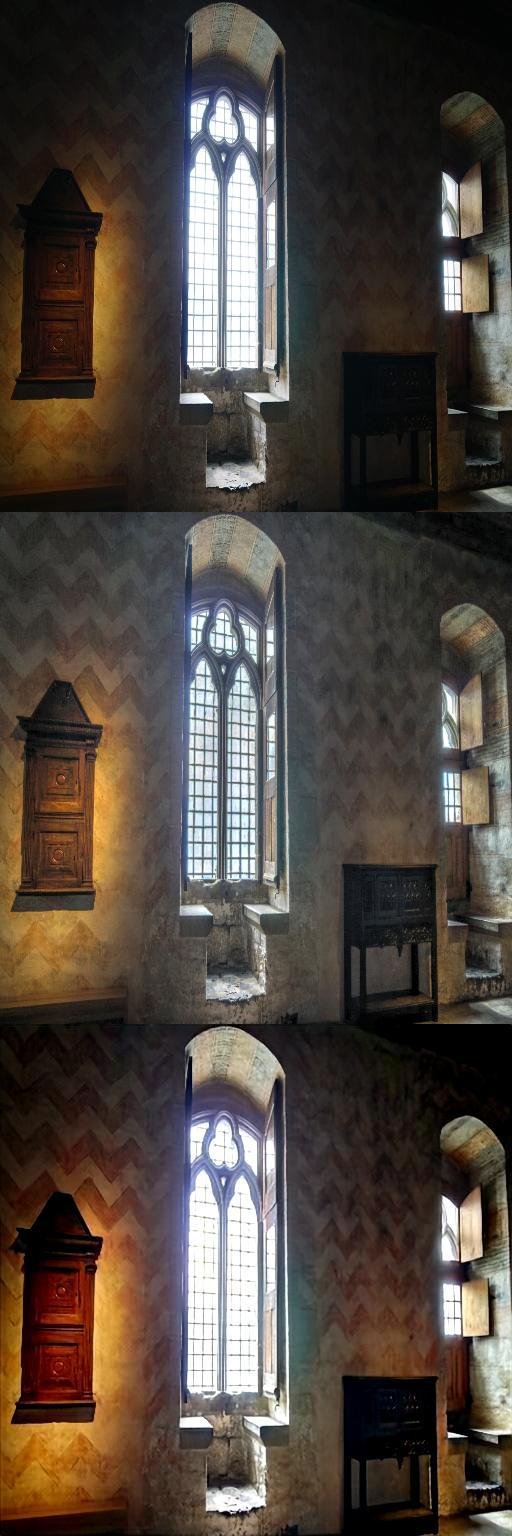}
    \end{subfigure}
    \begin{subfigure}{0.16\textwidth}
        \includegraphics[width=\textwidth]{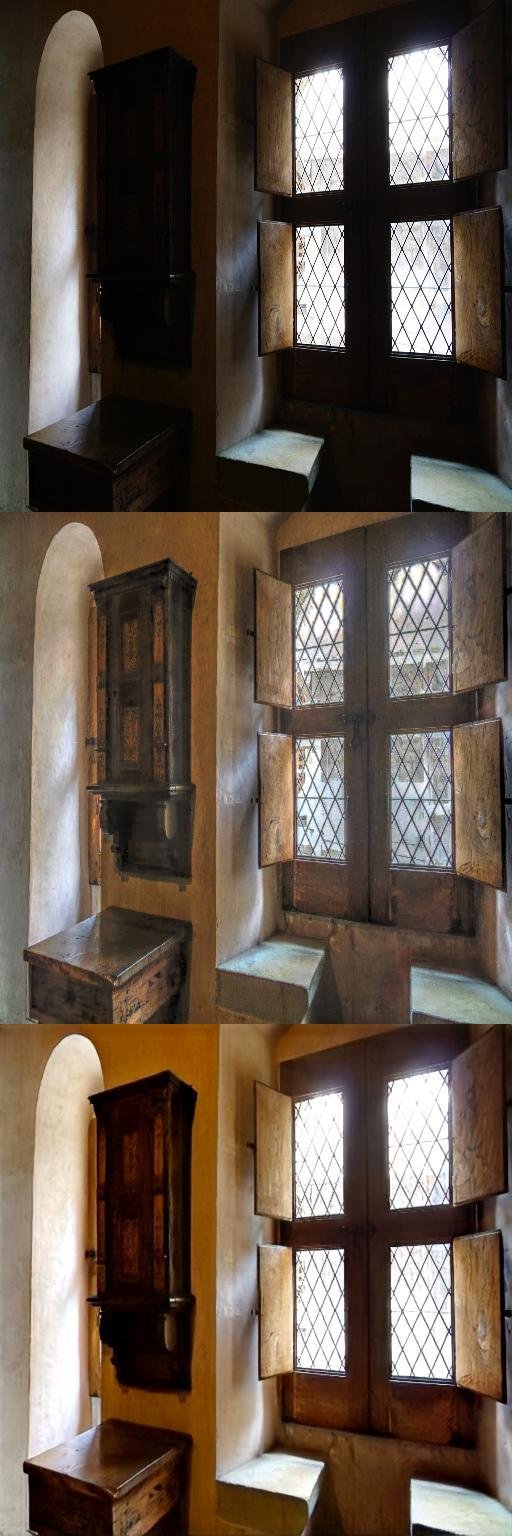}
    \end{subfigure}
    \begin{subfigure}{0.16\textwidth}
        \includegraphics[width=\textwidth]{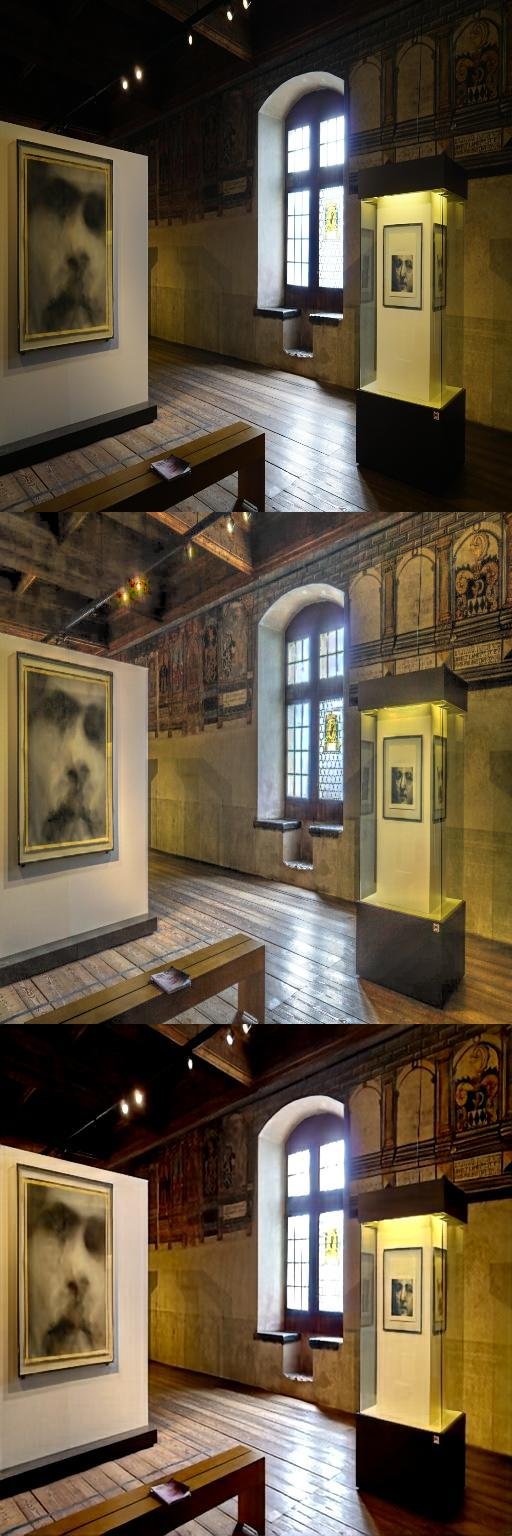}
    \end{subfigure}
    \begin{subfigure}{0.16\textwidth}
        \includegraphics[width=\textwidth]{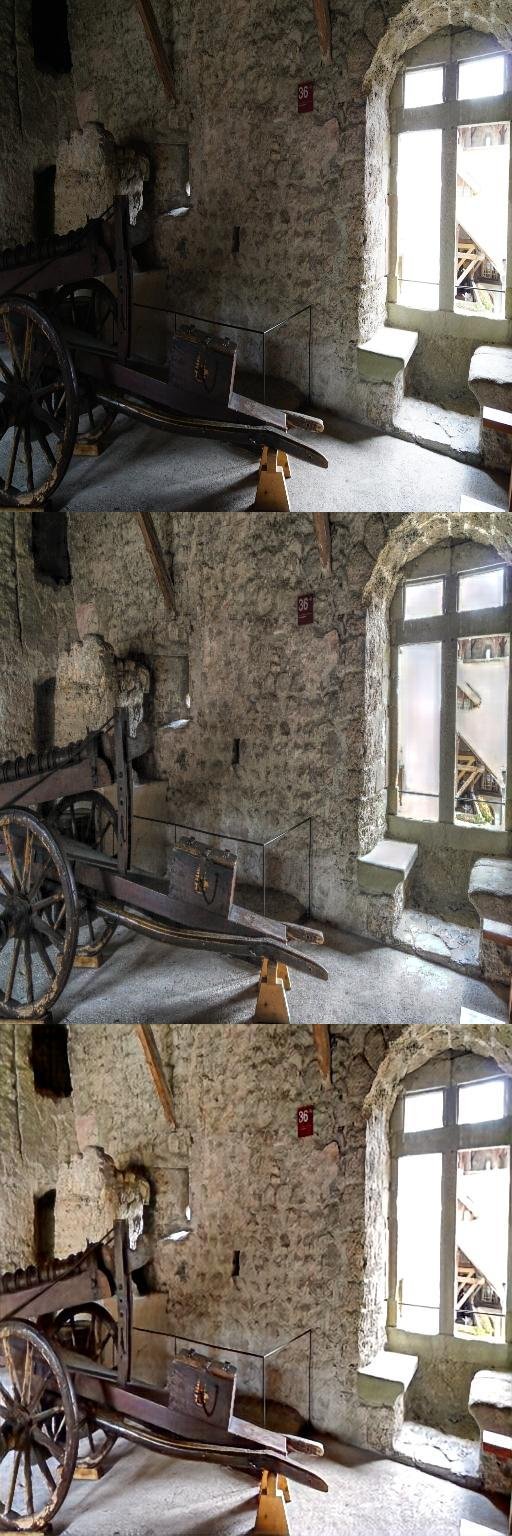}
    \end{subfigure}
    \caption{Visual comparison with GT on the HDREye~\cite{liu2020reverse} dataset. }
    \label{fig:GT_CMP_HDREye}
\end{figure*}

\begin{figure*}
    \centering
    \makebox[\textwidth][c]{Input \hspace{2cm} w/o fusion \hspace{2cm} Ours \hspace{2cm} GT} \\
    \includegraphics[width=0.8\linewidth]{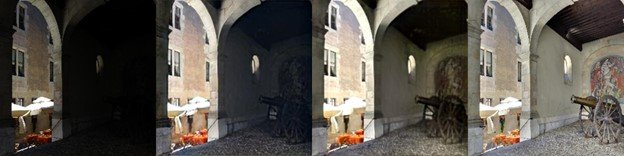} \\
    \includegraphics[width=0.8\linewidth]{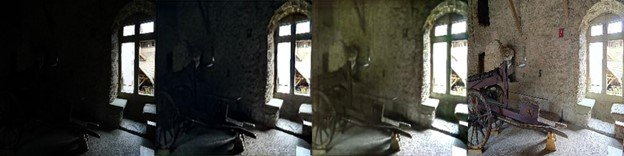} \\
    \includegraphics[width=0.8\linewidth]{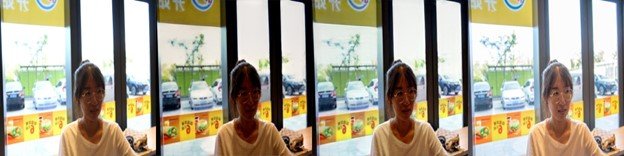} \\
    \includegraphics[width=0.8\linewidth]{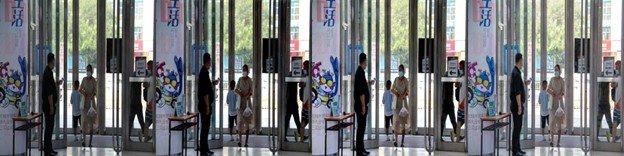} \\
    \includegraphics[width=0.8\linewidth]{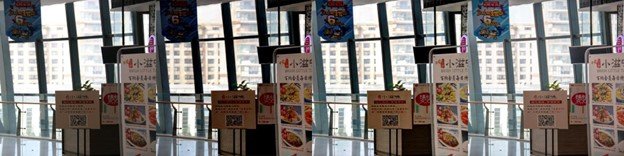}
    \caption{Visualized results of ablation study.}
    \label{fig:ablation_study}
\end{figure*}

\clearpage



\end{document}